\documentclass[]{jair}

\setcopyright{cc}
\copyrightyear{2026}
\acmDOI{10.1613/jair.1.20433}

\JAIRAE{Chang Xu}
\JAIRTrack{} 
\acmVolume{86}
\acmArticle{16}
\acmMonth{06}
\acmYear{2026}

\RequirePackage[
  datamodel=acmdatamodel,
  style=acmauthoryear,
  backend=biber,
  giveninits=true
  ]{biblatex}

\usepackage{theapa}
\usepackage{float}
\usepackage{booktabs}
\usepackage{graphicx,psfrag,epsf}
\usepackage{amsfonts}

\usepackage{amssymb}
\usepackage{amsthm}
\usepackage{amsmath}
\usepackage{stackengine}
\usepackage{tablefootnote}
\newcommand{\Lagr}{\mathcal{L}}
\newcommand{\prob}{\mathbb{P}}
\newcommand{\A}{\mathbb{A}}

\newcommand{\E}{\mathbb{E}}

\newcommand{\reals}{\mathbb{R}}
\newcommand{\nat}{\mathbb{N}}
\newcommand{\N}{\mathcal{N}}
\newcommand{\F}{\mathcal{F}}

\newcommand{\M}{\mathcal{M}}
\newcommand{\indicator}{\mathcal{I}}
\usepackage{caption}
\usepackage{subcaption}
\usepackage{multirow}
\newtheorem{proposition}{Proposition}
\newtheorem{corollary}{Corollary}
\newtheorem{remark}{Remark}

\begin{document}

\title[Input-Skip Latent Binary Bayesian Neural Networks]{Explainable Bayesian Deep Learning through Input-skip Latent Binary Bayesian Neural Networks}


\author{Eirik Høyheim}
\orcid{0009-0007-4626-756X}
\authornote{Corresponding Author.}
\email{eirik.hoyheim@ffi.no}
\author{Lars Skaaret-Lund}
\orcid{0009-0002-7043-7841}
\email{lars.skaaret-lund@nmbu.no}
\author{Solve Sæbø}
\orcid{0000-0001-8699-4592}
\email{solve.sabo@nmbu.no}
\author{Aliaksandr Hubin}
\orcid{0000-0002-3244-6571}
\email{aliaksandr.hubin@nmbu.no}
\affiliation{%
  \institution{KBM, Bioinformatics and applied statistics (BIAS), Norwegian University of Life Sciences}
  \city{Ås}
  \country{Norway}}

\renewcommand{\shortauthors}{Høyheim, Skaaret-Lund, Sæbø \& Hubin}

\begin{abstract}
Modeling natural phenomena with artificial neural networks (ANNs) often provides highly accurate predictions. However, ANNs often suffer from over-parameterization, complicating interpretation and raising uncertainty issues. Bayesian neural networks (BNNs) address the latter by representing weights as probability distributions, allowing for predictive uncertainty evaluation. Latent binary Bayesian neural networks (LBBNNs) further handle structural uncertainty and sparsify models by removing redundant weights. This article advances LBBNNs by enabling covariates to skip to any succeeding layer or be excluded, simplifying networks and clarifying input impacts on predictions. This further allows us to learn simpler structures (e.g., linear or even constant intercept only models) when appropriate. Furthermore, the input-skip LBBNN (ISLaB) approach reduces network density significantly compared to standard LBBNNs, achieving over 99\% reduction for small networks and over 99.9\% for larger ones, while still maintaining high predictive accuracy and uncertainty quantification. For example, on MNIST, we reached 97\% accuracy and great calibration with just 935 weights, reaching state-of-the-art for compression of neural networks. Furthermore, the proposed method accurately identifies the true covariates and adjusts for system non-linearity. The main contribution is the introduction of active paths, enhancing directly designed global and local explanations within the LBBNN framework. The latter are exact with theoretical guarantees and do not require post hoc external tools.
\end{abstract}

\received{15 September 2025}
\received[accepted]{8 January 2026}

\maketitle

\section{Introduction}

In recent years, artificial intelligence (AI) and machine learning (ML) have entered the colloquial language, largely due to the availability of highly sophisticated neural network models like GPT-4 \citep{openai2023gpt4}. These models are extensions of standard multilayer perceptrons, utilizing more advanced transformer architectures with billions of weights to map inputs to predictions. A key theoretical result associated with ANNs is the universal approximation theorem \citep{cybenko1989approximation, hornik1989multilayer}, which states that an ANN can approximate any given function when provided with sufficient data and given that the number of neurons in the hidden layer is large enough. Hence, there is no surprise that huge GPT-4 models trained on vast amounts of data collected from the internet perform well on approximating common knowledge. Yet, despite these impressive capabilities in function approximation, ANNs lack interpretability due to their complexity, making it difficult to understand predictive uncertainty or the contribution of individual covariates. Moreover, the lack of uncertainty handling does not facilitate proper ``doubt'' handling or ``I don't know'' responses in ANNs like GPT-4 \citep{papamarkou2024position}.

Traditional Bayesian neural networks (BNNs) provide a principled framework for uncertainty quantification by placing priors on network parameters and inferring a posterior distribution \citep{neal1992bayesian}. However, exact inference is intractable for modern architectures, which has motivated approximate approaches \citep{papamarkou2024position}. Two particularly influential methods are Monte Carlo Dropout \citep{gal2016dropout} and Deep Ensembles \citep{lakshminarayanan2017simple}. MC Dropout can be interpreted as approximate variational inference by applying dropout at test time, while Deep Ensembles estimate uncertainty by averaging predictions from multiple independently trained networks. Although both methods have demonstrated empirical success in improving calibration and robustness, they remain heuristic approximations to Bayesian inference: MC Dropout relies on a variational relaxation that does not capture the full posterior, and ensembles approximate uncertainty through model diversity rather than posterior probabilities. Furthermore, they remain memory- and computation-intensive, since they do not induce model sparsity and require either multiple forward passes (MC Dropout) or training several large networks (ensembles). Moreover, these methods do not yield compact representations that can facilitate explainability, but rather focus exclusively on predictive uncertainty. This highlights the need for alternative Bayesian approaches that can simultaneously provide well-calibrated uncertainty, efficiency through sparsity, and built-in interpretability.

Latent binary Bayesian neural networks (LBBNNs) \citep{hubin2019combining, bai2020efficient} address complexity by structure learning for model uncertainties. This is done by introducing trainable indicator variables for each weight. Thus, each combination of on and off weights produces a separate model. The probability of inclusion of each weight is then approximated. Finally, heavy sparsification is possible through a median probability model (MPM), which consists of weights with inclusion probabilities above 0.5. Thus, one can drastically reduce weight density by learning the appropriate neural network structure for a given problem. Also, incorporating model uncertainties in addition to parameter uncertainties was empirically shown to provide better predictive calibration for both images and tabular data in \citet{skaaret2023sparsifying}. Despite these important benefits, LBBNN models can still be unnecessarily complex, with thousands of weights and nonlinear transformations, hindering interpretability.

Specifically, the most significant limitation for existing LBBNNs is that covariates must pass through all hidden layers and nonlinear activation functions even if the MPM is extremely sparse. This approach might be unnecessary, as the underlying relationship between the input and output might be much simpler, e.g. linear. Furthermore, using overly complex models complicates interpretability, as more connections must be considered when examining the global explanation of a model. Additionally, larger models with more applications of matrix multiplications than necessary have negative environmental impacts due to the increased computational requirements for predictions and the additional storage space needed to save the models.

Standard LBBNNs force covariates through all layers, obscuring input-output relationships and inflating model size. To address these issues, we extend the LBBNN methods by introducing input-skip, allowing all covariates to enter directly any given layer. This approach enables the model to learn the most optimal structure for the problem at hand and automatically infer optimal depth and width of the layers. For instance, if the true underlying relationship between the input and output is linear, the learned model should be a linear function, meaning all other weights are redundant and should be removed. Similarly, if the relationship is nonlinear, the model should utilize the appropriate amount of nonlinear activation functions in the hidden layers and ignore all excessive connections in the initialized model. This is in stark contrast to traditional approaches, where one has to decide on a model \textit{a priori}. It also avoids extensive combinatorial exploration like in neural architecture search \citep{elsken2019neural} or in Bayesian generalized nonlinear models \citep{hubin2021flexible}. Further, local and global explanation methods are provided for LBBNN with input-skip. These explanations are built into the design of our model and do not need any external approximations like SHAP \citep{lundberg2017unified} or SAGE \citep{covert2020understanding}, thus avoiding both additional expensive computations and using approximations to explain other approximations that do not propagate uncertainties to the downstream task properly.  
The innovative input-skip LBBNN (ISLaB) approach reduces network density by over 99\% for small networks and over 99.9\% for larger ones, maintaining high accuracy and uncertainty evaluation. Specifically, on MNIST, ISLaB achieves 97\% accuracy with just 935 weights, which is a new efficiency frontier compared to prior sparse models like Lottery Ticket ($\sim$1,300 weights, 95\%) \citep{frankle2019lottery}. In addition, we demonstrate that our method consistently obtains much sparser networks than the baseline methods, while maintaining high predictive power and good calibration. Our method also has the advantage of not requiring any post-training, ad-hoc pruning, or post-hoc temperature scaling for that; \textit{sparsification, explainability and uncertainty handling are inherently built into the method's design}.
To summarize, this article introduces the following innovations and contributions:

\begin{itemize}
    
\item Proposes ISLaB, an LBBNN method that allows the covariates to skip to any given layer or be entirely excluded from the model, simplifying the network architecture and clarifying input contributions to predictions.
\item A novel method that enhances global and local understanding of trained networks, providing deeper insights into the contributions of individual covariates to the overall predictions.
\end{itemize}

\section{Materials and Methods}

\begin{figure}[htp]
\centering
\includegraphics[width=0.5\linewidth]{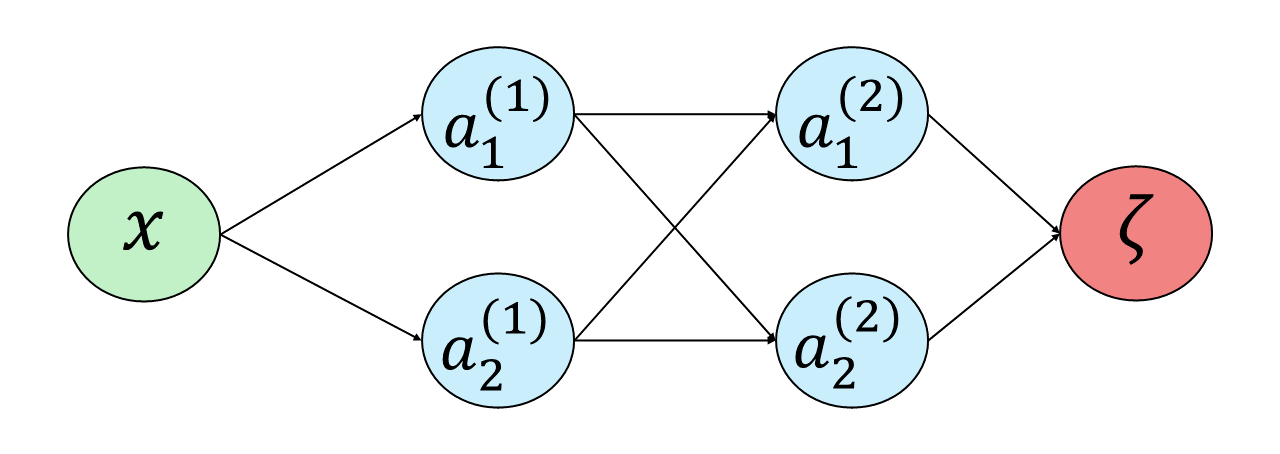}
\caption{A simple ANN architecture without bias nodes.}\label{fig:simpleANN}
\end{figure}

A neural network model connects the output $\textbf{\textit{y}}_i \in \reals^c$ to the input $\textbf{\textit{x}}_i \in \reals^v$ via a probability distribution parametrized with a trainable mapping $\zeta(\textbf{\textit{x}}_i)$, as shown in Figure \ref{fig:simpleANN}:

\begin{equation}
\textbf{\textit{y}}_i \sim f(\cdot; \zeta(\textbf{\textit{x}}_i),\phi).
\end{equation}
Here, $f$ represents the distribution modeling the relationship between covariates and responses through its trainable parameters $\zeta(\textbf{\textit{x}}_i)$ conditioning on the former and known fixed parameters $\phi$, which could be the dispersion parameter for instance. For example, in binary classification ($c=1$), a Bernoulli distribution might be used with $\zeta(\textbf{\textit{x}}_i)$ being the probability of success and $\phi$ being 1, while a Gaussian distribution with $\zeta(\textbf{\textit{x}}_i)$ consisting of the mean and the variance being $\phi$ is suitable for regression tasks. For multivariate classification problems ($c > 2$), a categorical distribution is typically assumed. The parameters $\zeta(\cdot)$, necessary for these distributions, are obtained by passing the covariates through a composition of hidden layers. Each hidden layer consists of nodes computed using a semi-affine transformation of the signal coming from the preceding layer:

\begin{equation}\label{eq:nn}
a^{(j)}_p = o^{(j)} \bigg(a_0^{(j-1)} w_{0,p}^{(j)} + \sum_{k=1}^K a_k^{(j-1)} w_{k,p}^{(j)}\bigg).
\end{equation}

In Equation (\ref{eq:nn}), $k=1,2,..., K$ represents the number of hidden nodes in the preceding layer, $p=1,2,..., P$ indicates the number of hidden nodes in the succeeding layer, and $j=1,2,..., J$ denotes the layer positions, where $0$ is the input layer, $\textbf{a}^{(0)} = \textbf{\textit{x}}$, and $J$ is the output layer, $\textbf{a}^{(J)} = \zeta(\textbf{\textit{x}}_i)$. The weight connecting the $k$-th node in layer $j-1$ to the $p$-th node in layer $j$ is denoted by $w_{k,p}^{(j)}$, $a_k^{(j-1)}$ is the activation of the $k$-th node in layer $j-1$, and $a_p^{(j)}$ is the activation to be computed for the $p$-th node in layer $j$. The bias node and its associated weight are represented by $a_0^{(j-1)}$ and $w_{0,p}^{(j)}$, respectively. For all layers, $a_0^{(j-1)}=1$, thus it is omitted in subsequent equations.

\subsection{Bayesian Neural Networks}
\begin{figure}[htp]
  \centering
    \includegraphics[width=\linewidth]{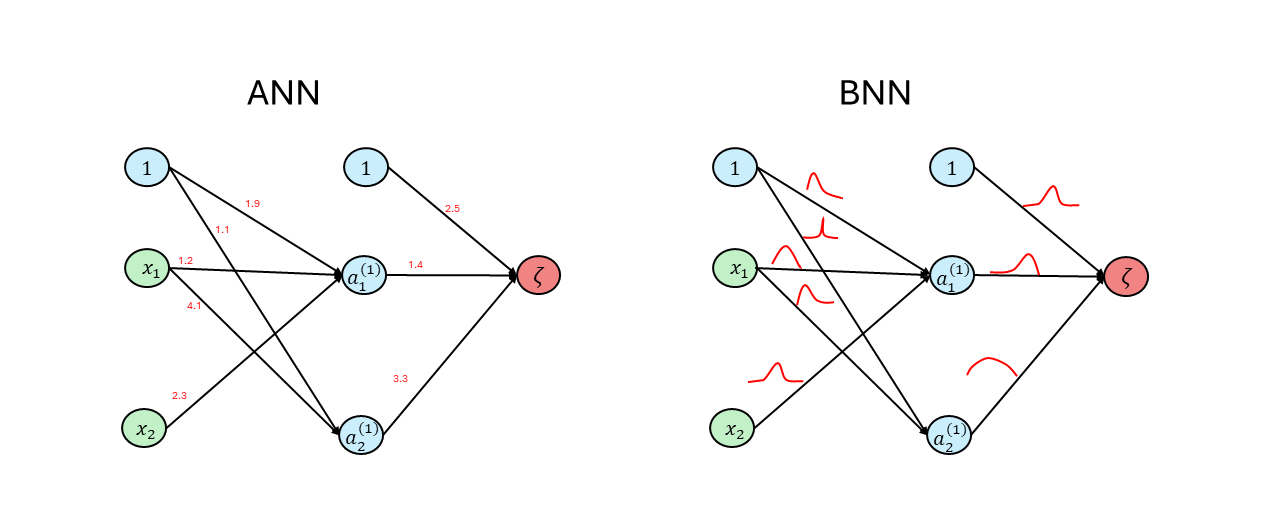} 
  \caption{Illustration of simple ANN (on the left) and BNN (on the right) architectures.}\label{fig:annVSbnn}
\end{figure}

A Bayesian neural network (BNN) is a variation of an ANN where weights are represented as probability distributions instead of scalars, as illustrated in Figure \ref{fig:annVSbnn}. 

In the Bayesian paradigm, all unknown parameters are treated as random variables. Given a prior distribution over the weights, $p(\mathbf{W})$, with $\mathbf{W} = \{w_{k,p}^{(j)}, k = 1,
\cdots, K, p = 1, \cdots P, j = 1, \cdots J\}$, a likelihood function  $\Lagr(D|\mathbf{W})$, and marginal likelihood, $\F(D)$, we find the posterior weight distribution $\pi(\mathbf{W}|D)$ through Bayes' theorem:

\begin{equation}\label{eq:BayesThm}
    \pi(\mathbf{W}|D) = \frac{\Lagr(D|\mathbf{W})p(\mathbf{W})}{\F(D)}.
\end{equation}
Here, $\mathbf{W}$ represents the weight parameters and $D$ is the available data. The posterior weight distribution allows making new predictions, $y_{\text{new}}$, given unseen data, $\textbf{\textit{x}}_{\text{new}}$ by the posterior predictive distribution: 

\begin{align}
    \pi(y_{\text{new}}|\textbf{\textit{x}}_{\text{new}}, D) &= \int_\mathbf{W} \pi(y_{\text{new}}, \mathbf{W}| \textbf{\textit{x}}_{\text{new}}, D) \text{d}\mathbf{W} \nonumber \\ 
    &= \int_\mathbf{W} \pi(y_{\text{new}}|\textbf{\textit{x}}_{\text{new}}, \mathbf{W}, D)\pi(\mathbf{W}|D) \text{d}\mathbf{W} \nonumber \\ 
    &= \E_{\pi(\mathbf{W}|D)}[\pi(y_{\text{new}}|\textbf{\textit{x}}_{\text{new}}, \mathbf{W}, D)].\label{eq:predBNN} 
\end{align}

Since exact evaluations of the integral \eqref{eq:predBNN} are computationally prohibitive due to the ultra-high dimensionality of $\mathbf{W}$, approximations like Monte Carlo (MC) are used:

\begin{equation}
    \pi(y_{\text{new}}|\textbf{\textit{x}}_{\text{new}}, D) \approx \frac{1}{N} \sum_{i=1}^N \pi(y_{\text{new}}|\textbf{\textit{x}}_{\text{new}}, \mathbf{W}_{(i)}, D), \mathbf{W}_{(i)} \sim \pi(\mathbf{W}|D). \label{eq:predMC}
\end{equation}

Equation (\ref{eq:predMC}) is the MC approximation of the integral, where random weights, $\mathbf{W}_{(i)}$, are drawn $N$ times from the posterior weight distribution. A credibility interval for the prediction can be obtained as empirical quantiles from these $N$ samples, allowing to describe the confidence in a prediction.

\subsection{Latent Binary Bayesian Neural Networks}

Neural networks, with their potential to incorporate millions or even billions of weights, present significant challenges in terms of interpretability and computational cost. This complexity is further amplified in BNNs, as weights are represented through probability distributions. To attain more interpretable models, structural reduction in the number of weights within a network is imperative.

\begin{figure}[htp]
  \centering
    \includegraphics[width=\linewidth]{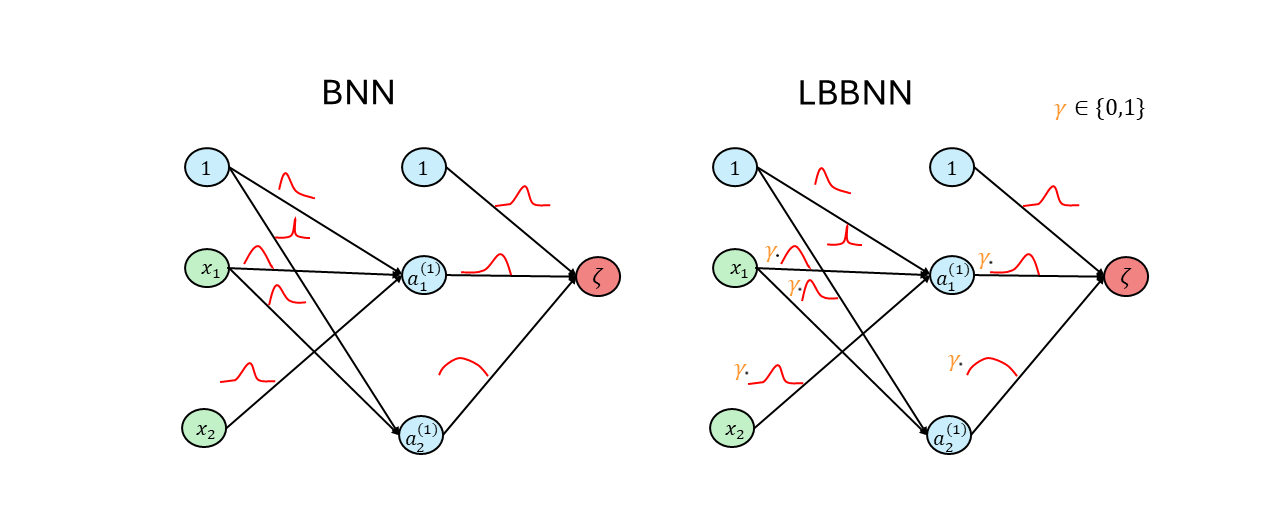} 
  \caption{Illustration of the difference between BNNs (left) and LBBNNs (right).}\label{fig:bnnVSlbbnn}
\end{figure}

To learn the neural network structures over which weights are required in BNNs, \citet{hubin2019combining} proposed associating each weight with a trainable latent indicator variable. These indicator variables facilitate sparsification by learning whether a weight should be included or excluded, and thus identifying an appropriate network structure for a specific problem through the posterior probabilities of weight inclusions allowing for automatic edge pruning. This model, known as latent binary Bayesian neural networks (LBBNNs), incorporates an inclusion parameter, $\gamma_{k,p}^{(j)}\in\{0,1\}$, into the node $p$ output in a given layer $j$ of a model as follows:

\begin{equation}\label{eq:SLNodeCalc}
    a^{(j)}_p = o^{(j)}\bigg(w_{0,p}^{(j)} + \sum_{k=1}^K \gamma_{k,p}^{(j)} a_k^{(j-1)} w_{k,p}^{(j)}\bigg),
\end{equation}
where $w_{k,p}^{(j)}$ is an entry of $\mathbf{W}$ corresponding to the $k$-th weight of neuron $p$ in layer $j$. 
In Equation (\ref{eq:SLNodeCalc}), $\gamma$ is not included in front of the bias weight, following \citet{skaaret2023sparsifying} and \citet{hubin2024sparse}, so the bias distribution remains consistent with the one presented for BNNs. An LBBNN model comprises all weights, $\mathbf{W}$, and their associated inclusion parameters $\gamma_{k,p}^{(j)}$ forming a matrix $\mathbf{\Gamma}$, that defines the network structure. Thus, we have one extra parameter per weight compared to standard BNNs, but with the potential to prune away a large proportion of the weights by inferring the structure of the MPM. To incorporate both weight and structural uncertainty into an LBBNN model, the spike-and-slab prior is used, assuming independence between weights within and between layers \citep{hubin2019combining,skaaret2023sparsifying}:

\begin{align}
    p\left(w_{k,p}^{(j)}|\gamma_{k,p}^{(j)}\right) &= \gamma_{k,p}^{(j)} \text{ } \N\left(w_{k,p}^{(j)}; 0, \left(\tau_{k,p}^{(j)}\right)^2\right) + \left(1-\gamma_{k,p}^{(j)}\right) \delta_0\left(w_{k,p}^{(j)}\right) \label{eq:priorLBBNN} \\
    p\left(\gamma_{k,p}^{(j)}\right) &= \text{Bernoulli}\left(\gamma_{k,p}^{(j)}; \psi_{k,p}^{(j)}\right), \qquad \psi_{k,p}^{(j)} \in [0, 1].
\end{align}

The ``spike'' part of the prior is denoted by $\delta_0$, a probability (mass) at zero, and the ``slab'' part is represented by a Gaussian distribution in the most common definition, although other continuous distributions can be used, for instance, the wider t-distribution, as in \citet{hubin2024sparse}. The main objective is to find the posterior distribution, $\pi(\mathbf{W},\mathbf{\Gamma}|D)$, to make predictions on unseen data. The predictive distribution extends Equation (\ref{eq:predBNN}) to consider all $2^{||\mathbf{\Gamma}||}$ models forming a model space $\M$, where each model is defined by a specific pattern of on and off weights and $||\mathbf{\Gamma}||$ counts the number of elements of the matrix of latent binary indicators. The predictive distribution accounting for model uncertainty is:

\begin{equation}\label{eq:predDistLBBNN}
\pi(y_{\text{new}}|\textbf{\textit{x}}_{\text{new}}, D) = \sum_{\mathbf{\Gamma}\in \M} \int_\mathbf{W} \pi(y_{\text{new}}|\textbf{\textit{x}}_{\text{new}}, \mathbf{W}, \mathbf{\Gamma}, D)\pi(\mathbf{W}, \mathbf{\Gamma}|D) \text{d}\mathbf{W}.
\end{equation} 

Even more severely than for standard BNNs, the predictive distribution is unattainable due to the difficulty of obtaining the exact posterior distribution. This makes it more convenient to use an approximate posterior distribution through the variational inference (VI)-based framework. 

A common choice for the approximate posterior distribution $\pi(\mathbf{W}, \mathbf{\Gamma}|D)$ is obtained by factorizing into the product of independent approximations of  $\pi(w_{k,p}^{(j)},\gamma_{k,p}^{(j)}|D)\approx q_{\eta_{k,p}^{(j)}}(w_{k,p}^{(j)},\gamma_{k,p}^{(j)}) = q_{\theta_{k,p}^{(j)}}(w_{k,p}^{(j)}|\gamma_{k,p}^{(j)})q_{\alpha_{k,p}^{(j)}}(\gamma_{k,p}^{(j)})$ with $\eta_{k,p}^{(j)} = (\theta_{k,p}^{(j)}, \alpha_{k,p}^{(j)}) = (\mu_{k,p}^{(j)}, \sigma_{k,p}^{(j)},\alpha_{k,p}^{(j)})$ through the components defined as follows:
\begin{equation}\label{eq:varPostLBBNN}
\begin{split}
    q_{\theta_{k,p}^{(j)}}\left(w_{k,p}^{(j)}|\gamma_{k,p}^{(j)}\right) &= \gamma_{k,p}^{(j)} \N\left(w_{k,p}^{(j)};\mu_{k,p}^{(j)},\left({\sigma_{k,p}^{(j)}}\right)^2\right) + \left(1-\gamma_{k,p}^{(j)}\right) \delta_0\left(w_{k,p}^{(j)}\right) \\
    q_{\alpha_{k,p}^{(j)}}\left(\gamma_{k,p}^{(j)}\right) &= \text{Bernoulli}\left(\gamma_{k,p}^{(j)}; \alpha_{k,p}^{(j)}\right)
    , \qquad \alpha_{k,p}^{(j)} \in [0, 1].
\end{split}
\end{equation}
This is known as mean-field approximations of LBNNN's posterior \citep{hubin2024sparse}. The goal is to find the parameters of the variational posterior, $$q_{\boldsymbol\eta}(\mathbf{W}, \mathbf{\Gamma}) = \prod_{k,p,j} q_{\eta_{k,p}^{(j)}}(w_{k,p}^{(j)},\gamma_{k,p}^{(j)})$$ that makes it as close as possible to the true posterior $\pi(\mathbf{W}, \mathbf{\Gamma}|D)$, where closeness is typically measured by the KL-divergence. Optimization with respect to the vector of parameters $\boldsymbol\eta$ can be done using numerical optimization algorithms like batch stochastic gradient descent or Adam \citep{kingma2014adam}:

\begin{align}
    \boldsymbol\eta^* &= \stackunder{arg min}{$\eta$} \text{ KL}\bigg(q_\eta(\mathbf{W}, \mathbf{\Gamma}) || \pi(\mathbf{W}, \mathbf{\Gamma}|D)\bigg) \nonumber \\
    &= \stackunder{arg min}{$\boldsymbol\eta$} \int_\mathbf{W} \sum_{\mathbf{\Gamma}} q_\eta(\mathbf{W}, \mathbf{\Gamma})\left(\log \frac{q_\eta(\mathbf{W}, \mathbf{\Gamma})}{p(\mathbf{W}, \mathbf{\Gamma})} - \log \Lagr(D|\mathbf{W}, \mathbf{\Gamma}) \right) \text{d}\mathbf{W} \nonumber \\
    &= \stackunder{arg min}{$\boldsymbol\eta$} \text{ KL}\bigg(q_\eta(\mathbf{W}, \mathbf{\Gamma}) || p(\mathbf{W}, \mathbf{\Gamma})\bigg) - \E_{q_\eta(\mathbf{W}, \mathbf{\Gamma})}\bigg[\log\Lagr(D|\mathbf{W}, \mathbf{\Gamma})\bigg]. \label{eq:negELBOLBBNN}
\end{align}

To ensure that $\alpha$ remains between $0$ and $1$ during the training process, it is conventional to define $\alpha = \frac{1}{1+\exp(-\lambda)}$, where $\lambda\in\reals$. 
The reparameterization trick and a gradient-based method are commonly used to find an appropriate $\boldsymbol\eta^*$ efficiently. However, as entries of $\mathbf{\Gamma}$ are binary, the loss is not differentiable. Solutions include using a relaxation to make entries of $\mathbf{\Gamma}$ continuous \citep{hubin2024sparse} or using the local reparametrization trick \citep{kingma2015variational,skaaret2023sparsifying}, with the latter being more computationally efficient due to sampling the (approximate) Gaussian pre-activations directly, in addition to not needing any relaxations. 

Mean-field approximations, like Equation \eqref{eq:varPostLBBNN}, ignore all dependencies in the posterior distributions, which may give suboptimal solutions. To resolve this, \citet{skaaret2023sparsifying} suggested to use dependent latent variables for the mean parameters of the approximate posteriors through  multiplicative normalizing flows \citep{louizos2017multiplicative} within a layer, resulting in the following approximations for a given layer $j$:

\begin{align}
\label{eq:varpost}
q_{\boldsymbol{\theta}}({\mathbf{W}^{(j)}|\boldsymbol{\mathbf{\Gamma}}^{(j)}},\boldsymbol{z}^{(j)})&=\prod_{p,k}\left(\gamma_{k,p}^{(j)}\mathcal{N}(w_{ij};z_k^{(j)} \mu_{k,p}^{(j)},{\sigma_{k,p}^{(j)}}^{2}) + (1-\gamma_{k,p}^{(j)})\delta(w_{k,p}^{(j)})\right); \\
q_{\tilde \alpha_{k,p}^{(j)}}(\gamma_{k,p}^{(j)}) &= \text{Bernoulli}(\gamma_{k,p}^{(j)}; \alpha_{k,p}^{(j)}).
\end{align}
Here, $\boldsymbol z^{(j)}$ follows a distribution $q_{\boldsymbol\phi}(\boldsymbol z^{(j)})$ modeled by normalizing flows \citep{louizos2017multiplicative} and thus introduces flexible spatial dependencies between the parameters through multiplicative neuron-specific random effects. The details on integrating out $\boldsymbol z^{(j)}$ in the KL divergence are further provided in \citet{skaaret2023sparsifying}.

\subsection{Input-skip in Latent Binary Bayesian Neural Networks}

\begin{figure}[htp]
  \centering
    \includegraphics[width=\linewidth]{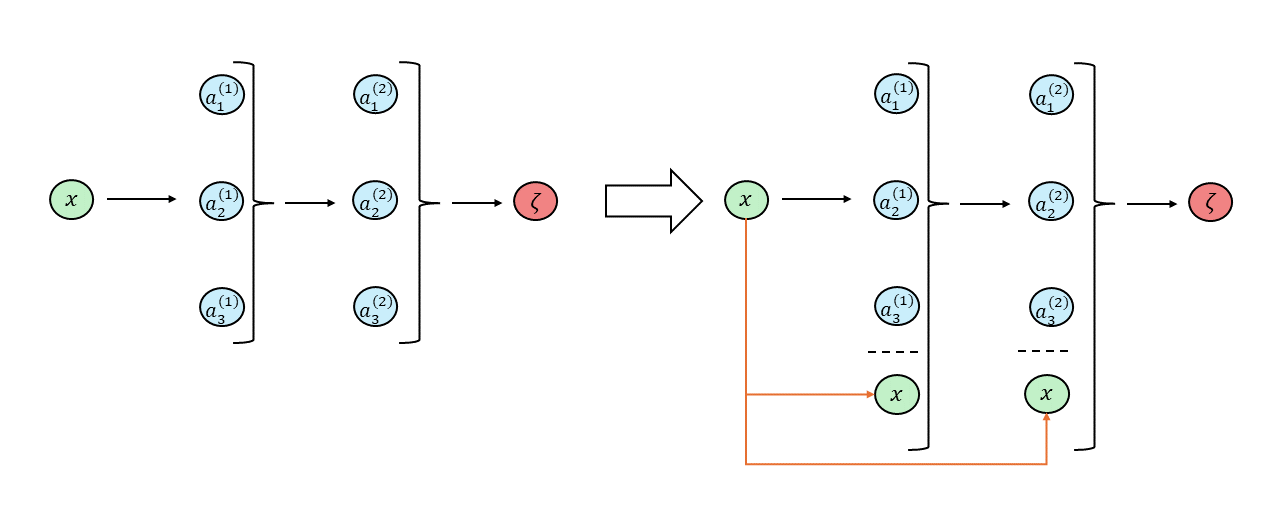} 
  \caption{Concatenation is used to allow the input to skip directly to any given hidden layer.} \label{fig:inputSkipIlluConcatination}
\end{figure}

Input-skip is implemented in LBBNNs by concatenating the input covariates to all hidden layers, as illustrated in Figure \ref{fig:inputSkipIlluConcatination}. While concatenation has been previously utilized in convolutional neural networks such as DenseNets \citep{huang2017densely} and U-nets \citep{ronneberger2015u}, as well as in multilayer perceptron models like DenseNet Regression \citep{jiang2022densely} and $l_1$-regularized nets \citep{lundby2023sparse}, this method has never been associated with an LBBNN model. Input-skip for latent binary Bayesian neural networks, hereafter abbreviated ISLaB, is therefore the first of its kind.

The extension of the LBBNN method to the ISLaB method is as follows:

\begin{equation}
    a_p^{(j)} = o^{(j)} \bigg(w_{0,p}^{(j)} + \sum_{k=1}^{K+v} \gamma_{k,p}^{(j)} a_k^{(j-1)} w_{k,p}^{(j)}\bigg),
\end{equation}
where \(v\) is the number of covariates. The main difference between the ISLaB method and the previously implemented LBBNN is that \(a^{(j-1)}\) will be extended to always include the covariates, resulting in:

\begin{equation*}
    a^{(j-1)} = \left[a_{1}^{(j-1)}, ..., a_{P}^{(j-1)}, \textbf{\textit{x}}^T\right]^T.
\end{equation*}

For instance, the connection between the input layer and the first hidden layer will consist of \(a^{(0)} = \textbf{\textit{x}}\), as there are no other previous hidden nodes, and for the connection between the second-to-last and last layer, \(a^{(J-1)} = \left[a_1^{(J-1)}, ..., a_P^{(J-1)}, \textbf{\textit{x}}^T \right]^T\) will apply.

Another difference lies in the necessity to extend the dimension of the weight matrices and inclusion parameter matrices as the covariates are concatenated, implying that more parameters will need to be considered. This contrasts with residual connections, where the previous information is simply added into succeeding nodes, which does not increase the number of parameters in the model. Inference on the models remains the same as in \citet{hubin2024sparse} and \citet{skaaret2023sparsifying} as independence is assumed between layers.

\subsection{Active Paths}

\begin{figure}[htp]
  \centering
    \includegraphics[width=\linewidth]{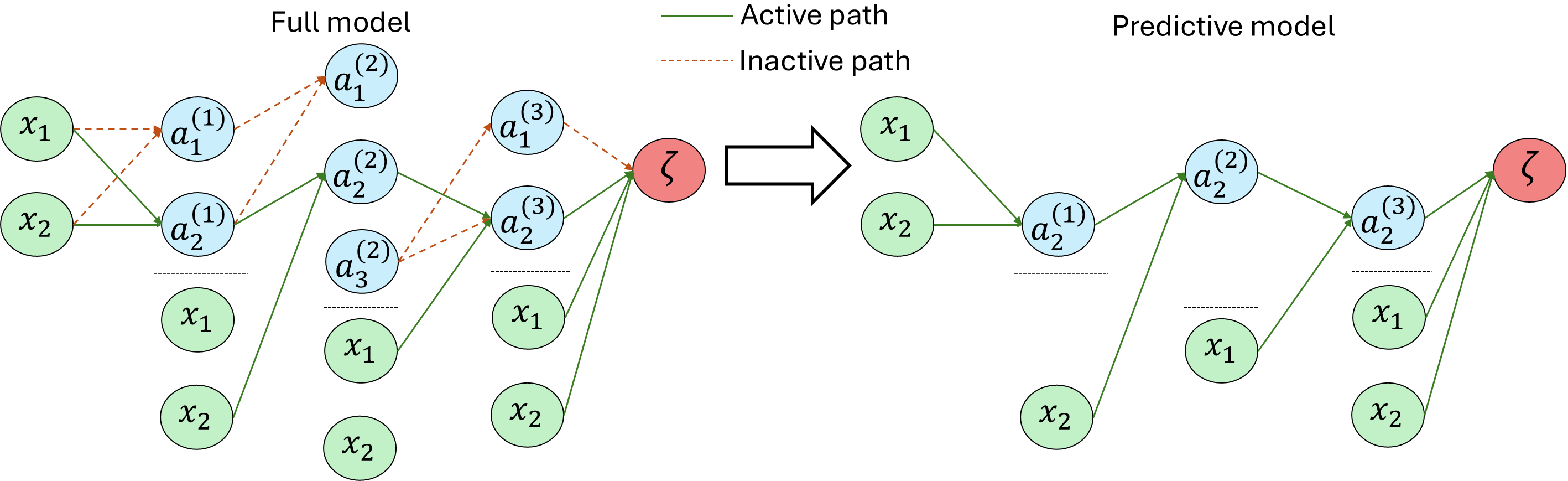} 
  \caption{Examples of inactive paths after removing connections where $\gamma=0$ or $w=0$.}\label{fig:activeAndInactivePaths}
\end{figure}

Model pruning by eliminating redundant weights in neural networks, either through LBBNNs or other methods, like $l_1$-regularized ANNs, can lead to weights being indirectly inactive, which further deem nodes to be inactive (see Figure \ref{fig:activeAndInactivePaths}). This means that no information going into, or out of, these nodes will contribute to the prediction. With this in mind, we define an \textit{active path} in a neural network as follows:

\begin{description}
    \item[\textbf{An active path}]\label{def:activePath}
\textit{ in a neural network is a collection of adjacent weights that connects a covariate directly, or via one or more hidden nodes, to an output node.}
\end{description}

Figure \ref{fig:activeAndInactivePaths} illustrates active and inactive paths, emphasizing that only paths with non-zero weighted connections contribute to the prediction. This definition helps in creating sparser, more interpretable network representations by highlighting only the contributing connections. A set of all active paths in a model is further denoted as $\A\prob$.

It should be noted that a path that only sends information about the bias is not considered an active path. The reason for excluding these paths is twofold: Firstly, our primary interest lies in how the covariates behave, like interactions and the amount of non-linear activations needed to contribute to a prediction; Secondly, a path represented solely by bias information could be summarized in a single bias node, meaning that these connections are mainly hindering an even sparser representation of the network. 

\subsection{Max and Average Depth in ISLaB Models}

The ISLaB model's architecture allows covariates to skip layers, directly influencing any part of the network. This flexibility, while powerful, introduces ambiguity regarding how and where the covariates contribute to the model's predictions. In a traditional ANN without skip connections, contributions are traced from the input layer through successive layers. However, in an ISLaB model, contributions can originate from any layer, making it crucial to understand how the layers are utilized.

To gain insights into the network's behavior, it is useful to measure which layers are most frequently engaged and how many connections are required for inputs to influence the output. This can be analyzed using the concept of \textit{contribution depth}, which we define as follows.

\begin{description}\item[Contribution Depth]\label{def:depth}
\textit{is the path length from a covariate (in some layer) along an active path to the output}
\end{description}

For instance, in a network with four hidden layers, if an active path starts from the input layer, the contribution depth would be five, as five connections are needed to reach the output.

Using this framework, we can measure the contribution depths (denoted as \textbf{depth} in the formula) for any covariate \(\textbf{\textit{x}}_i\), where \(i=1,2,...,v\), in an ISLaB or other relevant neural network model \textit{from a specific layer} \(j\), represented by the following set:

\begin{equation}\label{eq:depth}
    \text{depth}(\textbf{\textit{x}}_i, j) = \bigcup_{p=1}^P \{J-j|w_{k,p}^{(j)} \in \A\prob\}.
\end{equation}

In this equation, \(k\) corresponds to the \(i\)-th covariate in the current layer \(j\). Thus, contributing depth shows the path length from a covariate (in some layer) along an active path to the output. This depth set can be used to derive two key metrics that describe the network's structure: \textit{max depth} and \textit{average depth}.

The max depth metric is defined as the longest active path in the network: 

\begin{equation}\label{eq:maxDepth}
    \text{max depth} = \max \left(\bigcup_{j=1}^J \bigcup_{i=1}^v \text{depth}(\textbf{\textit{x}}_i, j) \right).
\end{equation}

For example, if a covariate contributes from the input layer, the max depth would be \(J-0=J\). Conversely, if the deepest contribution originates from the last hidden layer, the max depth would be \(J-(J-1) = 1\). Thus, max depth reflects the network's most complex path. 

To understand how deep contributions generally are within the network, we use the \textit{average depth} metric:

\begin{equation}\label{eq:avgDepth}
    \text{average depth} = \text{mean} \left(\bigcup_{j=1}^J \bigcup_{i=1}^v \text{depth}(\textbf{\textit{x}}_i, j) \right).
\end{equation}
Average depth provides a general understanding of the depth of contributions across the network. In scenarios where a network lacks active paths, the depth set is empty, resulting in both the max depth and average depth metrics being zero. This situation only occurs when the network outputs a constant prediction for all inputs, indicating no meaningful internal activity. Thus, if the covariate is not contributing to the predictions its max and average depth would be \(0\). 

\subsection{Global Understanding of Predictions}

\begin{figure}[htp]
  \centering
  \begin{tabular}{@{}c@{}}
    \includegraphics[width=1.\linewidth]{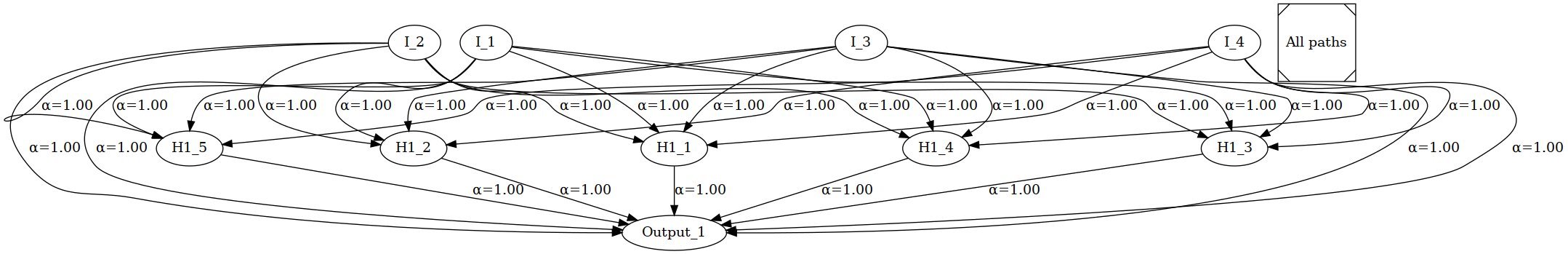} \\[\abovecaptionskip]
    \small (a) Full network before training. 
  \end{tabular}

  \vspace{\floatsep}

  \begin{tabular}{@{}c@{}}
    \includegraphics[height=100pt]{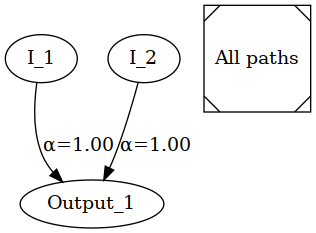} \\[\abovecaptionskip]
    \small (b) Active paths after training.
  \end{tabular}

  \caption{Before (a) and after (b) training an LRT-based ISLaB model for the linear problem $y = x_1 + x_2 + 0 x_3 + 0 x_4$. $\alpha$ is the inclusion probability for the specific weight, where $\alpha=1.00$ indicates that the weight will almost surely be included in the full model. Displayed is the MPM with its active paths. In this plot (and in similar plots to follow), $I_1$ corresponds to the input $x_1$, and similarly for the other inputs, whereas $Output \_1$ corresponds to $y$.}\label{fig:simpleLRTillu}
\end{figure}

Interpreting the predictions of a neural network globally enhances trust in the model. Active paths facilitate a clearer understanding by focusing only on contributing weights. In LBBNNs, this can be achieved by considering active paths in the median probability model (MPM), where a weight $w_{k,p}^{(j)}$ is included if $\alpha_{k,p}^{(j)} > 0.5$. Figure \ref{fig:simpleLRTillu} demonstrates this concept with a simple linear problem\footnote{A similar approach can be employed in $l_1$-regularized ANNs. However, as weights are not associated with an inclusion probability, active paths must be identified through a threshold:
$|w| > \epsilon
$, $\epsilon$ is typically set to a value close to zero.}.

\subsection{Local Explainability}\label{sec:local_explain}

In addition to providing global explanations, active paths allow for scalable local explanations of covariate contribution to a prediction. This is achievable by using activation functions that give linear contributions to the output layer, for instance, the ReLU activation function. Without loss of generality, we shall only consider ReLU activation in the context of local explanations. We do, however, emphasize that other piece-wise linear functions, like leaky ReLU, could give local explanations in a neural network.

The non-linear element of the ReLU function is its capability to eliminate nodes. Elimination of nodes will simplify networks locally as weights going into, and out of, these nodes will be redundant, and the remaining network will only use linear functions to predict. This simpler, sparsified network can be used to find how much each covariate contributes to the prediction. More explicitly, by defining the set of active (non-zero) nodes as $\A\nat$, we can propagate each covariate through a network that uses Equation (\ref{eq:localExplainNode}) to compute the node activations:

\begin{equation}\label{eq:localExplainNode}
    a_p^{(j)} = \indicator\left(a_p^{(j)} \in \A\nat\right)\cdot\left(\sum_{k=1}^K \indicator\left(w_{k,p}^{(j)}\in \A\prob\right) a_k^{(j-1)} w_{k,p}^{(j)} \right),
\end{equation}
Below we shall see that the contributions conditional on a specific observation are linear.

\begin{proposition}\label{prop:localGLM}
Let \(\{(y_i, \mathbf{x}_i)\}_{i=1}^n\) be a dataset with \(\mathbf{x}_i \in \mathbb{R}^v\), and suppose \(y_i \sim \mathcal{F}(\mu_i)\) follows an exponential family distribution with mean \(\mu_i\) and canonical link \(g(\mu_i) = \zeta(\mathbf{x}_i)\). Let \(\zeta: \mathbb{R}^v \to \mathbb{R}\) be the output of a feedforward neural network with \(L\) layers, weights \(\mathbf{W} = \{w_{k,p}^{(j)}, w_{0,p}^{(j)}\}_{j=1}^L\), where \(k \in \{1, \dots, d_{j-1}\}, p \in \{1, \dots, d_j\}\), ReLU activations \(o^{(j)}(z) = \max(0, z)\) for \(j=1,\dots,L-1\), and output activation \(o^{(L)} = g^{-1}(z)\). Assume pre-activation values \(z_p^{(j)} = w_{0,p}^{(j)} + \sum_{k=1}^{d_{j-1}} a_k^{(j-1)} w_{k,p}^{(j)} \neq 0\) almost everywhere. Then, \(\forall \mathbf{x}_i \in \mathbb{R}^v, \exists \varepsilon > 0\) such that \(\forall \mathbf{x} \in B_\varepsilon(\mathbf{x}_i) := \{ \mathbf{x} : \|\mathbf{x} - \mathbf{x}_i\|_2 < \varepsilon \}\), the linear predictor is:
\[
\zeta(\mathbf{x}) = \beta_0 + \sum_{j=1}^v \beta_j x_j,
\]
where \(\beta_0, \beta_j \in \mathbb{R}\) depend on \(\mathbf{W}\) and the ReLU activation pattern at \(\mathbf{x}_i\). Thus, \(\zeta\) locally approximates a GLM. See Appendix \ref{app:proofs} for proof.
\end{proposition}

\begin{remark}\label{rem:localLinearity}
The ReLU network divides the input space into regions where \(\zeta(\mathbf{x}) = \beta_0 + \sum_{j=1}^v \beta_j x_j\).  \(\forall \mathbf{x}_i \in \mathbb{R}^v\),  \(\exists\boldsymbol{\beta} = (\beta_1, \dots, \beta_v)\) and intercept \(\beta_0\) in that region enable interpretable GLM explanations.
\end{remark}

\begin{remark}\label{rem:localVsGlobal}
This result extends to any continuous piecewise-linear activation function in hidden layers.\(\forall \mathbf{x}_i \in \mathbb{R}^v\), \(\exists B_\varepsilon(\mathbf{x}_i) := \{ \mathbf{x} : \|\mathbf{x} - \mathbf{x}_i\|_2 < \varepsilon \}\) where the activation pattern is fixed, resulting in an affine \(\zeta(\mathbf{x}) = \beta_0 + \sum_{j=1}^v \beta_j x_j\).
\end{remark}

\begin{remark}\label{rem:lbnn}
In an LBBNN with weights \(\mathbf{W} = \{w_{k,p}^{(j)}, w_{0,p}^{(j)}\}_{j=1}^L\) and indicators \(\mathbf{\Gamma} = \{\gamma_{k,p}^{(j)}\}\), define the MPM \(\mathbf{\Gamma}^*\) where \(\gamma_{k,p}^{*(j)} = 1\) if \(p(\gamma_{k,p}^{(j)} = 1 \mid D) > 0.5\), else 0. Explanations use the coefficients \(\boldsymbol{\beta}^*, \beta_0^*\) computed from the posterior mean \(\E(\mathbf{W} \mid \mathbf{\Gamma}^*, D)\), satisfying Proposition \ref{prop:localGLM} in a fixed activation region. That means, \(\forall \mathbf{x}_i \in \mathbb{R}^v\),  \(\exists \varepsilon > 0\) such that, \(\forall \mathbf{x}  \in B_\varepsilon(\mathbf{x}_i) := \{ \mathbf{x} : \|\mathbf{x} - \mathbf{x}_i\|_2 < \varepsilon \}\), sampling \(\mathbf{W}\sim \pi(\mathbf{W} \mid \mathbf{\Gamma}^*, D)\) provides a distribution over \(\boldsymbol{\beta}, \beta_0\), yielding credible intervals for interpretability. See Appendix \ref{app:proofs} for proof.
\end{remark}

\begin{corollary}\label{coro:GradientExplain}
Let \(\mathbf{x}_i\) and \(\zeta(\mathbf{x}_i)\) be as in Proposition \ref{prop:localGLM} with continuous piecewise-linear activations. The local linear coefficients \(\boldsymbol{\beta} = (\beta_1, \dots, \beta_v)\) are:
\[
\boldsymbol{\beta} = \nabla_{\mathbf{x}} \zeta(\mathbf{x}) \big|_{\mathbf{x} = \mathbf{x}_i}.
\]
This directly provides explanation coefficients. See Appendix \ref{app:proofs} for proof.
\end{corollary}

Although gradient-based methods have previously been used to give local explanations of predictions \citep{shrikumar2016not,shrikumar2017learning}, similar to Corollary \ref{coro:GradientExplain}, to the best of our knowledge it has never been applied in the context of Proposition \ref{prop:localGLM}. Specifically, the explainable slope coefficients are equivalent to those used to produce the linear prediction for a given input. These can be obtained directly by computing the gradients with respect to the covariates, as long as piecewise linear activation functions are utilized. This is also applicable to LBBNNs.

\begin{corollary}\label{coro:LBBNNGradient}
In an LBBNN with MPM \(\mathbf{\Gamma}^*\) where \(\gamma_{k,p}^{*(j)} = 1\) if \(p(\gamma_{k,p}^{(j)} = 1 \mid D) > 0.5\), else 0, let \(\mathbf{x}_i\) and \(\zeta(\mathbf{x}_i)\) be as in Proposition \ref{prop:localGLM}. For each sample \(\mathbf{W} \sim \pi(\mathbf{W} \mid \mathbf{\Gamma}^*, D)\), \(\exists \varepsilon > 0\) such that, \(\forall \mathbf{x} \in B_\varepsilon(\mathbf{x}_i) := \{ \mathbf{x} : \|\mathbf{x} - \mathbf{x}_i\|_2 < \varepsilon \}\), the local linear coefficients are:
\[
\boldsymbol{\beta} = \nabla_{\mathbf{x}} \zeta(\mathbf{x}) \big|_{\mathbf{x} = \mathbf{x}_i}.
\]
The explanation coefficients \(\boldsymbol{\beta}^*, \beta_0^*\) are computed using \(\E(\mathbf{W} \mid \mathbf{\Gamma}^*, D)\). Sampling from \(\pi(\mathbf{W} \mid \mathbf{\Gamma}^*, D)\) provides a distribution over \(\boldsymbol{\beta}\), with credible intervals for \(\boldsymbol{\beta}\). See Appendix \ref{app:proofs} for proof.
\end{corollary}

\begin{figure}[htp]
  \centering
    \includegraphics[width=0.7\linewidth]{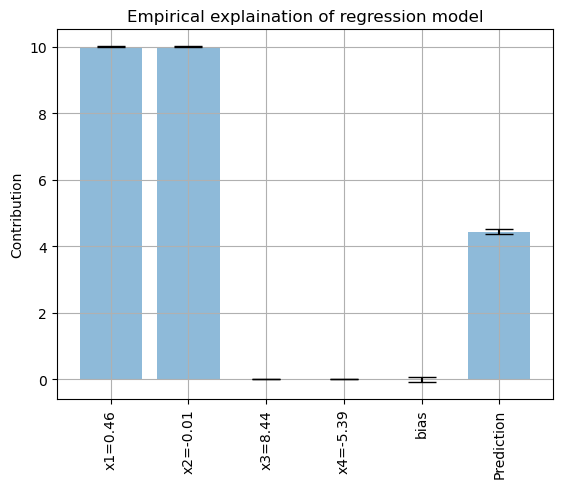} 
  \caption{Local explanation for a prediction made on an LBBNN model trained on the linear problem $y = 10 x_1 + 10 x_2 + 0 x_3 + 0 x_4$.}\label{fig:exampleLocalExplainGraph}
\end{figure}

In the LBBNN setting, it will also be feasible to compute credibility intervals for covariate contributions, providing insights into the magnitude and certainty of each covariate's impact. Figure \ref{fig:exampleLocalExplainGraph} exemplifies this local explanation for a prediction made from an LBBNN model. 

Furthermore, Proposition \ref{prop:localGLM} enables us to establish theoretical connections to two 
widely-used explanation methods: LIME \citep{ribeiro2016should} and Integrated Gradients 
(IG) \citep{sundararajan2017axiomatic}. However, our approach is fundamentally different: we 
provide \textit{point-wise, path-independent} exact explanations derived directly from the model, 
whereas LIME, by contrast, explains predictions 
through post-hoc surrogate model fitting in a local neighborhood, while IG aggregates 
gradients along a path from a baseline to the test input, inherently making it path-dependent. 
This fundamental distinction means our approach and these post-hoc methods serve complementary 
but different explanatory goals. For LBBNNs, our method further provides credible 
intervals for explanation coefficients through posterior sampling, a feature not naturally 
provided by LIME or IG.

\begin{corollary}\label{cor:limeFixed}
For a neural network as in Proposition \ref{prop:localGLM}, \(\exists \varepsilon > 0\) such that, in \(B_\varepsilon(\mathbf{x}_i) := \{ \mathbf{x} : \|\mathbf{x} - \mathbf{x}_i\|_2 < \varepsilon \}\), the LIME coefficients \(\hat{\boldsymbol{\beta}}, \hat{\beta}_0\), obtained via MLE for a GLM using \(n \to \infty\) samples \(\{\mathbf{x}_k, \tilde{y}_k\}_{k=1}^n\) from a distribution \(\pi\) in \(B_\varepsilon(\mathbf{x}_i)\) with \(\tilde{y}_k = \zeta(\mathbf{x}_k)\), satisfy:
\[
\hat{\boldsymbol{\beta}} \overset{p}{\to} \boldsymbol{\beta}, \quad \hat{\beta}_0 \overset{p}{\to} \beta_0,
\]
where \(\boldsymbol{\beta}, \beta_0\) are from Proposition \ref{prop:localGLM}. This shows LIME recovers the local GLM. See Appendix \ref{app:proofs} for proof.
\end{corollary}

\begin{corollary}\label{cor:limeLBBNN}
In an LBBNN with MPM \(\mathbf{\Gamma}^*\) where \(\gamma_{k,p}^{*(j)} = 1\) if \(p(\gamma_{k,p}^{(j)} = 1 \mid D) > 0.5\), else 0, let \(\mathbf{x}_i\) and \(\zeta(\mathbf{x}_i)\) be as in Proposition \ref{prop:localGLM}. The explanation coefficients \(\boldsymbol{\beta}^*, \beta_0^*\) are computed using \(\E(\mathbf{W} \mid \mathbf{\Gamma}^*, D)\). For each sample \(\mathbf{W} \sim \pi(\mathbf{W} \mid \mathbf{\Gamma}^*, D)\), \(\exists \varepsilon > 0\) such that, \(\forall  \mathbf{x} \in B_\varepsilon(\mathbf{x}_i) := \{ \mathbf{x} : \|\mathbf{x} - \mathbf{x}_i\|_2 < \varepsilon \}\), the LIME coefficients \(\hat{\boldsymbol{\beta}}, \hat{\beta}_0\) from a GLM with \(n \to \infty\) samples \(\{\mathbf{x}_k, \tilde{y}_k\}_{k=1}^n\) from \(\pi\) in \(B_\varepsilon(\mathbf{x}_i)\) with \(\tilde{y}_k = \zeta(\mathbf{x}_k)\), satisfy:
\[
\hat{\boldsymbol{\beta}} \overset{p}{\to} \boldsymbol{\beta}, \quad \hat{\beta}_0 \overset{p}{\to} \beta_0.
\]
The explanation uses \(\boldsymbol{\beta}^*, \beta_0^*\), with credible intervals for \(\boldsymbol{\beta}, \beta_0\) from sampling. See Appendix \ref{app:proofs} for proof.
\end{corollary}

\begin{remark}\label{rem:limeVsProp}
\(\forall \mathbf{x}_i \in \mathbb{R}^v\), unlike LIME, which fits a GLM in \(B_\varepsilon(\mathbf{x}_i) := \{ \mathbf{x} : \|\mathbf{x} - \mathbf{x}_i\|_2 < \varepsilon \}\), Proposition \ref{prop:localGLM} computes \(\boldsymbol{\beta}, \beta_0\) directly from \(\mathbf{W}\) and \(\mathbf{\Gamma}^*\). In an LBBNN, the explanation uses \(\boldsymbol{\beta}^*, \beta_0^*\) from \(\E(\mathbf{W} \mid \mathbf{\Gamma}^*, D)\), avoiding retraining. Sampling from \(\pi(\mathbf{W} \mid \mathbf{\Gamma}^*, D)\) provides a distribution over \(\boldsymbol{\beta}, \beta_0\), yielding credible intervals for interpretability.
\end{remark}

The following corollary examine IG under a restrictive assumption: that the network 
operates in a single piecewise-linear region along the entire integration path. We show that under this idealized scenario, IG can be related to our point-wise coefficients.

\begin{corollary}\label{coro:IGExplain}
Let \(\mathbf{x}_i\) and \(\zeta(\mathbf{x}_i)\) be as in Proposition \ref{prop:localGLM} with continuous piecewise-linear activations. For a baseline \(\mathbf{x}' \in \mathbb{R}^v\), define the Integrated Gradients attribution for feature \(j\) as:
\[
\text{IG}_j(\mathbf{x}_i) = (x_{i,j} - x'_j) \int_0^1 \frac{\partial \zeta}{\partial x_j} (\mathbf{x}' + \alpha (\mathbf{x}_i - \mathbf{x}')) \, d\alpha.
\]
Assuming \(\mathbf{x}' + \alpha (\mathbf{x}_i - \mathbf{x}') \in B_\varepsilon(\mathbf{x}_i) := \{ \mathbf{x} : \|\mathbf{x} - \mathbf{x}_i\|_2 < \varepsilon \}\) for \(\alpha \in [0,1]\), the attributions are:
\[
\text{IG}_j(\mathbf{x}_i) = (x_{i,j} - x'_j) \beta_j,
\]
where \(\boldsymbol{\beta} = (\beta_1, \dots, \beta_v) = \nabla_{\mathbf{x}} \zeta(\mathbf{x}) \big|_{\mathbf{x} = \mathbf{x}_i}\). This provides explanation coefficients scaled by the input difference from the baseline. See Appendix \ref{app:proofs} for proof.
\end{corollary}

\begin{corollary}\label{coro:LBBNNIG}
In an LBBNN with MPM \(\mathbf{\Gamma}^*\), let \(\mathbf{x}_i\) and \(\zeta(\mathbf{x}_i)\) be as in Proposition \ref{prop:localGLM}. For each sample \(\mathbf{W} \sim \pi(\mathbf{W} \mid \mathbf{\Gamma}^*, D)\), \(\exists \varepsilon > 0\) such that, \(\forall \mathbf{x} \in B_\varepsilon(\mathbf{x}_i) := \{ \mathbf{x} : \|\mathbf{x} - \mathbf{x}_i\|_2 < \varepsilon \}\), define the Integrated Gradients attribution for feature \(j\) as:
\[
\text{IG}_j(\mathbf{x}_i) = (x_{i,j} - x'_j) \int_0^1 \frac{\partial \zeta}{\partial x_j} (\mathbf{x}' + \alpha (\mathbf{x}_i - \mathbf{x}')) \, d\alpha.
\]
Assuming \(\mathbf{x}' + \alpha (\mathbf{x}_i - \mathbf{x}') \in B_\varepsilon(\mathbf{x}_i)\) for \(\alpha \in [0,1]\), the attributions are:
\[
\text{IG}_j(\mathbf{x}_i) = (x_{i,j} - x'_j) \beta_j,
\]
where \(\boldsymbol{\beta} = \nabla_{\mathbf{x}} \zeta(\mathbf{x}) \big|_{\mathbf{x} = \mathbf{x}_i}\). The explanation coefficients \(\boldsymbol{\beta}^*, \beta_0^*\) are computed using \(\E(\mathbf{W} \mid \mathbf{\Gamma}^*, D)\), yielding \(\text{IG}_j(\mathbf{x}_i) = (x_{i,j} - x'_j) \beta_j^*\). Sampling from \(\pi(\mathbf{W} \mid \mathbf{\Gamma}^*, D)\) provides a distribution over \(\text{IG}_j(\mathbf{x}_i)\), with credible intervals. See Appendix \ref{app:proofs} for proof.
\end{corollary}

\section{Experiments}

In this section, we present experiments that demonstrate our methods and compare them to related approaches. We first describe how the global-explanation experiments are set up, followed by the results and comparisons to existing methods. Finally, we illustrate how local explanations can be retrieved.

\subsection{Experimental Set-up} 

\begin{figure}
    \centering
    \includegraphics[width=0.65\textwidth]{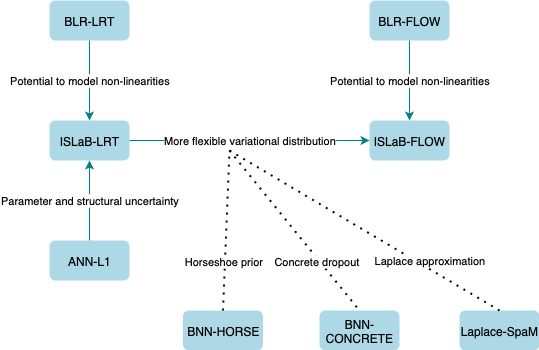}
    \caption{Illustration of the relations between the different methods considered for the experiments. With the solid lines, exactly one change in design is present for each edge allowing for a fully transparent ablation with clear treatments (ISLaB modifications with BCNN and ViT  are not included in the graph for better readability, since but are obviously directly linked to the respective ISLaB methods). The dotted lines represent the additional baselines where we have no such clear relation to the other methods.}\label{fig:methods_relations}
\end{figure}

In the design of experiments, we primarily want to have clear ablation rather than comparing fundamentally different entities. Thus, first and foremost, we benchmark our proposed ISLaB method against its closest related baselines methods. When we use the mean-field posterior, we denote it ISLaB-LRT, and ISLaB-FLOW when we in addition use normalizing flows. For the baselines, we use a simple Bayesian linear model with covariate selection, corresponding to a special case of LBBNN with only one neuron. Here, we again report results with the mean-field posterior (denoted BLR-LRT) and with normalizing flows (denoted BLR-FLOW). Lastly, we compare to a frequentist neural network, with L1 regularization on the weights, and input-skip (denoted IS-ANN-L1), representing a frequentist counterpart to our ISLaB method. For a graphical illustration of how the methods are related, see Figure \ref{fig:methods_relations}. Within these set of models, direct treatment effects are introduced on each edge. Furthermore, Bayesian convolutional networks (BCNNs) \citep{skaaret2023sparsifying} with ISLaB as the fully connected part is included for image classification problems. BCNNs with ISLaB is implemented with both LRT and normalizing flows, denoted BCNN-ISLaB-LRT and BCNN-ISLaB-FLOW, respectfully. We also extended LBBNN to be combined with  a visual transformer from \citet{dosovitskiy2020image} to demonstrate how one can boost the performance on CIFAR-10 (these results are reported in Appendix \ref{appendix:cifar10exp}). Note that the depths metrics and active paths are not defined for either convolutional or transformer layers, hence will not be reported for them.

Additionally, to perform sanity checks on how reasonable the performance is with respect to strong baselines, we compare against other popular BNN sparsification methods that are more loosely related to our approach. We consider, concrete dropout \citep{gal2017concrete}, denoted BNN-CONCRETE, which obtains sparse networks through learned dropout probabilities. Moreover, we have BNNs with horseshoe priors \citep{louizos2017bayesian, carvalho2009handling}, denoted BNN-HORSE. Finally, we consider the recent work of \citet{dhahri2024shaving}, which obtains sparse networks by using a Laplace approximation of the marginal likelihood, denoted Laplace-SpaM. We use an unstructured pruning approach, where we prune the weights based on the optimal posterior damage criterion proposed in \citet{dhahri2024shaving}. We also follow their suggestion and do not prune the last layer, for better performance. This is in contrast to our approach where all layers are treated equally. For this part (comparing to loosely related methods) of the experimental design, direct ablation is not possible and we cannot explain which particular design changes lead to a better or worse performance.  

To measure performance we consider a variety of metrics that allow to evaluate methods from different perspectives. For the classification problems, we report accuracy, denoted  ``ACC full'' when we are using all the weights, and ``ACC sparse'' when we are using the sparsified networks. For the ISLaB and BLR methods, the sparse networks are obtained using the median probability model (MPM) of \citet{barbieri2004optimal}, i.e. removing the weights with a corresponding posterior inclusion probability less than 0.5. For the frequentist approach, we obtain the sparse results by only using the parameters that are left after removing the ones falling below a pre-determined threshold of 0.005 on weights, for all experiments. For the regression problem, we report the Pearson correlation, and root mean squared error, denoted ``CORR'' and ``RMSE'' respectively. Furthermore, we separate the dataset into two different parts when using the ISLaB models. We use a training set to learn the appropriate model parameters for the specific problem at hand, and a test set which is used to report the model performance post-training.

Additionally, to check the calibration of the predictions, we consider two metrics. First, the expected calibration error (ECE) \citep{guo2017calibration}, which estimates how well the model predictions match the true probabilities, and secondly the negative test log-likelihood. For the regression problem, we report the pinball loss \citep{chung2021beyond} instead of ECE. 

Finally, we are interested in checking what kind of structure our networks have, and how sparse they are. For this, we measure the number of weights in active paths after sparsifying the networks, denoted ``Used weights''. We also report the average depth (``Avg depth'') and max depth (``Max depth'') of the networks.

\begin{figure}
    \centering
    \includegraphics[width=1.0\textwidth]{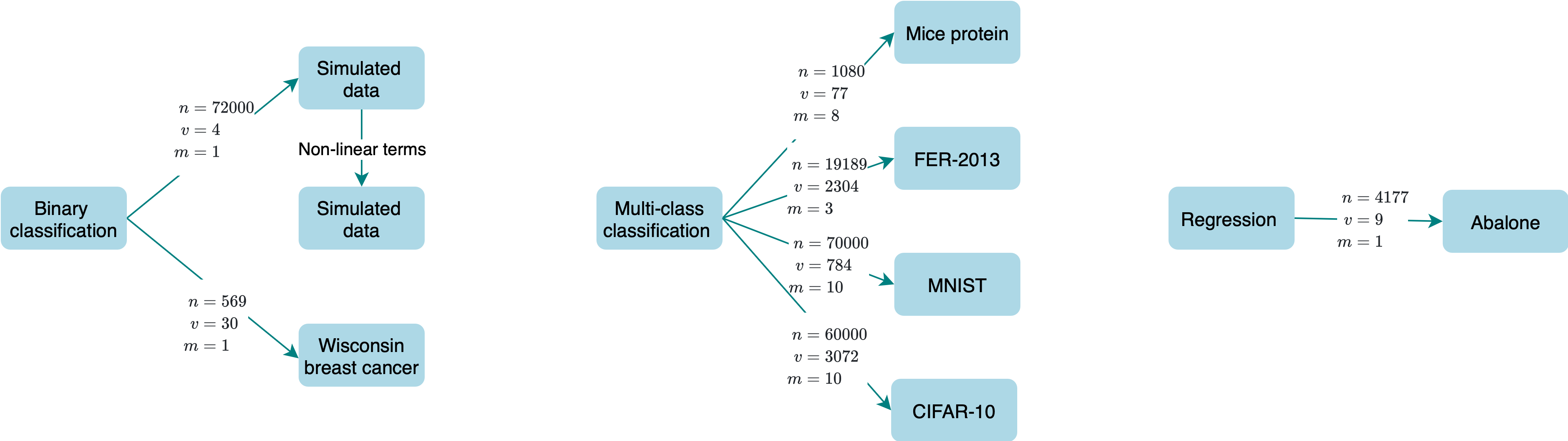}
    \caption{Datasets used in the experiments, where $n$ denotes the number of samples, $v$ the number of covariates, and $m$ the number of output neurons.}\label{fig:datasets}
\end{figure}

To assess and compare our methods, we use both synthetic and real-world datasets. For an overview, we refer to Figure \ref{fig:datasets}. To account for random variability, for all experiments, we run the different methods ten times on different seeds and then report minimum, median, and maximum results. All hyperparameters and settings used for these experiments can be found in Table \ref{tab:paramMat}, while initialization for inclusion probabilities can be found in Table \ref{tab:initInclusionProb}, both found in Appendix \ref{appendix:A}. All networks will be trained using the logistic function, also known as the sigmoid function, as the activation function for all hidden nodes. 

\subsection{Global Explanations and Model Performance}
In the first section, we start with a synthetic dataset to illustrate how our method performs with different levels of correlations, first with linear, and then non-linear data. We also check how well our method does at picking out the correct data-generating covariates. Following this, we compare results on real-world datasets. Additionally, we explore global explanations on these datasets i.e. checking which variables are influencing predictions in general. All experiments can be found and reproduced from \url{https://github.com/eirihoyh/ISLaB-LBBNN/tree/JAIR}.

\subsubsection{Simulated Linear Data}\label{sec:SimulatedLinearData}
At first, we explore two toy examples to demonstrate how the models perform when the underlying structure of the data is known. In the first example, $\textbf{X} = [\textbf{\textit{x}}_1, \textbf{\textit{x}}_2, \textbf{\textit{x}}_3, \textbf{\textit{x}}_4]$, will be sampled independently from uniform distributions, spanning from $-10$ to $10$.  Additionally, $\textbf{\textit{x}}_3$ will be made gradually more dependent on $\textbf{\textit{x}}_1$ through the following equation:

\begin{equation}\label{eq:dependenceFunc}
    \textbf{\textit{x}}_3 \leftarrow \varrho \cdot \textbf{\textit{x}}_1 + (1-\varrho)\cdot \textbf{\textit{x}}_3,
\end{equation}
where the dependence is varied over the set of values $\varrho=\{0.0, 0.1, 0.5, 0.9\}$. However, only $\textbf{\textit{x}}_1$ and $\textbf{\textit{x}}_2$ will be used as data generative covariates, which means that both $\textbf{\textit{x}}_3$ and $\textbf{\textit{x}}_4$ are redundant covariates and should be ignored when making decisions. We thus have:
\begin{equation}\label{eq:LinearProb}
    \eta(\textbf{X}) = \textit{w}_0 + \textit{w}_1 \textbf{\textit{x}}_1 + \textit{w}_2 \textbf{\textit{x}}_2 + \boldsymbol{\epsilon},
\end{equation}
with $w_0 = 100$, $w_1=w_2 = 1$ and $\epsilon_i \sim \N(0,0.01^2)$.

The problem will be transformed into a classification problem by forcing the output to be either $0$ or $1$ using the following equation:

\begin{equation}\label{eq:LabelSimData}
   \textbf{y} = 
    \begin{cases}
        0, \quad \eta(\textbf{X}) < \text{median}(\eta(\textbf{X})) \\
        1, \quad \eta(\textbf{X}) \geq \text{median}(\eta(\textbf{X})).
    \end{cases}
\end{equation}
 We use 64000 samples for training, and 8000 for testing the trained model. For the neural network models, 4 hidden layers, each having 20 hidden nodes, will be initialized, resulting in 1544 weights initialized. The results are in Tables \ref{tab:linAcc}, \ref{tab:lincalibration}, and \ref{tab:linInclution}. 
 \begin{table}[htp]
    \caption{Results from problem described in Equation (\ref{eq:LinearProb}). The neural networks have 1544 weights, while the linear methods have 4. We report median (min, max) for all measures. Bold indicates the best score, while italic is the second best.}
    \resizebox{\textwidth}{!}{
    \centering
    \begin{tabular}{p{0.18\linewidth}p{0.20\linewidth}p{0.20\linewidth}p{0.15\linewidth}p{0.15\linewidth}p{0.1\linewidth}}
     \hline
     \multicolumn{6}{c}{Accuracy, used weights and depth metrics for the linear problem: $\textbf{\textit{y}} = \textbf{\textit{x}}_1 + \textbf{\textit{x}}_2 + 100$} \\
     \hline
     Model; $\varrho$ & ACC full & ACC sparse & Used weights & Avg depth & Max depth \\
     \hline
     ISLaB-LRT; 0.0  & \textbf{99.9} (\textbf{99.9}, \textbf{100})\%  & \textbf{99.9} (\textbf{99.9}, \textbf{100})\%  & 2 (2, 2) & 1.0 (1.0, 1.0) & 1 (1, 1)\\ 
     ISLaB-LRT; 0.1  & \textbf{99.9} (\textbf{99.9}, \textit{99.9})\% & \textbf{99.9} (\textbf{99.9}, \textbf{100})\%  & 2 (2, 2) & 1.0 (1.0, 1.0) & 1 (1, 1)\\
     ISLaB-LRT; 0.5  & \textbf{99.9} (\textbf{99.9}, \textbf{100})\%  & \textbf{99.9} (\textbf{99.9}, \textbf{100})\%  & 2 (2, 2) & 1.0 (1.0, 1.0) & 1 (1, 1)\\
     ISLaB-LRT; 0.9  & \textbf{99.9} (\textbf{99.9}, \textit{99.9})\% & \textbf{99.9} (\textbf{99.9}, \textbf{100})\%  & 3 (3, 3) & 1.0 (1.0, 1.0) & 1 (1, 1)\\
     \hline
     ISLaB-FLOW; 0.0  & \textbf{99.9} (\textit{99.8}, \textbf{100})\% & \textbf{99.9} (\textit{99.8}, \textit{99.9})\% & 2 (2, 2)  & 1.0 (1.0, 1.0)&1 (1, 1)\\
     ISLaB-FLOW; 0.1  & \textbf{99.9} (\textbf{99.9}, \textbf{100})\% & \textbf{99.9} (\textit{99.8}, \textbf{100})\%  & 2 (2, 2) & 1.0 (1.0, 1.0)&1 (1, 1) \\
     ISLaB-FLOW; 0.5  & \textbf{99.9} (\textit{99.8}, \textit{99.9})\% & \textbf{99.9} (99.7, \textbf{100})\% & 3 (3, 3) & 1.0 (1.0, 1.0)&1 (1, 1) \\
     ISLaB-FLOW; 0.9  & \textbf{99.9} (\textit{99.8}, \textit{99.9})\% & \textbf{99.9} (\textit{99.8}, \textit{99.9})\% & 3 (3, 3) & 1.0 (1.0, 1.0)&1 (1, 1)\\
     \hline
     BLR-LRT; 0.0  & \textbf{99.9} (\textbf{99.9}, \textit{99.9})\% & \textbf{99.9} (\textbf{99.9}, \textit{99.9})\% & 2 (2, 2)&1 (-,-)&1 (-,-)  \\
     BLR-LRT; 0.1  & \textbf{99.9} (\textbf{99.9}, \textit{99.9})\% & \textbf{99.9} (\textbf{99.9}, \textit{99.9})\% & 2 (2, 2)&1 (-,-)&1 (-,-)  \\
     BLR-LRT; 0.5  & \textbf{99.9} (\textbf{99.9}, \textit{99.9})\% & \textbf{99.9} (\textbf{99.9}, \textit{99.9}\% & 2 (2, 2)&1 (-,-)&1 (-,-) \\
     BLR-LRT; 0.9  & \textbf{99.9} (\textbf{99.9}, \textit{99.9})\% & \textbf{99.9} (\textbf{99.9}, \textit{99.9})\% &3 (3, 3) &1 (-,-)&1 (-,-) \\
     \hline
     BLR-FLOW; 0.0  & \textbf{99.9} (99.7, \textit{99.9})\% & \textbf{99.9} (99.7, \textit{99.9})\% &2 (2, 2)&1 (-,-)&1 (-,-) \\
     BLR-FLOW; 0.1  & \textbf{99.9} (99.7, \textit{99.9})\% & \textbf{99.9} (99.7, \textbf{100})\% &2 (2, 2)&1 (-,-) &1 (-,-) \\
     BLR-FLOW; 0.5  & \textbf{99.9} (99.7, \textit{99.9})\% & \textbf{99.9} (99.7, \textit{99.9})\% & 2 (2, 2)&1 (-,-)&1 (-,-)  \\
     BLR-FLOW; 0.9  & \textbf{99.9} (\textit{99.8}, \textit{99.9})\% & \textbf{99.9} (\textit{99.8}, \textbf{100})\% & 3 (2, 3)&1 (-,-)&1 (-,-)  \\
     \hline
     IS-ANN-L1; 0.0  & \textbf{99.9} (\textbf{99.9}, \textit{99.9})\% & \textbf{99.9} (\textbf{99.9}, \textbf{100})\% &8 (5, 8)& 1.7 (1.5, 1.7)& 2 (2, 2)\\
     IS-ANN-L1; 0.1  & \textbf{99.9} (\textbf{99.9}, \textit{99.9})\% & \textbf{99.9} (\textbf{99.9}, \textbf{100})\% &8 (8, 8) & 1.7 (1.7, 1.7) & 2 (2, 2)\\
     IS-ANN-L1; 0.5  & \textbf{99.9} (\textbf{99.9}, \textit{99.9})\% & \textbf{99.9} (\textbf{99.9}, \textbf{100})\% & 8 (8, 8) & 1.7 (1.7, 1.7) & 2 (2, 2)\\
     IS-ANN-L1; 0.9  & \textbf{99.9} (\textbf{99.9}, \textit{99.9})\% & \textbf{99.9} (\textbf{99.9}, \textit{99.9})\% & 9 (9, 10) & 1.7 (1.7, 1.8)& 2 (2, 2)\\
     \hline
    \end{tabular}
}
\label{tab:linAcc}
\end{table}

 \begin{table}[htp]
    \caption{Expected calibration error and NLL on simulated linear data (lower is better).}
    \resizebox{\textwidth}{!}{
    \centering
    \begin{tabular}{p{0.20\linewidth}p{0.20\linewidth}p{0.20\linewidth}p{0.20\linewidth}p{0.20\linewidth}}
     \hline
     \multicolumn{5}{c}{Calibration and negative log-likelihood for the linear problem: $\textbf{\textit{y}} = \textbf{\textit{x}}_1 + \textbf{\textit{x}}_2 + 100$} \\
     \hline
     Model; $\varrho$ & ECE & ECE sparse &  NLL & NLL sparse  \\
     \hline
     ISLaB-LRT; 0.0  & \textit{0.004} (\textit{0.003}, 0.004) & \textit{0.004} (\textit{0.003}, 0.004) & \textit{0.005} (\textit{0.005}, \textit{0.006}) & \textit{0.005} (\textit{0.005}, \textit{0.006}) \\ 
     ISLaB-LRT; 0.1  & \textit{0.004} (\textit{0.003}, 0.004) & \textit{0.004} (\textit{0.003}, 0.004) & \textit{0.005} (\textit{0.005}, \textit{0.006}) & \textit{0.005} (\textit{0.005}, \textit{0.006}) \\
     ISLaB-LRT; 0.5  & \textit{0.004} (\textit{0.003}, 0.004) & \textit{0.004} (\textit{0.003}, 0.004) & \textit{0.005} (\textit{0.005}, \textit{0.006}) & \textit{0.005} (\textit{0.005}, \textit{0.006}) \\
     ISLaB-LRT; 0.9  & \textit{0.004} (\textit{0.003}, 0.004) & \textit{0.004} (\textit{0.003}, 0.004) & \textit{0.005} (\textit{0.005}, \textit{0.006}) & \textit{0.005} (\textit{0.005}, \textit{0.006}) \\
     \hline
     ISLaB-FLOW; 0.0  & 0.007 (0.006, 0.008) & 0.007 (0.006, 0.008) & 0.010 (0.010, 0.011) & 0.010 (0.010, 0.011) \\ 
     ISLaB-FLOW; 0.1  & 0.008 (0.007, 0.008) & 0.007 (0.007, 0.008) & 0.010 (0.010, 0.011) & 0.010 (0.010, 0.011) \\
     ISLaB-FLOW; 0.5  & 0.007 (0.006, 0.008) & 0.007 (0.006, 0.008) & 0.009 (0.009, 0.010) & 0.009 (0.009, 0.010) \\
     ISLaB-FLOW; 0.9  & 0.006 (0.005, 0.006) & 0.006 (0.005, 0.006) & 0.009 (0.008, 0.009) & 0.009 (0.008, 0.009) \\
     \hline
     BLR-LRT; 0.0  & 0.007 (0.007, 0.007) & 0.007 (0.006, 0.007) & 0.009 (0.009, 0.009)  & 0.009 (0.009, 0.009)\\ 
     BLR-LRT; 0.1  & 0.007 (0.007, 0.007) & 0.007 (0.006, 0.007) & 0.009 (0.009, 0.009) & 0.009 (0.009, 0.009)\\
     BLR-LRT; 0.5  & 0.007 (0.006, 0.007) & 0.007 (0.006, 0.007) &0.009 (0.009, 0.009) & 0.009 (0.009, 0.009)\\
     BLR-LRT; 0.9  & 0.007 (0.006, 0.007) & 0.007 (0.006, 0.007) & 0.009 (0.009, 0.009) & 0.009 (0.009, 0.009)\\
     \hline
     BLR-FLOW; 0.0  & \textbf{0.002} (\textbf{0.001}, \textbf{0.002}) & \textbf{0.002} (\textbf{0.001}, \textit{0.003}) & \textbf{0.004} (\textbf{0.003}, \textit{0.006}) & \textbf{0.004} (\textbf{0.003}, \textit{0.006})\\ 
     BLR-FLOW; 0.1  & \textbf{0.002} (\textbf{0.001}, \textbf{0.002}) & \textbf{0.002} (\textbf{0.001}, \textbf{0.002}) & \textbf{0.004} (\textbf{0.003}, \textbf{0.005}) & \textbf{0.004} (\textbf{0.003}, \textbf{0.005})\\
     BLR-FLOW; 0.5  & \textbf{0.002} (\textbf{0.001}, \textbf{0.002}) & \textbf{0.002} (\textbf{0.001}, \textbf{0.002}) &\textbf{0.004} (\textbf{0.003}, \textit{0.006}) & \textbf{0.004} (\textbf{0.003}, \textit{0.006})\\
     BLR-FLOW; 0.9  & \textbf{0.002} (\textbf{0.001}, \textit{0.003}) & \textbf{0.002} (\textbf{0.001}, \textit{0.003}) & \textbf{0.004} (\textbf{0.003}, \textbf{0.005}) & \textbf{0.004} (\textbf{0.003}, \textbf{0.005})\\
     \hline
     IS-ANN-L1; 0.0  & 0.007 (0.007, 0.007) & 0.007 (0.007, 0.007) &0.009 (0.009, 0.009)& 0.009 (0.009, 0.009) \\
     IS-ANN-L1; 0.1  & 0.007 (0.007, 0.007) & 0.007 (0.007, 0.007) &0.009 (0.009, 0.010) & 0.009 (0.009, 0.009)  \\
     IS-ANN-L1; 0.5  & 0.007 (0.007, 0.007) & 0.007 (0.007, 0.008) & 0.009 (0.009, 0.010) & 0.009 (0.009, 0.010)\\
     IS-ANN-L1; 0.9  & 0.007 (0.007, 0.007) & 0.007 (0.007, 0.007) & 0.009 (0.009, 0.009) & 0.009 (0.009, 0.009) \\
     \hline
    \end{tabular}
}
\label{tab:lincalibration}
\end{table}

\begin{figure}[htp]
\centering
  \includegraphics[width=.35\linewidth]{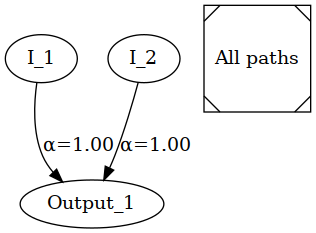}
\caption{Typical learned structures of Equation (\ref{eq:LinearProb}) when using $\varrho=0.0$ for an FLOW-based model. $\alpha$ indicates the inclusion probability and displayed is the median probability model with its active paths. Such graphs allow for transparent global explanations based on our models showing significant paths for each selected covariate (both in terms of MPM).}
\label{fig:linExamples}
\end{figure}

\begin{table}[htp]
    \caption{Same methods as used in Table \ref{tab:linAcc}. Here, the inclusion rates are reported for each covariate that is fed into the model.}
    \resizebox{\textwidth}{!}{
    \centering
    \begin{tabular}{p{0.2\linewidth}p{0.2\linewidth}p{0.2\linewidth}p{0.2\linewidth}p{0.15\linewidth}}
     \hline
     \multicolumn{5}{c}{Inclusion rates for the linear problem: $\textbf{\textit{y}} = \textbf{\textit{x}}_1 + \textbf{\textit{x}}_2 + 100$} \\
     \hline
     Model; $\varrho$ & Inclusion rate $\textbf{\textit{x}}_1$& Inclusion rate $\textbf{\textit{x}}_2$ & Inclusion rate $\textbf{\textit{x}}_3$ & Inclusion rate $\textbf{\textit{x}}_4$ \\
     \hline
     ISLaB-LRT; 0.0  & 100\% & 100\% & 0\% & 0\% \\
     ISLaB-LRT; 0.1  & 100\% & 100\% & 0\% & 0\% \\
     ISLaB-LRT; 0.5  & 100\% & 100\% & 0\% & 0\% \\
     ISLaB-LRT; 0.9  & 100\% & 100\% & 100\% & 0\% \\
     \hline
     ISLaB-FLOW; 0.0  & 100\% & 100\% & 0\% & 0\% \\
     ISLaB-FLOW; 0.1  & 100\% & 100\% & 0\% & 0\% \\
     ISLaB-FLOW; 0.5  & 100\% & 100\% & 100\% & 0\% \\
     ISLaB-FLOW; 0.9  & 100\% & 100\% & 100\% & 0\% \\
     \hline
     BLR-LRT; 0.0  & 100\% & 100\% & 0\% & 0\% \\
     BLR-LRT; 0.1  & 100\% & 100\% & 0\% & 0\% \\
     BLR-LRT; 0.5  & 100\% & 100\% & 0\% & 0\% \\
     BLR-LRT; 0.9  & 100\% & 100\% & 100\% & 0\% \\
     \hline
     
     BLR-FLOW; 0.0  & 100\% & 100\% & 0\% & 0\% \\
     BLR-FLOW; 0.1  & 100\% & 100\% & 0\% & 0\% \\
     BLR-FLOW; 0.5  & 100\% & 100\% & 0\% & 0\% \\
     BLR-FLOW; 0.9  & 100\% & 100\% & 80\% & 0\% \\
     \hline
     IS-ANN-L1; 0.0  & 100\% & 100\% & 0\% & 0\% \\
     IS-ANN-L1; 0.1  & 100\% & 100\% & 0\% & 0\% \\
     IS-ANN-L1; 0.5  & 100\% & 100\% & 0\% & 0\% \\
     IS-ANN-L1; 0.9  & 100\% & 100\% & 100\% & 0\% \\
     \hline
    \end{tabular}
}
\label{tab:linInclution}
\end{table}
The results indicate that all methods perform well with and without sparsification, with close to 100$\%$ accuracy. This is perhaps not surprising as this is a very simple linear problem. From Table \ref{tab:linInclution} we can see that all methods include $x_3$ with high correlation. We note that our methods typically do very well at pruning away all the irrelevant (of 1544) weights and only use the two linear terms, where an example of the typical learned structure is presented in Figure \ref{fig:linExamples}. This behavior is not seen in IS-ANN-L1, as it uses around 8 weights. For the calibration, ISLaB-LRT has lower ECE and NLL than the frequentist approach and ISLaB-FLOW, whereas the latter two are similar. Overall, the linear method BLR-FLOW has the best results here, which is expected as the data generative process is linear.

\subsubsection{Simulated Non-linear Data}

The setup in this experiment is the same as in the previous section, however here  we add an interaction term and squared covariates:

\begin{equation}\label{eq:NonLinearProb}
    \eta(\textbf{X}) = \textit{w}_0 + \textit{w}_1 \textbf{\textit{x}}_1 + \textit{w}_2 \textbf{\textit{x}}_2 + \textit{w}_3 \textbf{\textit{x}}_1 \textbf{\textit{x}}_2 + \textit{w}_4 \textbf{\textit{x}}_1^2 + \textit{w}_5 \textbf{\textit{x}}_2^2 + \boldsymbol{\epsilon},
\end{equation}
with $\textit{w}_3 = \textit{w}_4 = \textit{w}_5 = 1$. The results can be found in Table \ref{tab:nonLinAcc}, \ref{tab:nonlincalibration}, and \ref{tab:nonLinInclution}.

\begin{table}[htp]
    \caption{Results from problem described in Equation (\ref{eq:NonLinearProb}). See Table \ref{tab:linAcc} for more information.}
    \resizebox{\textwidth}{!}{
    \centering
    \begin{tabular}{p{0.18\linewidth}p{0.20\linewidth}p{0.20\linewidth}p{0.15\linewidth}p{0.15\linewidth}p{0.1\linewidth}}
    \hline
     \multicolumn{5}{c}{Accuracy, used weights and depth metrics for the non-linear problem: $\textbf{y} = \textbf{\textit{x}}_1 + \textbf{\textit{x}}_2 + \textbf{\textit{x}}_1 \textbf{\textit{x}}_2 + \textbf{\textit{x}}_1^2 + \textbf{\textit{x}}_2^2 + 100$} \\
     \hline
     Model; $\varrho$ & ACC full & ACC sparse & Used weights  & Avg depth & Max depth\\
     \hline
     ISLaB-LRT; 0.0  & \textbf{99.8} (\textbf{99.7}, \textbf{99.9})\% & \textbf{99.8} (\textbf{99.7}, \textbf{99.9})\% & 45.0 (25, 74)  & 2.33 (2.23, 2.63) & 3.0 (3, 4)\\
     ISLaB-LRT; 0.1  & \textbf{99.8} (99.4, \textbf{99.9})\% & \textbf{99.8} (99.4, \textbf{99.9})\% & 45.0 (16, 63)  & 2.33 (2.22, 2.54) & 3.0 (3, 4)\\
     ISLaB-LRT; 0.5  & \textbf{99.8} (\textit{99.6}, \textbf{99.9})\% & \textbf{99.8} (\textbf{99.7}, \textbf{99.9})\% & 47.5 (38, 71)  & 2.30 (2.13, 2.56) & 3.0 (3, 4)\\
     ISLaB-LRT; 0.9  & \textit{99.7} (99.4, \textit{99.8})\% & \textit{99.7} (99.4, \textit{99.8})\% & 60.5 (33, 74)  & 2.32 (2.22, 2.47) & 3.5 (3, 4)\\
     \hline
     ISLaB-FLOW; 0.0 & \textbf{99.8} (\textit{99.6}, \textit{99.8})\% & \textbf{99.8} (\textbf{99.7}, \textit{99.8})\% & 70.5 (18, 109) & 2.46 (2.00, 2.67) & 4.0 (3, 5) \\
     ISLaB-FLOW; 0.1 & \textbf{99.8} (99.1, \textbf{99.9})\% & \textbf{99.8} (99.1, \textbf{99.9})\% & 70.0 (14, 107) & 2.56 (2.00, 2.73) & 4.5 (3, 5)\\
     ISLaB-FLOW; 0.5 & \textbf{99.8} (\textbf{99.7}, \textbf{99.9})\% & \textbf{99.8} (\textbf{99.7}, \textbf{99.9})\% & 84.5 (59, 113) & 2.60 (2.20, 2.83) & 4.0 (3, 5) \\
     ISLaB-FLOW; 0.9 & \textbf{99.8} (\textbf{99.7}, \textit{99.8})\% & \textbf{99.8} (\textbf{99.7}, \textbf{99.9})\% & 85.0 (38, 125) & 2.45 (2.00, 2.61) & 4.0 (3, 5)\\
     \hline
     BLR-LRT; 0.0    & 55.5 (55.3, 55.5)\% & 55.4 (55.0, 56.0)\% & 2 (2, 2) &1 (-,-) &1 (-,-) \\
     BLR-LRT; 0.1    & 55.5 (55.3, 55.5)\% & 55.4 (55.0, 56.0)\% & 2 (2, 2) &1 (-,-) &1 (-,-)  \\
     BLR-LRT; 0.5    & 55.5 (55.3, 55.5)\% & 55.4 (55.0, 56.0)\% & 2 (2, 2) &1 (-,-)&1 (-,-)  \\
     BLR-LRT; 0.9    & 55.6 (55.4, 55.8)\% & 55.6 (55.0, 56.1)\% & 2 (2, 2) &1 (-,-)&1 (-,-)  \\
     \hline
     BLR-FLOW; 0.0   & 55.6 (55.4, 55.9)\% & 55.8 (55.1, 55.9)\% & 2 (2, 2)&1 (-,-) &1 (-,-)  \\
     BLR-FLOW; 0.1   & 55.6 (55.5, 56.0)\% & 55.7 (55.3, 56.0)\% & 2 (2, 2) &1 (-,-) &1 (-,-)  \\
     BLR-FLOW; 0.5   & 55.6 (55.4, 55.9)\% & 55.7 (55.4, 55.9)\% & 2 (2, 2)&1 (-,-)&1 (-,-)  \\
     BLR-FLOW; 0.9   & 55.9 (55.5, 55.9)\% & 55.7 (55.0, 56.0)\% & 2 (2, 2)&1 (-,-)&1 (-,-)  \\
     \hline
     IS-ANN-L1; 0.0  & 99.3 (98.8, 99.6)\% & 99.3 (98.8, 99.6)\% & 31.5 (16, 40) & 2.30 (2.13, 2.40) & 3 (3, 3)\\
     IS-ANN-L1; 0.1  & 99.1 (97.8, 99.5)\% & 99.1 (97.8, 99.5)\% & 31.0 (27, 35) &2.20 (1.73, 2.38) & 3 (2, 3)\\
     IS-ANN-L1; 0.5  & 99.4 (97.8, 99.6)\% & 99.4 (97.8, 99.6)\% & 30.0 (12, 44)&2.27 (1.67, 2.35) & 3 (2, 3)\\
     IS-ANN-L1; 0.9  & 99.5 (98.8, 99.7)\% & 99.4 (98.8, 99.7)\% & 36.5 (26, 43) &2.31 (2.06, 2.38) & 3 (3, 3) \\
     \hline
    \end{tabular}
}
\label{tab:nonLinAcc}
\end{table}

\begin{table}[htp]
    \caption{Expected calibration error and NLL on simulated non-linear data (lower is better).}
    \resizebox{\textwidth}{!}{
    \centering
    \begin{tabular}{p{0.18\linewidth}p{0.2\linewidth}p{0.2\linewidth}p{0.2\linewidth}p{0.18\linewidth}}
     \hline
     \multicolumn{5}{c}{Calibration and negative log-likelihood for the non-linear problem: $\textbf{y} = \textbf{\textit{x}}_1 + \textbf{\textit{x}}_2 + \textbf{\textit{x}}_1 \textbf{\textit{x}}_2 + \textbf{\textit{x}}_1^2 + \textbf{\textit{x}}_2^2 + 100$} \\
     \hline
     Model; $\varrho$ & ECE & ECE sparse &  NLL & NLL sparse  \\
     \hline
     ISLaB-LRT; 0.0  & \textbf{0.004} (\textit{0.003}, 0.009) & \textbf{0.004} (\textit{0.003}, 0.009) & 0.009 (\textbf{0.005}, 0.015) & 0.009 (\textbf{0.005}, 0.015) \\ 
     ISLaB-LRT; 0.1  & \textit{0.005} (\textit{0.003}, 0.012) & \textbf{0.004} (\textit{0.003}, 0.012) & 0.009 (\textit{0.006}, 0.024) & 0.009 (\textit{0.006}, 0.024) \\ 
     ISLaB-LRT; 0.5  & \textit{0.005} (\textit{0.003}, \textbf{0.007}) & \textit{0.005} (\textit{0.003}, \textbf{0.007}) & 0.009 (\textit{0.006}, \textbf{0.012}) & 0.009 (\textit{0.006}, \textbf{0.012}) \\ 
     ISLaB-LRT; 0.9  & \textbf{0.004} (\textit{0.003}, 0.011) & \textbf{0.004} (\textit{0.003}, 0.012) & 0.009 (0.007, 0.020) & 0.009 (0.007, 0.020) \\
     \hline
     ISLaB-FLOW; 0.0 & \textbf{0.004} (\textit{0.003}, 0.009) & \textbf{0.004} (\textit{0.003}, 0.009) & \textit{0.008} (0.007, 0.016) & \textit{0.008} (0.007, 0.016) \\ 
     ISLaB-FLOW; 0.1 & \textbf{0.004} (\textit{0.003}, 0.011) & \textbf{0.004} (\textit{0.003}, 0.011) & \textit{0.008} (\textit{0.006}, 0.026) & \textit{0.008} (\textit{0.006}, 0.026) \\ 
     ISLaB-FLOW; 0.5 & \textbf{0.004} (\textbf{0.002}, \textit{0.008}) & \textbf{0.004} (\textbf{0.002}, \textit{0.008}) & \textbf{0.007} (\textit{0.006}, \textit{0.014}) & \textbf{0.007} (\textit{0.006}, \textit{0.014}) \\ 
     ISLaB-FLOW; 0.9 & \textbf{0.004} (\textit{0.003}, \textit{0.008}) & \textbf{0.004} (\textit{0.003}, \textit{0.008}) & \textit{0.008} (0.007, \textit{0.014}) & \textit{0.008} (0.007, \textit{0.014}) \\ 
     \hline
     BLR-LRT; 0.0    & 0.024 (0.023, 0.025) & 0.024 (0.020, 0.029) & 0.689 (0.689, 0.689) & 0.689 (0.689, 0.689)\\ 
     BLR-LRT; 0.1    & 0.025 (0.023, 0.025) & 0.024 (0.020, 0.029) & 0.689 (0.689, 0.689) & 0.689 (0.689, 0.689)\\
     BLR-LRT; 0.5    & 0.025 (0.023, 0.025) & 0.024 (0.020, 0.029) & 0.689 (0.689, 0.689) & 0.689 (0.689, 0.689)\\
     BLR-LRT; 0.9    & 0.025 (0.024, 0.027) & 0.025 (0.020, 0.031) & 0.689 (0.688, 0.689) & 0.689 (0.688, 0.689)\\
     \hline
     BLR-FLOW; 0.0   & 0.023 (0.020, 0.036) & 0.024 (0.017, 0.033) & 0.689 (0.688, 0.690) & 0.689 (0.688, 0.690)\\ 
     BLR-FLOW; 0.1   & 0.020 (0.019, 0.025) & 0.022 (0.017, 0.025) & 0.688 (0.688, 0.689) & 0.688 (0.688, 0.689)\\
     BLR-FLOW; 0.5   & 0.020 (0.019, 0.024) & 0.021 (0.017, 0.026) & 0.688 (0.688, 0.689) & 0.688 (0.688, 0.689)\\
     BLR-FLOW; 0.9   & 0.022 (0.020, 0.024) & 0.022 (0.017, 0.026) & 0.688 (0.688, 0.689) & 0.688 (0.688, 0.689)\\
     \hline
     IS-ANN-L1; 0.0  & 0.032 (0.028, 0.037) & 0.032 (0.028, 0.037) & 0.049 (0.043, 0.051) & 0.048 (0.043, 0.051) \\
     IS-ANN-L1; 0.1  & 0.031 (0.028, 0.060) & 0.031 (0.028, 0.060) & 0.049 (0.046, 0.104) & 0.049 (0.046, 0.104)  \\
     IS-ANN-L1; 0.5  & 0.034 (0.028, 0.048) & 0.034 (0.027, 0.048) & 0.048 (0.043, 0.089) & 0.048 (0.043, 0.089)\\
     IS-ANN-L1; 0.9  & 0.035 (0.026, 0.038) & 0.035 (0.026, 0.037) & 0.047 (0.045, 0.052) & 0.047 (0.045, 0.052) \\
     \hline
    \end{tabular}
}
\label{tab:nonlincalibration}
\end{table}

\begin{table}[htp]
    \caption{Same models as used in Table \ref{tab:nonLinAcc}. The inclusion rates of each covariate fed into the model.}
    \resizebox{\textwidth}{!}{
    \centering
    \begin{tabular}{p{0.2\linewidth}p{0.2\linewidth}p{0.2\linewidth}p{0.2\linewidth}p{0.2\linewidth}}
     \hline
     \multicolumn{5}{c}{Inclusion rates for the non-linear problem: $\textbf{\textit{y}} = \textbf{\textit{x}}_1 + \textbf{\textit{x}}_2 + \textbf{\textit{x}}_1 \textbf{\textit{x}}_2 + \textbf{\textit{x}}_1^2 + \textbf{\textit{x}}_2^2 + 100$} \\
     \hline
     Model; $\varrho$ & Inclusion rate $\textbf{\textit{x}}_1$& Inclusion rate $\textbf{\textit{x}}_2$ & Inclusion rate $\textbf{\textit{x}}_3$ & Inclusion rate $\textbf{\textit{x}}_4$ \\
     \hline
     ISLaB-LRT; 0.0  & 100\% & 100\% & 0\% & 0\% \\
     ISLaB-LRT; 0.1  & 100\% & 100\% & 0\% & 0\% \\
     ISLaB-LRT; 0.5  & 100\% & 100\% & 0\% & 0\% \\
     ISLaB-LRT; 0.9  & 100\% & 100\% & 100\% & 0\% \\
     \hline
     ISLaB-FLOW; 0.0  & 100\% & 100\% & 0\% & 0\% \\
     ISLaB-FLOW; 0.1  & 100\% & 100\% & 0\% & 0\% \\
     ISLaB-FLOW; 0.5  & 100\% & 100\% & 0\% & 0\% \\
     ISLaB-FLOW; 0.9  & 100\% & 100\% & 100\% & 0\% \\
     \hline
     BLR-LRT; 0.0  & 100\% & 100\% & 0\% & 0\% \\
     BLR-LRT; 0.1  & 100\% & 100\% & 0\% & 0\% \\
     BLR-LRT; 0.5  & 100\% & 100\% & 0\% & 0\% \\
     BLR-LRT; 0.9  & 50\% & 100\% & 50\% & 0\% \\
     \hline
     BLR-FLOW; 0.0  & 100\% & 100\% & 0\% & 0\% \\
     BLR-FLOW; 0.1  & 100\% & 100\% & 0\% & 0\% \\
     BLR-FLOW; 0.5  & 100\% & 100\% & 0\% & 0\% \\
     BLR-FLOW; 0.9  & 60\% & 100\% & 40\% & 0\% \\
     \hline
     IS-ANN-L1; 0.0  & 100\% & 100\% & 0\% & 0\% \\
     IS-ANN-L1; 0.1  & 100\% & 100\% & 10\% & 0\% \\
     IS-ANN-L1; 0.5  & 100\% & 100\% & 60\% & 0\% \\
     IS-ANN-L1; 0.9  & 100\% & 100\% & 100\% & 0\% \\
     
     \hline
    \end{tabular}
}
\label{tab:nonLinInclution}
\end{table}

\begin{figure}[htp]
\centering
  \includegraphics[width=.7\linewidth]{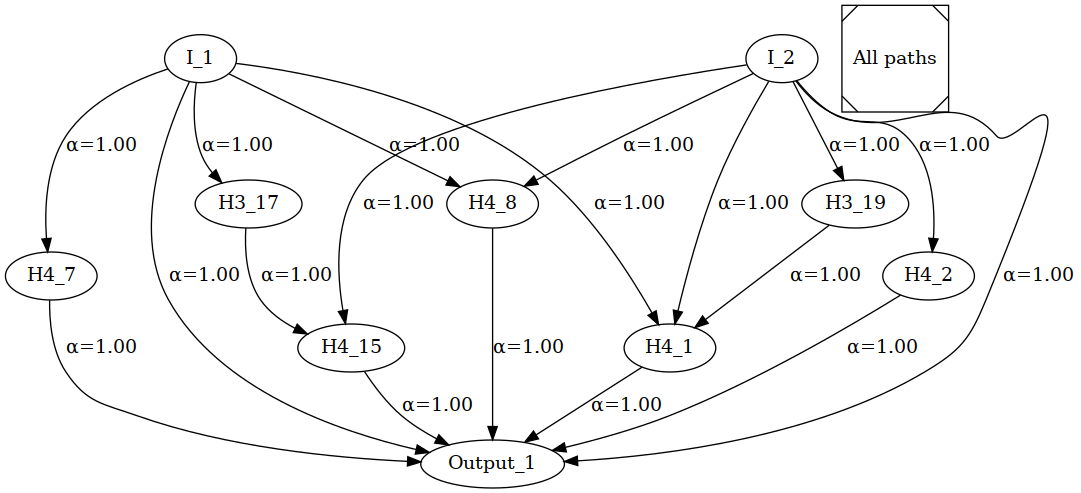}
\caption{Sparse structure learned of Equation (\ref{eq:NonLinearProb}) when using $\varrho=0.0$ for an FLOW-based model. $\alpha$ indicates the inclusion probability and displayed is the median probability model with its active paths.}
\label{fig:nonlinExamples}
\end{figure}

The results indicate that our methods perform well both with the full and sparse model, with high accuracy and correctly selecting the data-generating covariates. While the linear methods are also able to do this, they have no way to model the interaction and square terms, and therefore the accuracy is poor. As illustrated in Figure \ref{fig:nonlinExamples}, ISLaB learns a non-linear structure when presented with non-linear terms, which shows that it is able to adapt in accordance with the complexity of the problem at hand. IS-ANN-L1 has good accuracy with the full model and even sparser model, however, it has much higher ECE and NLL than  ISLaB methods, indicating significantly worse uncertainty handling. Furthermore,  IS-ANN-L1 struggles with the identification of the true data generating covariates for higher levels of correlations, unlike ISLaB models. These two toy experiments indicate that our novel method can perform well in a variety of settings, due to its flexibility and ability to $\emph{learn}$ the needed model complexity from the data, while maintaining predictive accuracy and uncertainty handling. 

\subsubsection{Wisconsin Breast Cancer}
Having established that the ISLaB method is capable of finding linear and non-linear structures when presented with simple problems where the functional form is known, the focus will now shift to real-world datasets. The first problem to be examined is the Wisconsin breast cancer (WBC) dataset. The data is labeled as either malignant or benign and can be found in \citet{wolberg1995breast}.

The neural network methods were initiated with 2 hidden layers, each having 50 hidden nodes, resulting in 5'580 weights in the full network. All covariates will be min-max scaled such that the largest value is equal to 1 and the smallest is equal to 0. We used 512 samples for training and 57 for testing the network after training.

\begin{figure}[htp]
\centering
  \includegraphics[width=.4\linewidth]{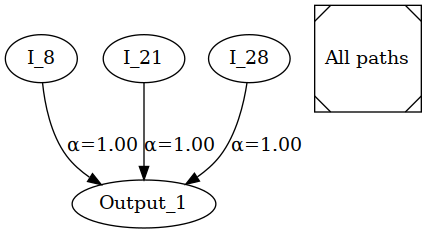}
\caption{Typical learned structure by an ISLaB-FLOW network trained on the WBC dataset.
$\alpha$ indicates the inclusion probability in the full model.}
\label{fig:WBCExample}
\end{figure}

The metrics related to the accuracy and the number of used weights are found in Table \ref{tab:WBCAcc}, and an example of a typical ISaB model is presented in Figure \ref{fig:WBCExample}. From Table \ref{tab:WBCAcc} we see that all methods have high accuracies for both full and sparse networks. We also see that the linear methods use about the same number of weights as our methods (again demonstrating our method's ability to prune away the vast majority of irrelevant covariates). Similar results were achieved in \citet{hubin2021flexible} and \citet{hubin2023fractional}, where a Bayesian generalized nonlinear model and Bayesian fractional polynomial model only learned linear terms. This underscores that the true structure might indeed be linear. The frequentist network uses roughly twice the number of weights, but still only linear connections. We note that the Laplace-SpaM method is using quite a lot more weights, partially due to not pruning the last layer. The results on calibration are in Table \ref{tab:WBCCalibration}. Here we see that our proposed method does better than the frequentist baseline, with both lower ECE and NLL, but worse than the other sparsification baselines. From Table \ref{tab:WBCAUC}, we see that all methods have nearly perfect ROC AUC scores, with IS-ANN-L1 performing the best.

\begin{table}[htp] 
    \caption{Results from Wisconsin Breast Cancer dataset. Two hidden layers, each consisting of 50 nodes, were used. 5'580 initial weights in the full network.}
    \resizebox{\textwidth}{!}{
    \centering
    \begin{tabular}{p{0.2\linewidth}p{0.20\linewidth}p{0.20\linewidth}p{0.17\linewidth}p{0.13\linewidth}p{0.1\linewidth}}
     \hline
     \multicolumn{6}{c}{Accuracy, used weights and depth metrics, WBC dataset} \\
     \hline
     Model & ACC full & ACC sparse & Used weights & Avg depth & Max depth \\
     \hline
     ISLaB-LRT     & \textit{96.5}          (\textit{96.5},          \textbf{98.2})\% & \textit{96.5}          (\textit{96.5},          \textbf{98.2})\% & 4.0 (3, 6)  & 1.0 (1.0, 1.0) & 1 (1, 1)\\
     ISLaB-FLOW    & \textit{96.5}          (93.0,          \textbf{98.2})\% & \textit{96.5}          (93.0,          \textbf{98.2})\% & 4.5 (3, 5)  & 1.0 (1.0, 1.0) & 1 (1, 1) \\
     BLR-LRT       & \textbf{98.2} (\textit{96.5},          \textbf{98.2})\% & \textbf{98.2} (\textit{96.5},          \textbf{98.2})\% & 5.0 (4, 6)  & 1 (-,-) & 1 (-,-)\\ 
     BLR-FLOW      & \textbf{98.2} (\textit{96.5},          \textbf{98.2})\% & \textbf{98.2} (93.0,          \textbf{98.2})\% & 5.0 (3, 5)  & 1 (-,-) & 1 (-,-) \\
     IS-ANN-L1     & \textbf{98.2} (\textbf{98.2}, \textbf{98.2})\% & \textbf{98.2} (\textbf{98.2}, \textbf{98.2})\% & 9.5 (6, 12) & 1.0 (1.0, 1.0) & 1 (1, 1)\\
     BNN-HORSE     & 93.0          (91.2,          \textit{94.7})\%          & 93.0          (89.5,          94.7)\%          & 7.0 (7, 7)  & 3 (-, -) & 3 (-, -)\\
     BNN-CONCRETE  & 93.0 (89.5, \textbf{98.2})\% & -                                                                & 55.0 (28, 187)\tablefootnote{Here and throughout the paper for BNN-CONCRETE we report the expected number of included weights based on learned dropout probabilities.}    & 3 (-, -) & 3 (-, -)\\
     Laplace-SpaM  & 91.2          (87.7,          \textbf{98.2})\% & 93.0          (89.5,          \textit{96.5})\%          & 140.0 (140, 140) & 3 (-, -) & 3 (-, -)\\
     \hline
    \end{tabular}
}
\label{tab:WBCAcc}
\end{table}

\begin{table}[htp] 
    \caption{Expected calibration error and NLL (lower is better).}
    \resizebox{\textwidth}{!}{
    \centering
    \begin{tabular}{p{0.2\linewidth}p{0.2\linewidth}p{0.2\linewidth}p{0.2\linewidth}p{0.18\linewidth}}
     \hline
     \multicolumn{5}{c}{Calibration and negative log-likelihood, WBC dataset} \\
     \hline
     Model & ECE full & ECE sparse & NLL full & NLL sparse \\
     \hline
     ISLaB-LRT    & 0.142          (0.132,          0.164)          & 0.146          (0.131,          0.165)          & 0.206          (0.192,          0.229)          & 0.207          (0.192,          0.230) \\
     ISLaB-FLOW   & 0.153          (0.131,          0.171)          & 0.155          (0.136,          0.170)          & 0.220          (0.206,          0.248)          & 0.213          (0.204,          0.247) \\
     BLR-LRT      & 0.153          (0.134,          0.170)          & 0.150          (0.139,          0.164)          & 0.207          (0.197,          0.227)          & 0.204          (0.190,          0.220)\\ 
     BLR-FLOW     & 0.172          (0.167,          0.195)          & 0.171          (0.127,          0.187)          & 0.234          (0.222,          0.287)          & 0.234          (0.222,          0.287) \\
     IS-ANN-L1    & 0.226          (0.219,          0.226)          & 0.226          (0.226,          0.227)          & 0.296          (0.295,          0.296)          & 0.296          (0.295,          0.297) \\
     BNN-HORSE    & \textit{0.054}          (0.034,          \textit{0.080})          & \textbf{0.053} (\textit{0.039},          \textit{0.093})          & \textbf{0.117} (\textbf{0.102}, \textbf{0.158}) & \textbf{0.123} (\textbf{0.105}, \textbf{0.186}) \\
     BNN-CONCRETE & \textbf{0.044} (\textit{0.030},          \textbf{0.070}) & -                                               & \textit{0.138}          (\textit{0.109},          \textit{0.178})          & - \\
     Laplace-SpaM & 0.060          (\textbf{0.026}, 0.112)          & \textit{0.057}          (\textbf{0.030}, \textbf{0.066}) & 0.197          (0.126,          0.369)          & \textit{0.156}          (\textit{0.128},          \textit{0.220}) \\
     \hline
    \end{tabular}
}
\label{tab:WBCCalibration}
\end{table}

\begin{table}[htp] 
    \caption{ROC AUC score on the WBC dataset, where a score of 1 represents a perfect classifier, and 0.5 is random guessing.}
    \resizebox{\textwidth}{!}{
    \small
    \centering
    \begin{tabular}{p{0.38\linewidth}p{0.38\linewidth}p{0.16\linewidth}}
     \hline
     \multicolumn{3}{c}{ROC AUC, WBC dataset} \\
     \hline
     Model & ROC AUC full & ROC AUC sparse \\
     \hline
     ISLaB-LRT     & \textit{0.998}          (\textit{0.995},          \textit{0.999})          & \textit{0.998}          (\textit{0.995},          \textit{0.999}) \\
     ISLaB-FLOW    & 0.997          (0.991,          \textit{0.999})          & 0.995          (0.992,          \textit{0.999}) \\
     BLR-LRT       & 0.997          (0.993,          \textbf{1.000}) & 0.997          (\textit{0.995},          \textbf{1.000})\\ 
     BLR-FLOW      & 0.995          (0.991,          0.997)          & 0.996          (\textit{0.995},          0.997) \\
     IS-ANN-L1     & \textbf{0.999} (\textbf{0.999}, \textit{0.999})          & \textbf{0.999} (\textbf{0.999}, \textit{0.999})  \\
     BNN-HORSE     & 0.991          (0.991,          0.991)          & 0.991          (0.991,          0.991) \\
     BNN-CONCRETE  & 0.989          (0.983,          0.989)          & -  \\
     Laplace-SpaM  & 0.989          (0.979,          0.991)          & 0.991          (0.989,          0.992)  \\
     \hline
    \end{tabular}
}
\label{tab:WBCAUC}
\end{table}

\subsubsection{Abalone Dataset}\label{sec:abaloneExp}
 This dataset presents a regression task, with the objective being to accurately predict the number of rings an abalone possesses, as this can serve as an indicator of its age. The data can be accessed from \citet{nash1995abalone}. For this problem, the neural network-based approaches have 2 hidden layers, each containing 200 nodes, giving 43'809 weights in the full model. The data will be partitioned into a training set with 3'759 observations, and a test set with 418 observations. 
 
\begin{figure}[htp]
\centering
  \includegraphics[width=.7\linewidth]{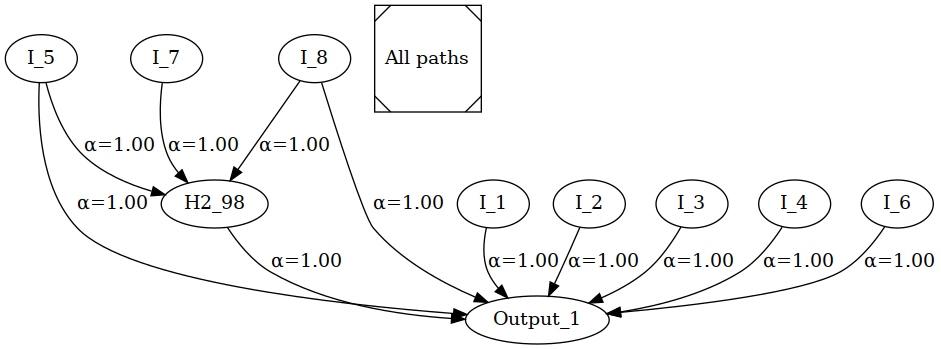}
\caption{Typical learned structure by ISLaB-FLOW network trained on the abalone dataset.
$\alpha$ indicates the inclusion probability in the full model.}
\label{fig:abaloneExample}
\end{figure}

 RMSE and Pinball metrics are presented in Table \ref{tab:abaloneCalibration}. Here, all neural network-based methods obtain similar results, on both the RMSE and Pinball loss, however, they outperform the linear models for these metrics. As seen in Table \ref{tab:abaloneAcc}, the ISLaB methods obtained much sparser networks than the other (non-linear) sparsification approaches, with an example of a learned structure is shown in Figure \ref{fig:abaloneExample}. It is further observed that our methods use more parameters than would be necessary for only linear terms. This is consistent with the results in \citet{hubin2023fractional} and \citet{hubin2021flexible}, where their models needed non-linear components to attain accurate predictions. 

\begin{table}[htp]
    \caption{Results from the Abalone dataset. Two hidden layers, each consisting of 200 nodes, were used. 43'809 initial weights in the full network.}
    \resizebox{\textwidth}{!}{
    \centering
    \begin{tabular}{p{0.2\linewidth}p{0.18\linewidth}p{0.18\linewidth}p{0.18\linewidth}p{0.15\linewidth}p{0.1\linewidth}}
     \hline
     \multicolumn{6}{c}{Correlation, used weights and depth metrics, Abalone dataset} \\
     \hline
     Model  & Corr full & Corr sparse &  Used weights & Avg depth & Max depth  \\
     \hline
     ISLaB-LRT     & \textit{0.78}          (\textbf{0.78}, \textit{0.79})          & \textit{0.78}          (\textbf{0.78}, \textbf{0.79}) & 24.5 (17, 34)       & 1.62 (1.50, 1.76) & 2 (2, 2)  \\
     ISLaB-FLOW    & \textit{0.78}          (0.75,          \textit{0.79})          & \textit{0.78}          (\textit{0.75},          \textbf{0.79}) & 11.0 (6, 20)        & 1.33 (1.00, 1.76) & 2 (1, 2)  \\
     BLR-LRT       & 0.74          (0.74,          0.74)          & 0.74          (0.74,          0.74)          & 9.0 (9, 9)          & 1 (-,-)           & 1 (-,-)   \\
     BLR-FLOW      & 0.76          (0.74,          0.76)          & 0.76          (0.74,          \textit{0.76})          & 7.0 (6, 8)          & 1 (-,-)           & 1 (-,-)   \\
     IS-ANN-L1     & \textit{0.78}          (\textit{0.77},          0.78)          & \textit{0.78}          (\textbf{0.78}, \textbf{0.79}) & 95.0 (80,104)       & 2.13 (2.00, 2.25) & 3 (3, 3)  \\
     BNN-HORSE     & \textbf{0.79} (\textbf{0.78}, \textit{0.79})          & \textbf{0.79} (\textbf{0.78}, \textbf{0.79}) & 141.5 (104, 180)    & 3 (-,-) & 3 (-,-)  \\
     BNN-CONCRETE  & \textbf{0.79} (\textbf{0.78}, \textit{0.79})          &  -                                           & 48.0 (26, 78)       & 3 (-,-) & 3 (-,-)  \\
      Laplace-SpaM & \textbf{0.79} (\textit{0.77},          \textbf{0.80}) & 0.77          (0.73,          \textbf{0.79}) & 2290.0 (2290, 2290) & 3 (-,-) & 3 (-,-)  \\
     
     \hline
    \end{tabular}
}
\label{tab:abaloneAcc}
\end{table}

\begin{table}[htp]
    \caption{Results from the Abalone dataset. Two hidden layers, each consisting of 200 nodes, were used.}
    \resizebox{\textwidth}{!}{
    \centering
    \begin{tabular}{p{0.22\linewidth}p{0.22\linewidth}p{0.22\linewidth}p{0.22\linewidth}p{0.15\linewidth}}
     \hline
     \multicolumn{5}{c}{RMSE and Pinball loss, Abalone dataset} \\
     \hline
     Model  & RMSE full & RMSE sparse & Pinball full & Pinball sparse \\
     \hline
     ISLaB-LRT    & 2.08          (2.06,          \textbf{2.10}) & \textit{2.08}          (\textit{2.06},          \textit{2.10})          & 0.73          (0.72,          \textbf{0.74}) & \textit{0.73}          (\textit{0.72},          \textbf{0.74}) \\
     ISLaB-FLOW   & 2.12          (2.07,          2.21)          & 2.12          (2.07,          2.21)          & 0.75          (0.72,          0.80)          & 0.75          (\textit{0.72},          0.80) \\
     BLR-LRT      & 2.24          (2.24,          2.25)          & 2.24          (2.24,          2.25)          & 0.80          (0.80,          0.81)          & 0.80          (0.80,          0.80) \\
     BLR-FLOW     & 2.19          (2.18,          2.27)          & 2.19          (2.18,          2.27)          & 0.79          (0.78,          0.82)          & 0.79          (0.78,          0.82) \\
     IS-ANN-L1    & 2.11          (2.10,          2.12)          & \textit{2.08}          (\textit{2.06},          \textbf{2.09}) & 0.78          (0.77,          0.78)          & 0.74          (0.73,          \textit{0.75})\\
     BNN-HORSE    & \textit{2.07}          (2.05,          \textit{2.11})          & \textbf{2.06} (\textbf{2.05}, \textit{2.10})          & \textit{0.72}          (0.71,          \textit{0.76})          & \textbf{0.72} (\textbf{0.71}, \textit{0.75})\\
     BNN-CONCRETE & \textbf{2.06} (\textit{2.04},          \textbf{2.10}) & -                                            & \textbf{0.71} (\textit{0.70},          \textbf{0.74}) &  -\\
     Laplace-SpaM & \textit{2.07}          (\textbf{2.03}, 2.13)          & 2.12          (2.07,          2.28)          & \textbf{0.71} (\textbf{0.69}, 0.79)          &  0.74         (\textit{0.72},          0.80)\\
     
     \hline
    \end{tabular}
}
\label{tab:abaloneCalibration}
\end{table}

\subsubsection{Mice Protein Dataset}

In the mice protein dataset from \citet{higuera2015self}, mice are categorized into one of 8 different classes, where the explanations for each class can be found in Table \ref{tab:miceClassDesc}. For the neural network approaches, 4 hidden layers will be used, where each hidden layer has 100 hidden nodes, which gives 62'216 weights in the full model. The training set will have 972 observations, and 108 observations will be used as a test set to evaluate the performance of the trained models. 

Table \ref{tab:miceAcc} presents the results related to the accuracy, used weights, and depth. We see that all methods generally have very good accuracy, with the linear methods slightly worse. The results on calibration can be found in Table \ref{tab:micecalibration}. Here, we see that BNN-HORSE appears to be the best calibrated method, with low ECE and NLL on both the full and sparse networks. We note here that our method does worse than the other sparsification approaches. 

\begin{figure}[htp]
\centering

\begin{subfigure}{\linewidth} 
\centering\includegraphics[width=\linewidth]{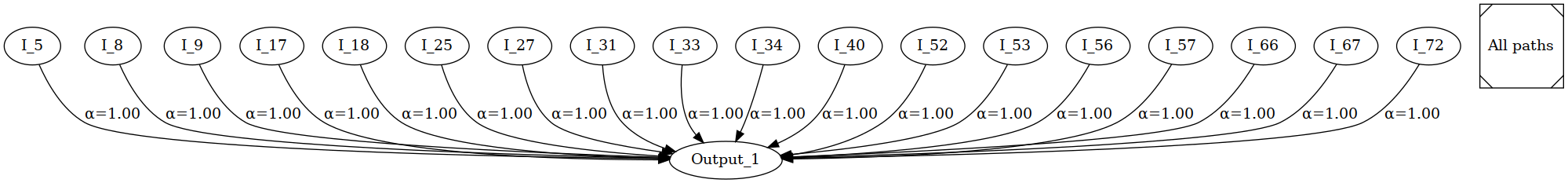} 
\end{subfigure}
\begin{subfigure}{\linewidth} 
\centering\includegraphics[width=\linewidth]{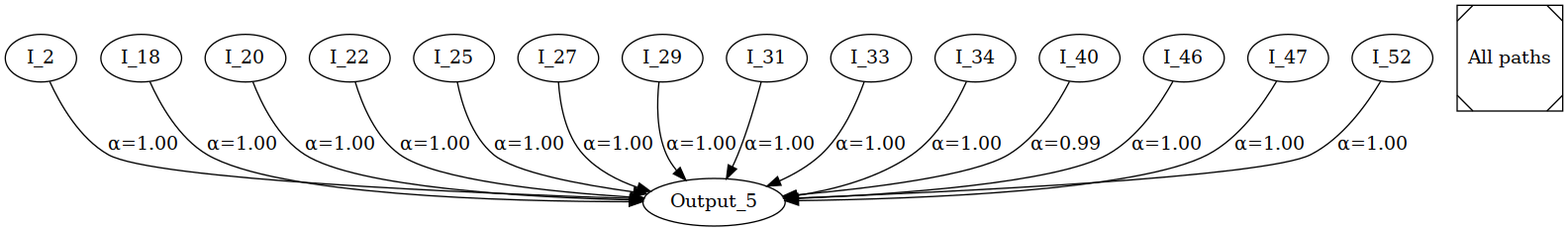} 
\end{subfigure}
\caption{Typical learned structure by ISLaB-FLOW network trained on the mice protein dataset. At the top, all connections go into class 1, with connections into class 5 at the bottom.}\label{fig:miceExample_flow} 
\end{figure}

On this dataset, the linear methods perform quite well, and we see that the non-linear methods are mostly reduced to linear methods as well. Again we see that our approach uses significantly fewer weights than the other sparsification methods. To illustrate how our methods can help with interpretability, Figure \ref{fig:miceExample_flow} shows all paths that lead to classes 1 and 5 for a FLOW-based network. This simplifies the task for researchers when choosing proteins to further investigate, as information about which proteins contribute to the different classes is available. Furthermore, since all connections are linear, it will be feasible to determine whether covariates contribute positively or negatively to a given class.

\begin{table}[htp]
    \caption{Results from the Mice protein dataset. Four hidden layers, each consisting of 100 nodes were used. 62'216 initial weights in the full network.}
    \resizebox{\textwidth}{!}{
    \centering
    \begin{tabular}{p{0.20\linewidth}p{0.20\linewidth}p{0.20\linewidth}p{0.20\linewidth}p{0.19\linewidth}p{0.1\linewidth}}
     \hline
     \multicolumn{6}{c}{Accuracy, used weights and depth metrics, Mice protein dataset} \\
     \hline
     Model & ACC full & ACC sparse & Used weights & Avg depth& Max depth\\
     \hline
     ISLaB-LRT     & \textit{99.1}          (98.1,          \textbf{100})\%  & \textbf{99.1} (\textit{97.2},          \textbf{100})\%  & 140.0 (134, 145) & 1.00 (1.00, 1.00) & 1 (1, 1) \\ 
     ISLaB-FLOW    & 97.2          (96.3,          \textit{99.1})\%          & \textit{97.2}          (96.3,          \textit{99.1})\%          & 98.5 (88, 111)   & 1.00 (1.00, 1.00) & 1 (1, 1) \\ 
     BLR-LRT       & 96.8          (96.3,          97.2)\%          & 96.8          (96.3,          98.1)\%          & 63.0 (60, 65) &1 (-,-)&1 (-,-)\\
     BLR-FLOW      & 96.8          (94.4,          \textit{99.1})\%          & 96.3          (94.4,          \textit{99.1})\%          & 103.0 (85, 115)&1 (-,-)&1 (-,-) \\
     IS-ANN-L1     & \textit{99.1}          (\textit{99.1},          \textbf{100})\%  & \textbf{99.1} (\textbf{99.1}, \textbf{100})\%  & 214.5 (209, 234) & 1.02 (1.00, 1.11) & 2 (1, 2)\\
     BNN-HORSE     & \textit{99.1}          (\textit{99.1},          \textbf{100})\%  & \textbf{99.1} (\textbf{99.1}, \textbf{100})\%  & 542.5 (483, 650) & 5 (-, -) & 5 (-, -)\\
     BNN-CONCRETE  & \textit{99.1}          (98.1,          \textbf{100})\%  & -                                              & 9282.0 (7566, 11154)  & 5 (-, -) & 5 (-, -)\\
     Laplace-SpaM  & \textbf{100}  (\textbf{100},  \textbf{100})\%  & 94.0          (89.8,          97.2)\%          & 1931.0 (1931, 1931) & 5 (-, -) & 5 (-, -)\\
     
     \hline
    \end{tabular}
}
\label{tab:miceAcc}
\end{table}

\begin{table}[htp]
    \caption{Expected calibration error and NLL on Mice data. Lower is better.}
    \resizebox{\textwidth}{!}{
    \centering
    \begin{tabular}{p{0.25\linewidth}p{0.25\linewidth}p{0.25\linewidth}p{0.25\linewidth}p{0.18\linewidth}}
     \hline
     \multicolumn{5}{c}{Calibration and negative log-likelihood, Mice protein dataset} \\
     \hline
     Model & ECE full & ECE sparse & NLL full & NLL sparse\\
     \hline
     ISLaB-LRT    & 0.072          (0.053,          0.092)          & 0.062          (0.032,          0.080)          & 0.203          (0.162,          0.238)          & 0.185          (0.140,          0.236) \\
     ISLaB-FLOW   & 0.053          (0.042,          0.067)          & 0.053          (0.035,          0.067)          & 0.340          (0.184,          0.649)          & 0.318          (0.167,          0.675) \\
     BLR-LRT      & 0.036          (0.017,          0.055)          & \textit{0.033}          (\textit{0.014},          \textit{0.050})          & 0.286          (0.207,          0.375)          & 0.278          (0.205,          0.378)\\
     BLR-FLOW     & 0.057          (0.033,          0.072)          & 0.040          (0.050,          0.085)          & 0.485          (0.262,          0.860)          & 0.485          (0.282,          0.806)\\
     IS-ANN-L1    & 0.066          (0.056,          0.068)          & 0.066          (0.056,          0.068)          & 0.079          (0.077,          0.081)          & \textit{0.079}          (\textit{0.077},          \textit{0.081})\\
     BNN-HORSE    & \textit{0.004}          (0.002,          \textit{0.013})          & \textbf{0.005} (\textbf{0.002}, \textbf{0.015}) & \textit{0.004}          (0.002,          \textit{0.061})          & \textbf{0.005} (\textbf{0.002} ,\textbf{0.073})\\
     BNN-CONCRETE & 0.008          (\textit{0.001},          0.018)          & -                                               & 0.044          (\textit{0.001},          0.176)          & -\\
     Laplace-SpaM & \textbf{0.000} (\textbf{0.000}, \textbf{0.002}) & 0.096          (0.069,          0.124)          & \textbf{0.000} (\textbf{0.000}, \textbf{0.003}) & 0.252          (0.139,          0.369)\\
     
     \hline
    \end{tabular}
}
\label{tab:micecalibration}
\end{table}

\subsubsection{FER2013 Dataset - Happy, Surprised and Neutral}\label{sec:FER}
The next example to be examined is the FER2013 image dataset, which is based on a Kaggle competition \citep{erhan2013challenges}. The dataset consists of $48\times 48$ images with expressions of anger, disgust, fear, happiness, sadness, surprise, and neutrality. However, for this experiment, the focus will be on the three expressions: happy, surprised, and neutral. For the neural network-based implementations, there will be 2 hidden layers, each containing 200 hidden nodes, resulting in 969'112 weights in the full model. The network will be trained with 17'270 observations, and tested with the remaining 1'919 observations.

The obtained accuracy and the final density after training can be found in Table \ref{tab:FERAcc}. We see that the convolutional approach with flows has much higher accuracy than other approaches, which is not surprising. From the feed forward networks, we see that BNNs with horseshoe priors have the highest accuracy, albeit with a orders of magnitude more weights than our approach. Indeed we see our approach as well as the frequentist baseline all collapse to a linear model on this dataset, and we see they have similar accuracy as the linear methods. The results on calibration are in Table \ref{tab:FERcalibration}, where we see quite similar results for all approaches, aside from BNN-CONCRETE which does worse. 

\begin{figure}[htp]
\centering
\begin{subfigure}{.25\textwidth}
  \centering
  \includegraphics[width=.97\linewidth]{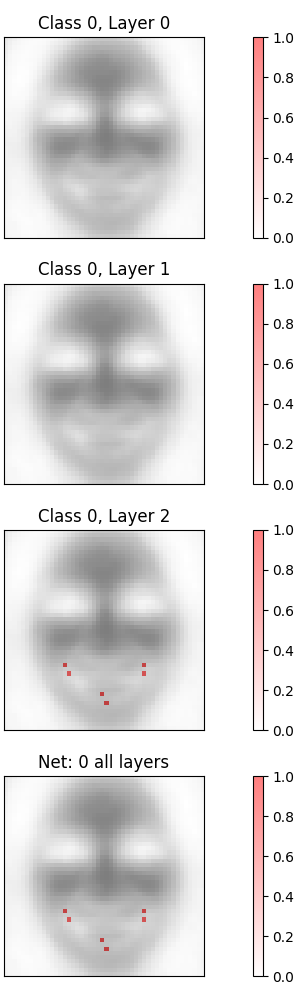}
  \caption{Happy expression}
  \label{fig:GlobalExplainFERlrtHappy}
\end{subfigure}%
\begin{subfigure}{.25\textwidth}
  \centering
\includegraphics[width=.98\linewidth]{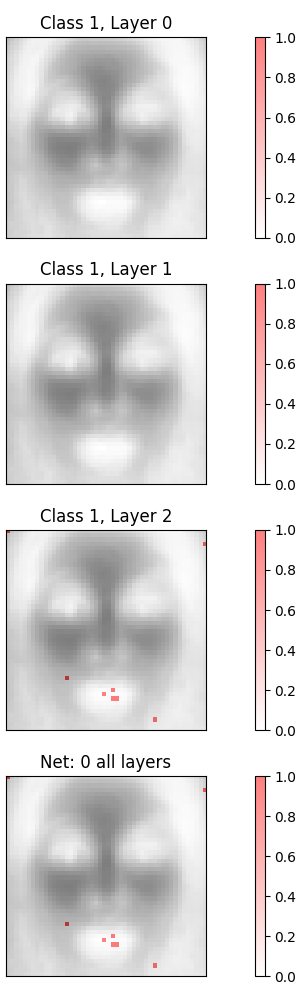}
  \caption{Surprised expression}
  \label{fig:GlobalExplainFERlrtSurprised}
\end{subfigure}
\begin{subfigure}{.25\textwidth}
  \centering
  \includegraphics[width=.99\linewidth]{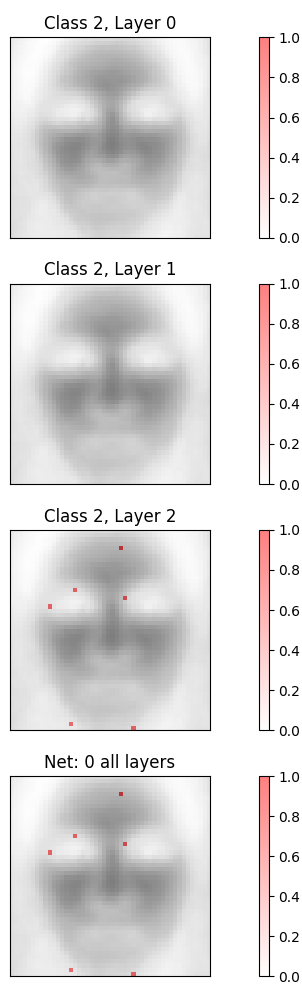}
  \caption{Neutral expression}
  \label{fig:GlobalExplainFERlrtNeural}
\end{subfigure}
\caption{Global explanation of a trained ISLaB-FLOW model showing pixels in the MPM that predict the addressed classes in the FER2013 dataset.}
\label{fig:GlobalExplainationFERlrt}
\end{figure}

Figure \ref{fig:GlobalExplainationFERlrt} illustrates the pixels used to make predictions in an ISLaB-FLOW network. A particularly interesting observation from these explanations is that the method appears to focus on the correct regions when making predictions. For instance, Figure \ref{fig:GlobalExplainFERlrtHappy}, which illustrates which pixels are used for predicting a happy facial expression, indicates that the method uses pixels around the mouth region where a smile is typically present. This seems reasonable as it is a region that humans also focus on when determining whether a person is happy. We can also see that the input layer and the first hidden layer are redundant and do not contribute to the predictions, indicating a linear model, which explains why the linear methods have comparable to neural networks accuracy on this dataset.

\begin{table}[htp]
    \caption{Results from FER2013 HSN dataset. Two hidden layers, each consisting of 200 nodes, were used. 969'112 initial weights in the full network.}
    \resizebox{\textwidth}{!}{
    \centering
    \begin{tabular}{p{0.22\linewidth}p{0.22\linewidth}p{0.22\linewidth}p{0.24\linewidth}p{0.22\linewidth}p{0.1\linewidth}}
     \hline
     \multicolumn{6}{c}{Accuracy, used weights and depth metrics, FER2013 HSN dataset} \\
     \hline
     Model & ACC full & ACC sparse & Used weights & Avg depth & Max depth\\
     \hline
     ISLaB-LRT       & 59.6          (58.5,          60.6)\%          & 59.5          (58.3,          60.4)\%          & 24.0 (22, 28) & 1.0 (1.0, 1.0) & 1 (1, 1)\\
     ISLaB-FLOW      & 58.4          (57.3,          59.5)\%          & 58.3          (57.4,          59.7)\%          & 24.0 (20, 36) & 1.0 (1.0, 1.0) & 1 (1, 1)\\
     BCNN-ISLaB-LRT  & 61.0          (58.9,          62.9)\%          & 56.0          (36.7,          61.3)\%          & - (-,-) & - (-,-) & - (-,-)\\
     BCNN-ISLaB-FLOW & \textbf{76.8} (\textbf{75.1}, \textbf{77.4})\% & \textbf{75.1} (\textbf{67.8}, \textbf{75.7})\% & - (-,-) & - (-,-) & - (-,-)\\
     BLR-LRT         & 60.3          (60.0,          61.3)\%          & 60.4          (60.1,          61.1)\%          & 34.0 (32, 38) & 1 (-,-) & 1 (-,-)\\
     BLR-FLOW        & 60.7          (60.3,          61.1)\%          & 60.9          (59.9,          61.6)\%          & 28.5 (25, 32) & 1 (-,-) & 1 (-,-)\\
     IS-ANN-L1       & 62.3          (62.3,          62.4)\%          & 63.9          (63.2,          64.4)\%          & 746.5 (736, 752) & 1.0 (1.0, 1.0) & 1 (1, 1) \\
     BNN-HORSE       & \textit{68.4}          (\textit{68.1},          \textit{69.3})\%          & \textit{68.4}          (\textit{67.5},          \textit{69.1})\%          & 3370.5 (2203, 4632) & 3 (-,-) & 3 (-,-) \\
     BNN-CONCRETE    & 64.6          (62.9,          66.0)\%          & -                                              & 10.0 (5, 20)  & 3 (-,-) & 3 (-,-) \\
     Laplace-SpaM    & 65.8          (63.7,          67.3)\%          & 64.7          (62.5,          66.2)\%          & 5608.0 (5608, 5608) & 3 (-,-) & 3 (-,-) \\
     \hline
    \end{tabular}
}
\label{tab:FERAcc}
\end{table}

\begin{table}[htp]
    \caption{Expected calibration error and NLL on the FER2013 HSN dataset, where lower values are better.}
    \resizebox{\textwidth}{!}{
    \centering
    \begin{tabular}{p{0.3\linewidth}p{0.3\linewidth}p{0.3\linewidth}p{0.3\linewidth}p{0.17\linewidth}}
     \hline
     \multicolumn{5}{c}{Calibration and negative log-likelihood, FER2013 HSN dataset} \\
     \hline
     Model & ECE full & ECE sparse & NLL full & NLL sparse\\
     \hline
     ISLaB-LRT       & \textbf{0.024} (\textit{0.014},          0.046)          & 0.023          (\textit{0.013},          0.043)          & 0.882          (0.875,          0.893)          & 0.883          (0.875,          0.894) \\
     ISLaB-FLOW      & \textit{0.023}          (0.017,          \textbf{0.030}) & \textit{0.022}          (0.014,          0.059)          & 0.938          (0.921,          0.954)          & 0.898          (0.888,          0.919) \\
     BCNN-ISLaB-LRT  & 0.033          (0.020,          0.042)          & 0.059          (0.015,          0.370)          & 0.891          (0.873,          0.911)          & 0.948          (0.881,          1.310) \\
     BCNN-ISLaB-FLOW & 0.060          (0.028,          0.080)          & 0.046          (0.030,          0.074)          & \textbf{0.667} (\textbf{0.644}, 0.895)          & \textbf{0.688} (\textbf{0.625}, \textit{0.813}) \\
     BLR-LRT         & 0.027          (\textbf{0.013}, 0.038)          & 0.029          (0.023,          \textit{0.034})          & 0.876          (0.873,          0.881)          & 0.876          (0.873,          0.883) \\
     BLR-FLOW        & 0.030          (\textbf{0.013}, 0.035)          & 0.031          (0.017,          0.042)          & 0.881          (0.873,          0.896)          & 0.876          (0.867,          0.889) \\
     IS-ANN-L1       & 0.025          (0.024,          \textit{0.028})          & \textbf{0.021} (\textbf{0.012}, \textbf{0.024}) & 0.847          (0.846,          0.847)          & 0.828          (0.828,          0.829) \\
     BNN-HORSE       & 0.028          (0.020,          0.037)          & 0.030          (0.021,          0.037)          & \textit{0.730}          (\textit{0.722},          \textbf{0.743}) & \textit{0.733}          (\textit{0.720},          \textbf{0.744}) \\
     BNN-CONCRETE    & 0.210          (0.187,          0.248)          & -                                               & 1.279          (1.131,          1.686)          &- \\
     Laplace-SpaM    & 0.058          (0.026,          0.088)          & 0.028          (0.014,          0.050)          & 0.790          (0.772,          \textit{0.806})          & 0.807          (0.772,          0.843) \\
     \hline
    \end{tabular}
}
\label{tab:FERcalibration}
\end{table}

\subsubsection{MNIST Dataset}

The final example in this section is the MNIST dataset from \citet{lecun1998gradient}. The MNIST dataset consists of $28\times 28$ images of digits between 0 and 9. The MNIST dataset has established train and test sets, which consist of 60'000 and 10'000 images, respectively. For these experiments, we initiated networks with two hidden layers, each consisting of 600 hidden nodes, giving 1'314'640 weights in the full model. For the L1 regularized networks, we trained networks using four different levels of regularization: $\lambda=\{0.01,0.1,0.2,1.0\}$. This is done to illustrate how different regularization levels will induce different sparsity levels and thus different performance results. 

From Table \ref{tab:MNISTAcc}, we have that the lowest regularization level, $\lambda=0.01$, gives higher accuracies than the networks with the highest regularization, $\lambda=1.0$. However, this comes at the cost of including approximately 11 times the amount of weights and using a much more complex network structure, as indicated by the average and max depth metrics. We see that with the stronger regularizations, the ANN becomes a linear model, with accuracies comparable to our linear Bayesian baselines (but still using more weights). We note that the ISLaB-FLOW method is able to get almost 97$\%$ accuracy, while only using around 900 weights (of a possible 1'314'640) corresponding to over 99.9\% sparsification. This is impressive, as it hardly uses more weights than the linear methods, but still includes many non-linear terms. To our best awareness this is the most sparse neural network model achieving over 96\% accuracy on MNIST data. 

The calibration results displayed in Table \ref{tab:mnistcalibration} show that the Laplace-SpaM method has the lowest ECE with the full network, while the convolutional networks have the smallest sparse ECE. In addition, the frequentist network with the smallest regularization term has the lowest NLL.

\begin{figure}[htp]
\centering
\begin{subfigure}{.413\textwidth}
  \centering
  \includegraphics[width=.65\linewidth]{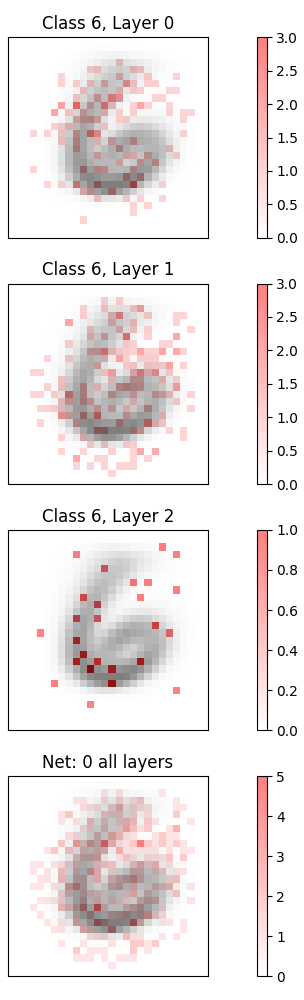}
  \caption{LRT model.}
  \label{fig:lrtGlobal6}
\end{subfigure}%
\begin{subfigure}{.413\textwidth}
  \centering
  \includegraphics[width=.647\linewidth]{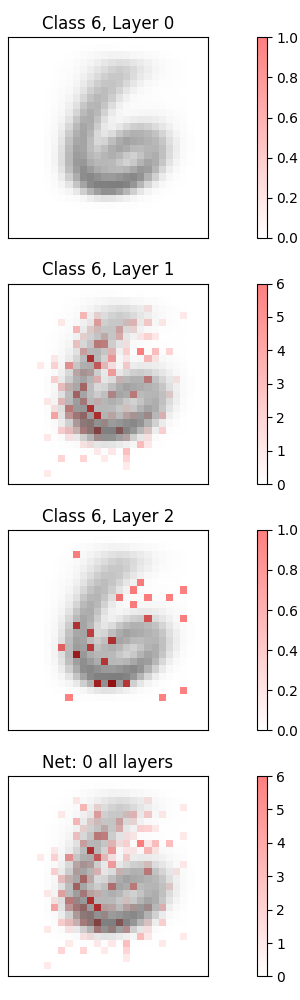}
  \caption{FLOW model.}
  \label{fig:flowGlobal6}
\end{subfigure}
\caption{Pixels used for predicting digit 6 in both ISLaB-LRT and ISLaB-FLOW implementations. Layer 0 indicates all pixels used in the input layer, while Layer 1 and Layer 2 indicate pixels used when the input skips to the respective layers. The last image shows all pixels utilized throughout the model. The 6 digit displayed is an average over all 6 images available in the train set.}
\label{fig:globalExplainMNIST6}
\end{figure}

Figure \ref{fig:globalExplainMNIST6} shows the global explanations for predicting digit 6 for an ISLaB-LRT model and an ISLaB-FLOW model. It can be seen that the model focuses on the most relevant regions of the images, as it ignores the redundant information around the edges, and focuses mainly around the center of the image, where the digit typically appears. The two models also seem to focus on crucial regions to distinguish a 6 from any other digit, like the lower left part of the digit, while also focusing on regions where a 6 usually does not attain any pixels, like the upper right part.

\begin{table}[htp]
    \caption{Results from the MNIST dataset, using two hidden layers consisting of 600 neurons each. 1'314'640 initial weights in the full network. The strength of the regularization, $\lambda$, is varied to induce different densities for the frequentist network.}
    \resizebox{\textwidth}{!}{
    \centering
    \begin{tabular}{p{0.22\linewidth}p{0.22\linewidth}p{0.22\linewidth}p{0.28\linewidth}p{0.22\linewidth}p{0.1\linewidth}}
     \hline
     \multicolumn{6}{c}{Accuracy, used weights and depth metrics, MNIST dataset} \\
     \hline
     Model                      & ACC full                                        & ACC sparse                                      & Used weights        & Avg depth         & Max depth\\
     \hline
     ISLaB-LRT                  & 96.8          (96.7,          97.0)\%           & 96.5          (96.4,          96.8)\%           & 1157.5 (1112, 1258) & 1.94 (1.89, 1.98) & 3 (3, 3) \\ 
     ISLaB-FLOW                 & 96.7          (96.6,          97.1)\%           & 96.7          (96.5,          97.0)\%           & 886.5 (824, 935)    & 1.65 (1.62, 1.72) & 3 (2, 3) \\
     BCNN-ISLaB-LRT             & \textbf{98.4} (97.6,          \textbf{98.7})\%  & \textit{98.3}          (97.4,          \textbf{98.7})\%  & - (-,-) & - (-,-) & - (-,-) \\ 
     BCNN-ISLaB-FLOW            & \textit{98.3}          (\textbf{98.2}, 98.4)\%           & 98.2          (\textbf{98.2}, \textit{98.4})\%           & - (-,-) & - (-,-) & - (-,-) \\
     BLR-LRT                    & 92.0          (91.9,          92.2)\%           & 91.9          (91.9,          92.2)\%           & 593.0 (581, 604) &1 (-,-)&1 (-,-)\\
     BLR-FLOW                   & 91.9          (88.2,          92.3)\%           & 91.9          (86.0,          92.1)\%           & 606.0 (585, 666)&1 (-,-)&1 (-,-) \\
     IS-ANN-L1, $\lambda = 1$   & 90.6          (90.3,          90.7)\%           & 90.6          (90.5,          90.6)\%           & 1092.5 (1076, 1106) & 1.00 (1.00, 1.00) & 1 (1, 1)\\
     IS-ANN-L1, $\lambda = 0.2$ & 92.3          (92.1,          92.4)\%           & 92.2          (92.1,          92.2)\%           & 1839.5 (1808, 1853)& 1.00 (1.00, 1.00) & 1 (1, 1)\\
     IS-ANN-L1, $\lambda = 0.1$ & 92.7          (92.3,          94.4)\%           & 92.4          (92.2,          93.9) \%          & 2196.0 (2169, 2701)& 1.00 (1.00, 1.24)  & 1 (1, 2)\\
     IS-ANN-L1, $\lambda = 0.01$& \textit{98.3}          (\textbf{98.2}, 98.4)\%           & \textbf{98.4} (\textbf{98.2}, \textit{98.4}) \%          & 12276.0 (10968, 13023)& 1.91 (1.86, 2.06)   & 3 (2, 3)\\
     BNN-HORSE                  & \textit{98.3}          (\textbf{98.2}, 98.3)\%           & 98.0          (97.9,          98.1) \%          & 29342.0 (28188, 30407)& 3 (-,-)   & 3 (-, -)\\
     BNN-CONCRETE               & \textbf{98.4} (97.8,          \textit{98.6})\%           & -                                               & 90139.0 (75455, 103151) & 3 (-,-)   & 3 (-, -)\\
     Laplace-SpaM               & 98.1          (\textit{98.0},          98.4)\%           & 98.0          (\textit{98.0},          98.1) \%          & 22608.0 (22608, 22608)& 3 (-,-)   & 3 (-, -)\\
     
     \hline
    \end{tabular}
}
\label{tab:MNISTAcc}
\end{table}

\begin{table}[htp]
    \caption{Expected calibration error and NLL on MNIST.}
    \resizebox{\textwidth}{!}{
    \centering
    \begin{tabular}{p{0.27\linewidth}p{0.27\linewidth}p{0.27\linewidth}p{0.27\linewidth}p{0.18\linewidth}}
     \hline
     \multicolumn{5}{c}{Calibration and negative log-likelihood, MNIST dataset} \\
     \hline
     Model & ECE full & ECE sparse & NLL full & NLL sparse\\
     \hline
     ISLaB-LRT                & 0.010          (0.006,          0.012)          & 0.010          (0.008,          0.011)          &  0.161         (0.151,           0.168)         & 0.152          (0.144,          0.161) \\
     ISLaB-FLOW               & \textit{0.005}          (\textbf{0.002}, 0.\textit{009})          & \textit{0.005}          (\textit{0.003},          0.009)          &  0.135         (0.118,           0.155)         & 0.129          (0.114,          0.144) \\
     BCNN-ISLaB-LRT           & 0.006          (\textit{0.005},          0.010)          & \textbf{0.003} (\textbf{0.002}, \textbf{0.004}) &  0.072         (0.064,           0.098)         & \textit{0.066}          (\textit{0.058},          0.093) \\
     BCNN-ISLaB-FLOW          & 0.009          (0.006,          0.013)          & \textbf{0.003} (\textit{0.003},          \textit{0.006})          &  0.084         (0.072,           0.094)         & 0.073          (0.064,          0.083) \\
     BLR-LRT                  & 0.019          (0.017,          0.020)          & 0.018          (0.015,          0.019)          &  0.311         (0.303,           0.320)         & 0.305          (0.301,          0.308)\\
     BLR-FLOW                 & 0.021          (0.018,          0.063)          & 0.019          (0.017,          0.061)          &  0.327         (0.310,           2.033)         & 0.316          (0.307,          2.515)\\
     ANN-L1, $\lambda = 1$    & 0.075          (0.073,          0.077)          & 0.077          (0.077,          0.078)          &  0.337         (0.335,           0.341)         & 0.357          (0.356,          0.357)\\
     ANN-L1, $\lambda = 0.2$  & 0.035          (0.033,          0.037)          & 0.036          (0.035,          0.037)          &  0.276         (0.275,           0.279)         & 0.286          (0.285,          0.288)\\
     ANN-L1, $\lambda = 0.1$  & 0.026          (0.021,          0.027)          & 0.028          (0.019,          0.029)          &  0.265         (0.191,           0.268)         & 0.277          (0.214,          0.283)\\
     ANN-L1, $\lambda = 0.01$ & 0.008          (0.007,          0.010)          & 0.008          (0.007,          0.009)          & \textbf{0.050} (\textbf{0.043}, \textbf{0.061}) & \textbf{0.055} (\textbf{0.053}, \textbf{0.057}) \\
     BNN-HORSE                & 0.008          (0.008,          \textit{0.009})          & 0.010          (0.008,          0.011)          &  0.067         (0.064,           \textit{0.071})         & 0.078          (0.072,          0.082) \\
     BNN-CONCRETE             & 0.010          (0.008,          0.012)          & -                                               &  0.086         (0.070,           0.111)         & - \\
     Laplace-SpaM             & \textbf{0.004} (\textbf{0.002}, \textbf{0.006}) & 0.013          (0.011,          0.015)          &  \textit{0.060}         (\textit{0.056},           \textit{0.071})         & 0.068          (0.066,          \textit{0.071}) \\
     
     \hline
    \end{tabular}
}
\label{tab:mnistcalibration}
\end{table}

\subsection{Local Explanations}

If a network employs activation functions that allow for linear contributions to the output, like the ReLU function, local explanations can be obtained for all predictions, as described in Section \ref{sec:local_explain}. Additionally, since ISLaB models use variational distributions to describe each weight, credibility intervals can be retrieved for each covariate contribution, alongside a credibility interval for the prediction. Hence, to illustrate this process and demonstrate how it can be plotted, several examples of local explanations will be presented based on three of the problems introduced above. Since the focus in this subsection is purely on displaying examples, we will only focus on an LRT-based ISLaB method in this section. Local experiments can be found and reproduced from  \url{https://github.com/eirihoyh/ISLaB-LBBNN/tree/JAIR/implementations/local_explain}.

\subsubsection{Linear Function}

\begin{figure}[htp]
  \centering
    \includegraphics[width=0.4\linewidth]{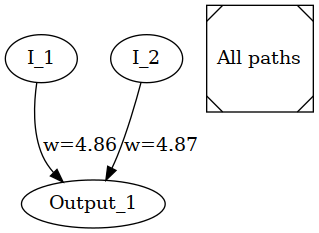} 
  \caption{Global explanation of the ISLaB-LRT network trained on the linear problem. The $w$ indicates the mean of the probability distribution of the weights.}\label{fig:lrtWeightsGlobal}
\end{figure}

In the first experiment under consideration, an ISLaB-LRT model with ReLU activations will be trained on the linear problem presented in Section \ref{sec:SimulatedLinearData}. The model will have 4 hidden layers with 20 hidden nodes per layer. The global explanation of the network after training is displayed in Figure \ref{fig:lrtWeightsGlobal}, and it attained an accuracy of 99.9\%.

\begin{figure}[htp]
\centering
\begin{subfigure}{.33\textwidth}
  \centering
  \includegraphics[width=.99\linewidth]{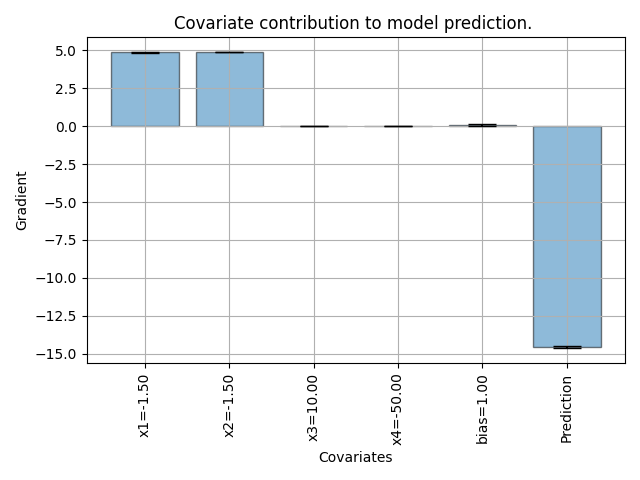}
  \caption{True class: 0}
  \label{fig:gradLinear0}
\end{subfigure}%
\begin{subfigure}{.33\textwidth}
  \centering
  \includegraphics[width=.99\linewidth]{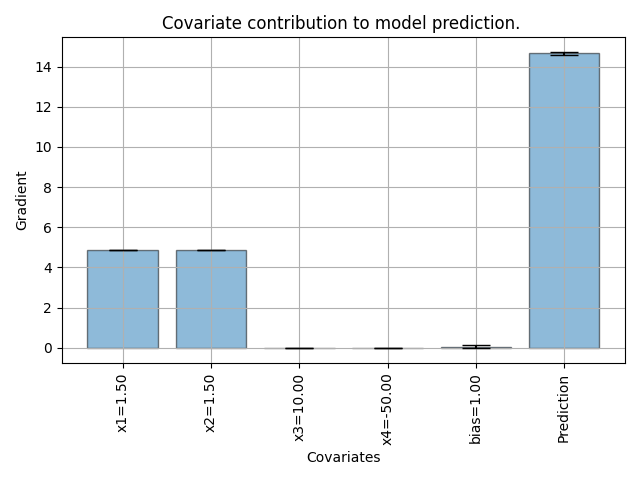}
  \caption{True class: 1}
  \label{fig:gradLinear1}
\end{subfigure}
\begin{subfigure}{.33\textwidth}
  \centering
  \includegraphics[width=.99\linewidth]{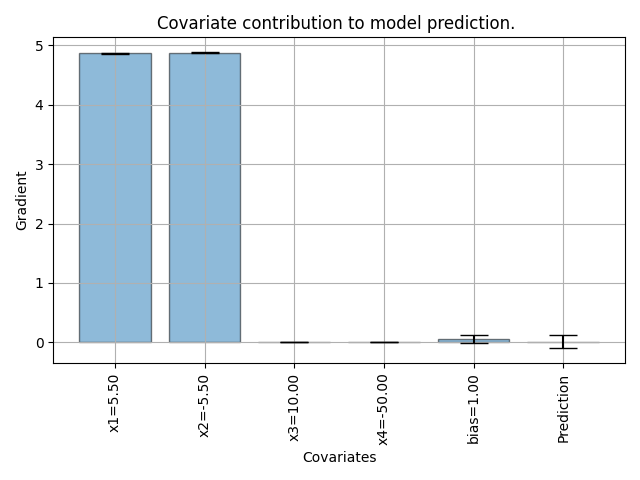}
  \caption{True class: 1}
  \label{fig:gradLinear0Uncertain}
\end{subfigure}
\caption{Explainable slope coefficients based on predictions made by the ISLaB-LRT model on the linear problem. The mean impact is displayed as a blue bin in the histogram plot, while the 95\% credibility interval is illustrated as error bars on top of each bin.}
\label{fig:linearGradExample}
\end{figure}

Figure \ref{fig:linearGradExample} shows the local explanations, in terms of the explainable slope coefficients, of three observations using the approach delineated in Section \ref{sec:local_explain}. As can be seen in the figures, it is clear that we retrieved the same slope coefficients for all three explanations, regardless of the final prediction. This indicates that the model has found a linear relationship between the input covariates and the output nodes as the magnitude of the coefficients does not change for different inputs. We can also note that the reason for the uncertain prediction in Figure \ref{fig:gradLinear0Uncertain} is due to the input covariates contributing with about the same magnitude to the prediction, but with different signs.

\subsubsection{Abalone Dataset}

\begin{figure}[htp]
  \centering
    \includegraphics[width=0.99\linewidth]{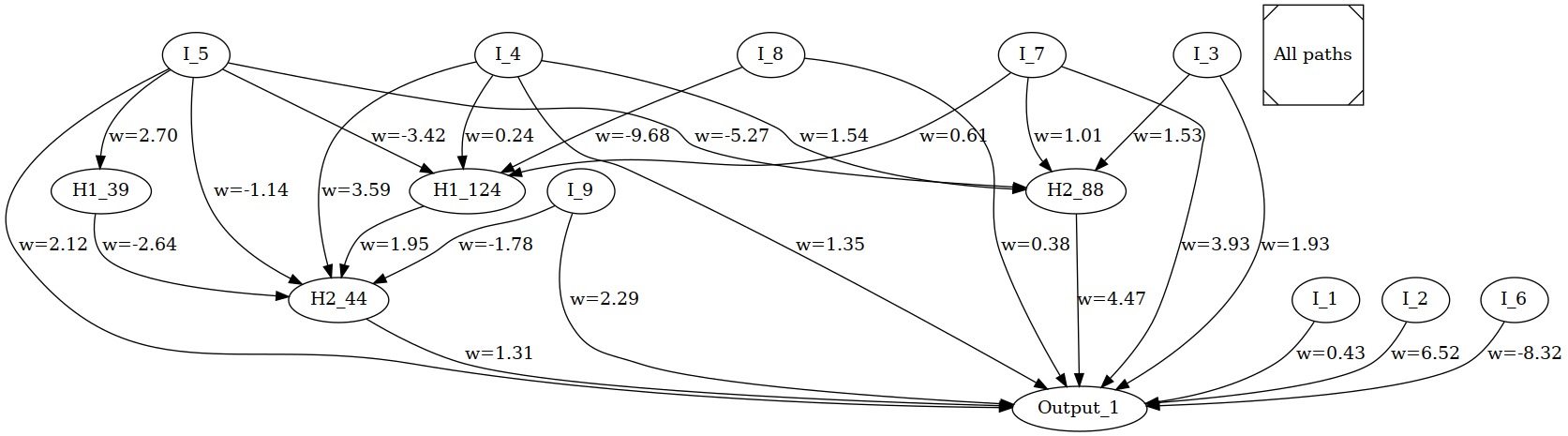} 
  \caption{Global explanation of the ISLaB-LRT network trained on the abalone dataset. The $w$ indicates the mean of the probability distribution of the weights.}\label{fig:lrtWeightsGlobalAbalone}
\end{figure}

To obtain local explanations of the predictions of the abalone dataset, a ReLU-activated ISLaB-LRT model will be trained, comprising two hidden layers and 200 hidden nodes per layer. The global explanation of the trained network is found in Figure \ref{fig:lrtWeightsGlobalAbalone}, and it attained an RMSE of 2.07 for the median probability model. 

\begin{figure}[htp]

\centering
\begin{subfigure}{.33\textwidth}
  \centering
  \includegraphics[width=.99\linewidth]{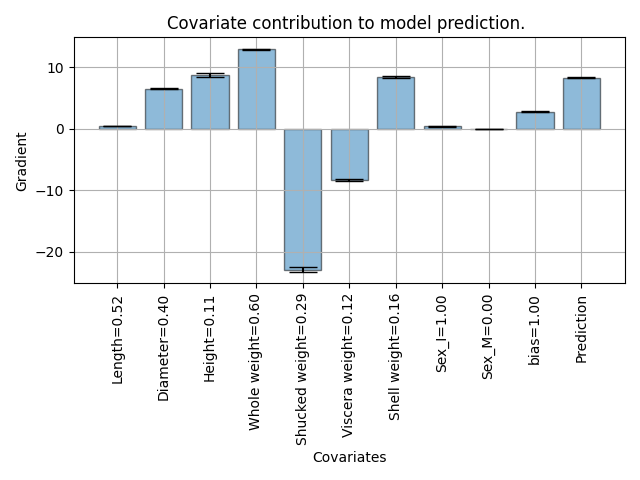}
  \caption{True value is 8}
  \label{fig:GradRegAbalone8}
\end{subfigure}%
\begin{subfigure}{.33\textwidth}
  \centering
  \includegraphics[width=.99\linewidth]{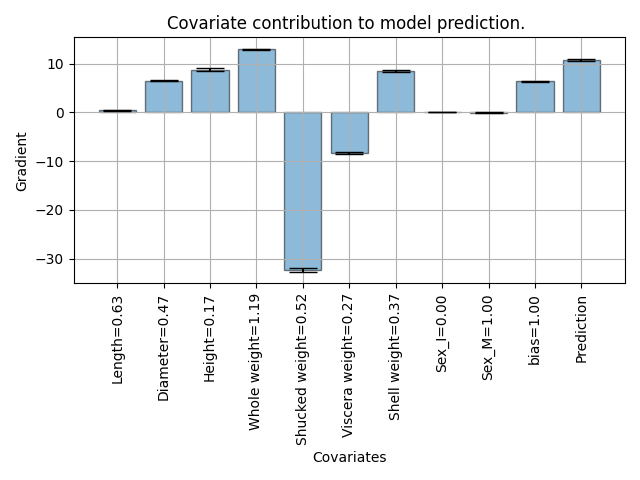}
  \caption{True value is 10}
  \label{fig:GradRegAbalone10}
\end{subfigure}
\begin{subfigure}{.33\textwidth}
  \centering
  \includegraphics[width=.99\linewidth]{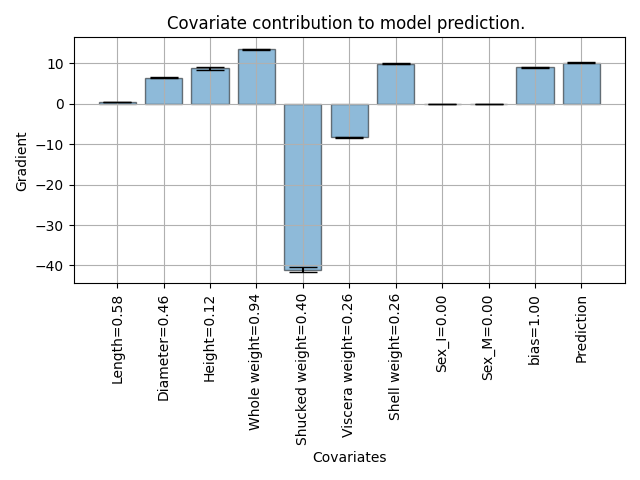}
  \caption{True value is 11}
  \label{fig:GradRegAbalone11}
\end{subfigure}
\captionsetup{justification=raggedright,singlelinecheck=false}
\caption{ISLaB-LRT model with gradient explanations of abalone predictions.}
\label{fig:abaloneGradExamples}
\end{figure}

The local explanations of observations with 8, 10, and 11 rings are shown in Figure \ref{fig:abaloneGradExamples}. Since this is a regression problem, the covariate contributions can be interpreted more directly, as the magnitude indicates how much each covariate influences the regression model to give its ring count prediction. A notable observation, which does not appear in the previous linear example, is that the selected covariates contributions vary across the problems presented. For instance, the slope coefficients for the \textit{Shucked weight} covariate are different for all three examples, which indicates that different active paths have been activated.

The bias contribution to the prediction, illustrated in Figure \ref{fig:abaloneGradExamples}, is more pronounced than that observed in the previous classification problem, shown in Figure \ref{fig:linearGradExample}. This observation is reasonable as the trained model for the abalone dataset is non-linear, which will make the bias vary more. Also, since the previous problem was a linear classification problem, only the input covariates needed to be considered as the bias will remain constant for all inputs.

\subsubsection{MNIST Dataset}

For the MNIST dataset, a ReLU-based ISLaB-LRT network will be initiated with two hidden layers and 600 hidden nodes per layer. The trained network contains 1'115 weights in the median probability model and obtained an accuracy of 95.47\%. 

\begin{figure}[htp]
\centering
\begin{subfigure}{.45\linewidth}
  \centering
  \includegraphics[width=\linewidth]{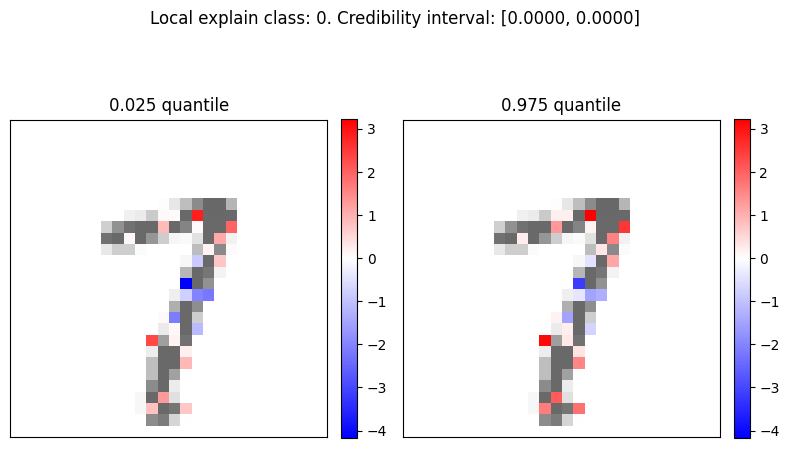}
  \caption{Explanation of 0.}
  \label{fig:7as0MNIST}
\end{subfigure}%
\begin{subfigure}{.45\linewidth}
  \centering
  \includegraphics[width=\linewidth]{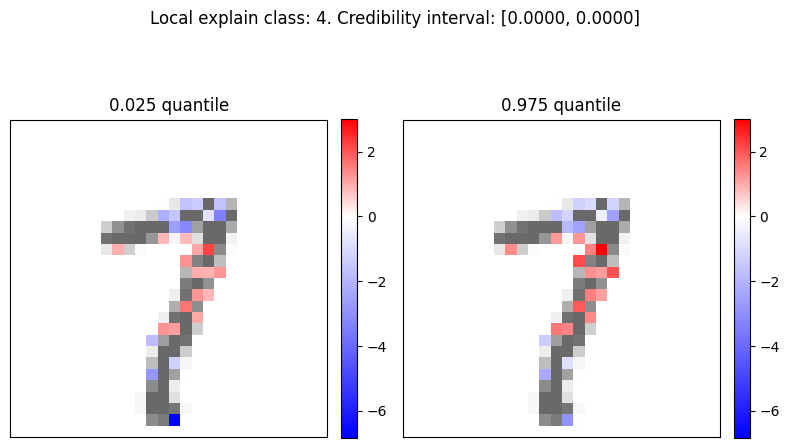}
  \caption{Explanation of 4.}
  \label{fig:7as4MNIST}
\end{subfigure}
\begin{subfigure}{.45\linewidth}
  \centering
  \includegraphics[width=\linewidth]{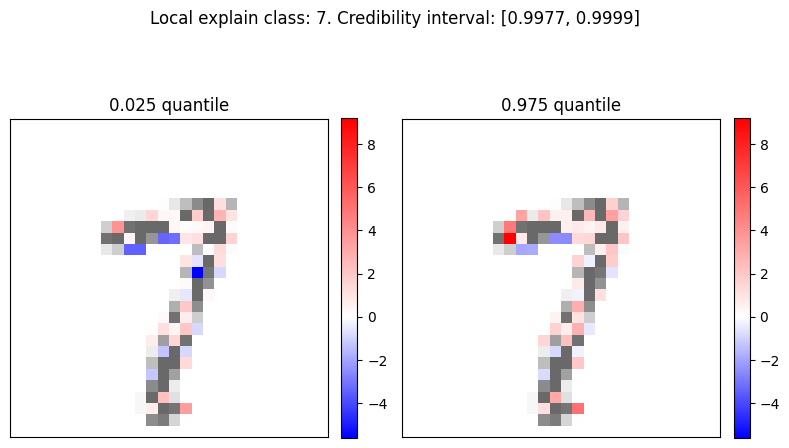}
  \caption{Explanation of 7.}
  \label{fig:7as7MNIST}
\end{subfigure}
\caption{ISLaB-LRT method with gradient explanations of a digit. The true class is seven and the credibility interval is the [0.025, 0.975] quantiles for the prediction probability.}
\label{fig:explainations7}
\end{figure}

\begin{figure}[htp]
\centering
\begin{subfigure}{.45\linewidth}
  \centering
  \includegraphics[width=\linewidth]{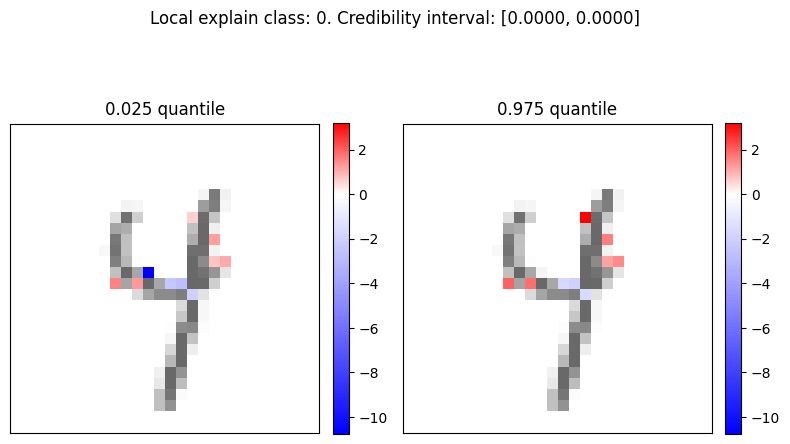}
  \caption{Explanation of 0.}
  \label{fig:4as0MNIST}
\end{subfigure}%
\begin{subfigure}{.45\linewidth}
  \centering
  \includegraphics[width=\linewidth]{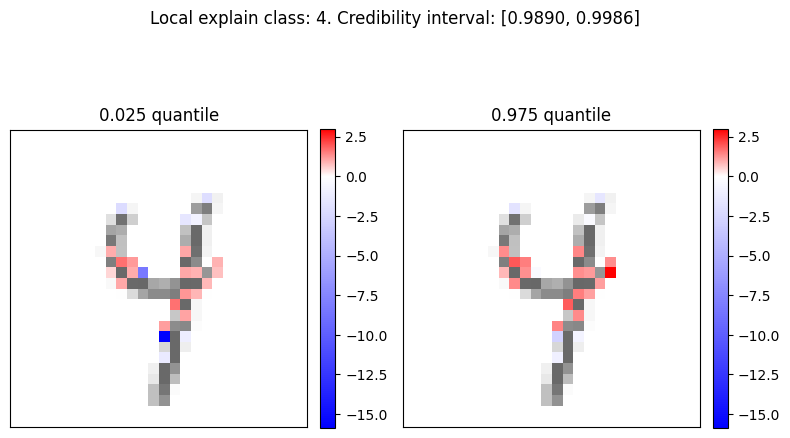}
  \caption{Explanation of 4.}
  \label{fig:4as4MNIST}
\end{subfigure}
\begin{subfigure}{.45\linewidth}
  \centering
  \includegraphics[width=\linewidth]{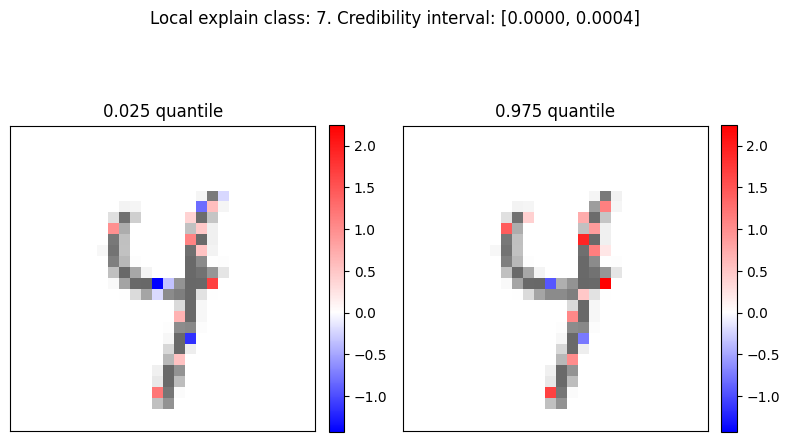}
  \caption{Explanation of 7.}
  \label{fig:4as7MNIST}
\end{subfigure}
\caption{ISLaB-LRT method with gradient explanations of a digit. The true class is four and the credibility interval is the [0.025, 0.975] quantiles for the prediction probability.}
\label{fig:explaination4}
\end{figure}

Local explanations of a 4-digit and a 7-digit are displayed in Figure \ref{fig:explainations7} and \ref{fig:explaination4}, where the contributions for class 0, 4, and 7 are explained in sub-figures (a), (b) and (c), respectively. For instance, Figure \ref{fig:7as4MNIST} presents the local explanations for predicting a 4 when a 7 digit is sent through the network. As can be seen in the top part of that image, present pixels give negative contributions, as these are regions where pixels are not typically observed when predicting for the 4 class. On the other hand, for the local explanation of predicting the 7 as a 7 in Figure \ref{fig:7as7MNIST}, this part contributes positively as these are areas where pixels are usually observed for sevens.

For the local explanation of class 0, both Figure \ref{fig:7as0MNIST} and Figure \ref{fig:4as0MNIST} show that having pixels around the middle of the image contributes negatively. This seems reasonable as the observed zeros in the training dataset do not usually have any pixels in this region as they are centered. Hence, when pixels are in the center of the image, the model will push the prediction toward not belonging to the zero class.

\newpage
\section{Discussion}

This work has predominantly focused on latent binary Bayesian neural networks (LBBNNs) with input skip, referred to as ISLaB, and has demonstrated how a sparser representation of the network can provide both global and local explanations of predictions. This is mainly accomplished through the concept of active paths, where a trained network is evaluated solely on weights in active paths. The evaluation of networks through active paths and the allowance for entire layers to be deemed redundant has proven to yield simpler structures than previous implementations. For instance, the reported density for the median probability model of an LBBNN-LRT network trained on the MNIST dataset was found to be around 10\% in \citet{skaaret2023sparsifying}, whereas this work demonstrates that the density can be further reduced to about 0.09\% if the ISLaB method is employed and only weights in active paths are considered.

A sparser representation, in conjunction with the model's ability to deem layers as redundant, can lead to more interpretable models globally as it becomes easier to see which input covariates are used, how many non-linear activations are employed, and how the input covariates interact when contributing to the output. Moreover, as local explanations can be obtained through the use of active paths and piecewise-linear activation functions, information about the covariate contribution can be acquired from each specific prediction. In the context of a Bayesian neural network, it is also possible to derive a full distribution for each covariate local contribution. These explanations differ from those obtained from post-hoc models as both local and global explanations are obtained directly from the predictive model. This is advantageous as it ensures that the obtained explanations are derived from the predictive model. It should be mentioned that local explanations 
describe how covariates contribute to the predictive node, not how covariates interact and affect each other within the network. It is therefore advantageous to evaluate the local explanations in the context of the global explanation. 
In terms of the predictive metrics, our proposed ISLaB methods demonstrate decent and consistent performance across multiple datasets, achieving a very good balance between predictive accuracy, model sparsity, interpretability, and uncertainty calibration, while maintaining a manageable training cost (see Appendix \ref{appendix:CompCost}). 

Furthermore, pre-trained models and other architectures can also be used to enhance ISLaB performance. We extend ISLaB to  BCNN in the simple vision applications. Moreover, Appendix \ref{appendix:cifar10exp} shows that one can further significantly outperform BCNN models by utilizing a hybrid model between ISLaB and a pre-trained Vision Transformer (ViT) on the CIFAR-10 dataset. However, the active paths framework, which gives ISLaB methods the ability to provide local and global explanations, is inherently designed for fully connected layers. Consequently, when extending ISLaB with convolutional or transformer-based architectures, the interpretability advantage is lost, as active paths cannot be applied directly. Future work should therefore investigate whether the active paths framework can be generalized to support convolutional and transformer models without compromising its explainability capabilities.

\subsection{Initialization and Optimization}\label{sec:initializationAndOptimization}

The initialization of inclusion probabilities in ISLaB plays a crucial role in shaping its learning dynamics and structural sparsification. A dense initialization, where most connections start active (e.g., probabilities close to 1), mimics a backward selection strategy \citep{hastie2009elements, sutter1993comparison, burdon2004forwards}, allowing the model to refine its structure by progressively pruning unnecessary paths during training, often resulting in better predictive power but higher density of the model. Conversely, sparse initialization (e.g., probabilities near 0) is more akin to forward selection \citep{hastie2009elements, sutter1993comparison, burdon2004forwards}, where only the most critical paths are activated incrementally, often resulting in lighter models but with somewhat lower predictive power. Experimental results using different initial inclusion probability of the abalone dataset can be found in Appendix \ref{appendix:abalon_diff_init}. In general, sensitivity to initialization is consistent with previous findings that standard optimization methods such as SGD and Adam may struggle with neural network optimization in the absence of well-informed starting points \citep{goodfellow2016deep, sutskever2013importance}. Poor initialization can lead to suboptimal minima \citep{choromanska2015loss}, underscoring the importance of strategically setting prior inclusion probabilities. On one hand, this challenge extends beyond LBBNNs and ISLaB to all neural networks—efficient convergence requires initialization schemes that align with both the structure and task \citep{glorot2010understanding}, but on the other hand motivates future work on customized optimization designed specifically for LBBNNs and ISLaB. Such future work could explore adaptive initialization strategies, where inclusion probabilities are aggressively adjusted based on early training dynamics, or mode jumping optimization strategies like in mode jumping MCMCs \citep{hubin2018mode, tjelmeland2001mode} allowing for multiple restarts guided by reversible Markov Chain to further improve model convergence under any initialization. 

\subsection{Theoretical Properties of Active Paths}
While our current work has primarily focused on the empirical utility of active paths for sparsity and explainability, an important direction for future research is to strengthen the theoretical underpinnings of this framework. In particular, analyzing the \emph{convergence, consistency, and generalization} properties of active paths and the induced local and global explanations could provide formal guarantees that complement the empirical evidence presented here. For instance, using concentration inequalities \citep{wainwright2019high}, stability analyses \citep{bousquet2002stability}, or PAC-Bayesian bounds \citep{mcallester1999pac,guedj2019primer}, one could study conditions under which the active path selection converges to a stable subset of weights, or whether the local linear explanations are consistent estimators of the underlying data-generating process. Similarly, exploring generalization bounds \citep{dziugaite2017computing,jiang2020fantastic} that link sparsity, uncertainty calibration, predictive performance, and interpretability  could help clarify why active paths offer advantages over post-hoc methods. 

The visualization of active paths provided in this work may not work as well in more complex and deeper models. In future work, we aim to create an R package that will contain additional summary information such as how many times each input variable was included from each layer, and the average inclusion probabilities.

\subsection{Similar Work}

The work presented in this paper is an extension of the framework already created by \citet{hubin2024sparse} and \citet{skaaret2023sparsifying}. In the LBBNN framework, both weight and structural uncertainty are incorporated into the model, which provides a measure of predictive uncertainty and allows for sparser representations of the network to be formed, respectively. Although this work reports sparser representations and densities compared to those found in earlier studies, it does not necessarily imply a noteworthy difference in the methods. Rather, it underscores the importance of active paths, as this ensures that only weights used for predictions are evaluated. However, an important distinction in this work is the introduction of input-skip, which allows active paths to be formed much closer to the output. This could potentially render earlier layers redundant, thereby giving sparser representations than what had been achievable with an LBBNN that is initialized with the same amount of layers and hidden nodes. Compared to other methods, ISLaB does not require post-hoc temperature scaling, as calibration is inherently handled by the probabilistic structure of the LBBNN. Sparsification is intrinsic to the model through the latent binary variables, eliminating the need for ad-hoc pruning. Finally, the input-skip architecture and the concept of active paths provide explainability without relying on post-hoc tools such as SHAP or LIME.

In the finalization of this work, we became aware of other methods that bear resemblance to the active path framework presented in this paper, such as critical data routing paths \citep{wang2018interpret}, critical pathways \citep{khakzar2021neural}, and activation patterns \citep{raghu2017expressive}. However, none of these methods define active paths through the inclusion and exclusion of weights within a network, and they do not take into account uncertainty. Also, \citet{sudjianto2020unwrapping} proposed a method for obtaining local explanations by unwrapping ReLU-based networks through local linear models and activation patterns. This approach diverges from the work conducted in this paper as active paths are employed to obtain local explanations.

\printbibliography

\appendix

\section{Proofs}\label{app:proofs}
\begin{proof}[Proof of Proposition \ref{prop:localGLM}]
{\textbf{Step} $\mathbf{I}$:} Consider a feedforward neural network with $L$ layers and weights $\mathbf{W} = \{w_{k,p}^{(j)}, w_{0,p}^{(j)}\}_{j=1}^L$, where $w_{k,p}^{(j)} \in \mathbb{R}$ connects node $k \in \{1, \dots, d_{j-1}\}$ in layer $j-1$ to node $p \in \{1, \dots, d_j\}$ in layer $j$, and $w_{0,p}^{(j)} \in \mathbb{R}$ is the bias. The activation is $o^{(j)}(z) = \max(0, z)$ for $j=1,\dots,L-1$, and $o^{(L)} = g^{-1}(z)$. The output is $\zeta(\mathbf{x}) \in \mathbb{R}$, where $y_i \sim \mathcal{F}(\mu_i)$, $g(\mu_i) = \zeta(\mathbf{x}_i)$. For $\mathbf{x} \in \mathbb{R}^v$, the forward pass is: input $a_p^{(0)} = x_p$; for $j=1,\dots,L-1$, pre-activation $z_p^{(j)} = w_{0,p}^{(j)} + \sum_{k=1}^{d_{j-1}} a_k^{(j-1)} w_{k,p}^{(j)}$, activation $a_p^{(j)} = o^{(j)}(z_p^{(j)})$; for $j=L$, $\zeta(\mathbf{x}) = w_{0,1}^{(L)} + \sum_{k=1}^{d_{L-1}} a_k^{(L-1)} w_{k,1}^{(L)}$. In an LBBNN, $z_p^{(j)} = w_{0,p}^{(j)} + \sum_{k=1}^{d_{j-1}} \gamma_{k,p}^{(j)} a_k^{(j-1)} w_{k,p}^{(j)}$. Assume $z_p^{(j)} \neq 0$ almost everywhere. 
\newline
{\textbf{Step} $\mathbf{II}$:} For $\mathbf{x}_i \in \mathbb{R}^v$, define the activation pattern $\sigma_p^{(j)} = 1$ if $z_p^{(j)} > 0$, else 0, for $j=1,\dots,L-1$. Since $o^{(j)}$ is continuous, $\exists \varepsilon > 0$ such that $\sigma_p^{(j)}$ is fixed $\forall \mathbf{x} \in B_\varepsilon(\mathbf{x}_i) := \{ \mathbf{x} : \|\mathbf{x} - \mathbf{x}_i\|_2 < \varepsilon \}$. Let $\mathbf{D}^{(j)} = \operatorname{diag}(\sigma_1^{(j)}, \dots, \sigma_{d_j}^{(j)})$. For $j=1,\dots,L-1$, $a_p^{(j)} = \sigma_p^{(j)} z_p^{(j)}$, making the layer affine. 
\newline
{\textbf{Step} $\mathbf{III}$:} 
Composing affine maps, define $a_p^{(j)} = \sum_{q=1}^v m_{p,q}^{(j)} x_q + c_p^{(j)}$, where for $j=1$, $m_{p,q}^{(1)} = \sigma_p^{(1)} w_{q,p}^{(1)}$, $c_p^{(1)} = \sigma_p^{(1)} w_{0,p}^{(1)}$; for $j=2,\dots,L-1$, recursively, $m_{p,q}^{(j)} = \sigma_p^{(j)} \sum_{k=1}^{d_{j-1}} w_{k,p}^{(j)} m_{k,q}^{(j-1)}$, $c_p^{(j)} = \sigma_p^{(j)} \left( w_{0,p}^{(j)} + \sum_{k=1}^{d_{j-1}} w_{k,p}^{(j)} c_k^{(j-1)} \right)$. In an LBBNN, $w_{k,p}^{(j)}$ is replaced by $\gamma_{k,p}^{*(j)} w_{k,p}^{(j)}$. Thus, $a_p^{(L-1)} = \sum_{q=1}^v m_{p,q}^{(L-1)} x_q + c_p^{(L-1)}$, and we get $\zeta(\mathbf{x}) = w_{0,1}^{(L)} + \sum_{k=1}^{d_{L-1}} \left( \sum_{q=1}^v m_{k,q}^{(L-1)} x_q + c_k^{(L-1)} \right) w_{k,1}^{(L)} = \beta_0 + \sum_{q=1}^v \beta_q x_q$, where $\beta_q = \sum_{k=1}^{d_{L-1}} w_{k,1}^{(L)} m_{k,q}^{(L-1)}$, $\beta_0 = w_{0,1}^{(L)} + \sum_{k=1}^{d_{L-1}} w_{k,1}^{(L)} c_k^{(L-1)}$. Thus, $\forall \mathbf{x} \in B_\varepsilon(\mathbf{x}_i)$, $\zeta(\mathbf{x})$ is a GLM linear predictor.
\end{proof}
\begin{proof}[Proof of Remark \ref{rem:lbnn}]
{\textbf{Step} $\mathbf{I}$:} For $\mathbf{x}_i \in \mathbb{R}^v$ and any $\mathbf{W} \sim \pi(\mathbf{W} \mid \mathbf{\Gamma}^*, D)$, Proposition \ref{prop:localGLM} gives $\zeta(\mathbf{x}) = \beta_0 + \sum_{j=1}^v \beta_j x_j$ in $B_\varepsilon(\mathbf{x}_i) := \{ \mathbf{x} : \|\mathbf{x} - \mathbf{x}_i\|_2 < \varepsilon \}$, where $\beta_j = \sum_{p=1}^{d_{L-1}} w_{p,1}^{(L)} m_{p,j}^{(L-1)}$, $\beta_0 = w_{0,1}^{(L)} + \sum_{p=1}^{d_{L-1}} w_{p,1}^{(L)} c_p^{(L-1)}$, and $m_{p,q}^{(j)}, c_p^{(j)}$ are defined as in Proposition \ref{prop:localGLM} with $w_{k,p}^{(j)}$ replaced by $\gamma_{k,p}^{*(j)}  w_{k,p}^{(j)}$. \newline{\textbf{Step} $\mathbf{II}$:} The explanation coefficients $\boldsymbol{\beta}^*, \beta_0^*$ are computed using $\E(\mathbf{W} \mid \mathbf{\Gamma}^*, D)$ with a fixed activation pattern $\sigma_p^{(j)} = 1$ if $z_p^{(j)} = w_{0,p}^{(j)} + \sum_{k=1}^{d_{j-1}} \gamma_{k,p}^{*(j)} a_k^{(j-1)} w_{k,p}^{(j)} > 0$, else 0. \newline{\textbf{Step} $\mathbf{III}$:} Sampling $\mathbf{W} \sim \pi(\mathbf{W} \mid \mathbf{\Gamma}^*, D)$ generates a distribution over $\boldsymbol{\beta}, \beta_0$ via varying weights and possibly $\sigma_p^{(j)}$, with credible intervals from sample quantiles.
\end{proof}
\begin{proof}[Proof of Corollary \ref{coro:GradientExplain}]{\textbf{Step} $\mathbf{I}$:}
For $\mathbf{x}_i \in \mathbb{R}^v$, Proposition \ref{prop:localGLM} gives $\zeta(\mathbf{x}) = \beta_0 + \sum_{j=1}^v \beta_j x_j$ in $B_\varepsilon(\mathbf{x}_i) := \{ \mathbf{x} : \|\mathbf{x} - \mathbf{x}_i\|_2 < \varepsilon \}$, with $\sigma_p^{(j)} = 1$ if $z_p^{(j)} = w_{0,p}^{(j)} + \sum_{k=1}^{d_{j-1}} a_k^{(j-1)} w_{k,p}^{(j)} > 0$, else 0, fixed in $B_\varepsilon(\mathbf{x}_i)$. \newline{\textbf{Step} $\mathbf{II}$:} The gradient is $\nabla_{\mathbf{x}} \zeta(\mathbf{x}) = (\beta_1, \dots, \beta_v) = \boldsymbol{\beta}$, $\forall \mathbf{x} \in B_\varepsilon(\mathbf{x}_i)$.
\end{proof}
\begin{proof}[Proof of Corollary \ref{coro:LBBNNGradient}]
{\textbf{Step} $\mathbf{I}$:} For any $\mathbf{W} \sim \pi(\mathbf{W} \mid \mathbf{\Gamma}^*, D)$, $\exists \varepsilon > 0$ such that $\forall \mathbf{x} \in B_\varepsilon(\mathbf{x}_i) := \{ \mathbf{x} : \|\mathbf{x} - \mathbf{x}_i\|_2 < \varepsilon \}$, Proposition \ref{prop:localGLM} gives $\zeta(\mathbf{x}) = \beta_0 + \sum_{j=1}^v \beta_j x_j$, where $\beta_j = \sum_{p=1}^{d_{L-1}} w_{p,1}^{(L)} m_{p,j}^{(L-1)}$, $\beta_0 = w_{0,1}^{(L)} + \sum_{p=1}^{d_{L-1}} w_{p,1}^{(L)} c_p^{(L-1)}$, and $m_{p,q}^{(j)}, c_p^{(j)}$ are defined as in Proposition \ref{prop:localGLM} with $w_{k,p}^{(j)}$ replaced by $\gamma_{k,p}^{*(j)}  w_{k,p}^{(j)}$. \newline{\textbf{Step} $\mathbf{II}$:} The gradient is $\nabla_{\mathbf{x}} \zeta(\mathbf{x}) = \boldsymbol{\beta}$, $\forall \mathbf{x} \in B_\varepsilon(\mathbf{x}_i)$. The explanation coefficients $\boldsymbol{\beta}^*, \beta_0^*$ are computed using $\E(\mathbf{W} \mid \mathbf{\Gamma}^*, D)$. \newline{\textbf{Step} $\mathbf{III}$:} Sampling $\mathbf{W} \sim \pi(\mathbf{W} \mid \mathbf{\Gamma}^*, D)$ generates a distribution over $\boldsymbol{\beta}$ via varying weights and possibly $\sigma_p^{(j)}$, with credible intervals from sample quantiles.
\end{proof}
\begin{proof}[Proof of Corollary \ref{cor:limeFixed}]
{\textbf{Step} $\mathbf{I}$:} Linear LIME fits a GLM ${\zeta}(\mathbf{x}) = \beta_0 + \sum_{j=1}^v \beta_j x_j$ to samples $\{\mathbf{x}_k \sim \pi, \tilde{y}_k = \zeta(\mathbf{x}_k)\}_{k=1}^n$ in $B_\varepsilon(\mathbf{x}_i) := \{ \mathbf{x} : \|\mathbf{x} - \mathbf{x}_i\|_2 < \varepsilon \}$, where $\pi$ has positive density. The explanations $\boldsymbol{\beta}^*, \hat{\beta}_0$ are obtained by maximizing the log likelihood $\ell(\boldsymbol{\beta}, \beta_0) = \sum_{k=1}^n \log p(\tilde{y}_k \mid {\zeta}(\mathbf{x}_k))$. \newline{\textbf{Step} $\mathbf{II}$:} By Proposition \ref{prop:localGLM}, $\exists \varepsilon > 0$ such that $\forall \mathbf{x}_k \in B_\varepsilon(\mathbf{x}_i)$, $\zeta(\mathbf{x}_k) = \beta_0^* + \sum_{j=1}^v \beta_j^* x_{kj}$, where $\boldsymbol{\beta}^*, \beta_0^*$ depend on $\mathbf{W}$, and $\sigma_p^{(j)} = 1$ if $z_p^{(j)} = w_{0,p}^{(j)} + \sum_{k=1}^{d_{j-1}} a_k^{(j-1)} w_{k,p}^{(j)} > 0$, else 0. \newline{\textbf{Step} $\mathbf{III}$:} Under regularity conditions ($\pi$ positive, design matrix full rank, identifiable parameters), the MLE satisfies $\hat{\boldsymbol{\beta}} \overset{p}{\to} \boldsymbol{\beta}^*, \hat{\beta}_0 \overset{p}{\to} \beta_0^*$ as $n \to \infty$.
\end{proof}
\begin{proof}[Proof of Corollary \ref{cor:limeLBBNN}]
{\textbf{Step} $\mathbf{I}$:} For any $\mathbf{W} \sim \pi(\mathbf{W} \mid \mathbf{\Gamma}^*, D)$, $\exists \varepsilon > 0$ such that $\forall \mathbf{x} \in B_\varepsilon(\mathbf{x}_i) := \{ \mathbf{x} : \|\mathbf{x} - \mathbf{x}_i\|_2 < \varepsilon \}$, Corollary \ref{cor:limeFixed} gives LIME coefficients $\hat{\boldsymbol{\beta}} \overset{p}{\to} \boldsymbol{\beta}$, $\hat{\beta}_0 \overset{p}{\to} \beta_0$ as $n \to \infty$, where $\beta_j = \sum_{p=1}^{d_{L-1}} w_{p,1}^{(L)} m_{p,j}^{(L-1)}$, $\beta_0 = w_{0,1}^{(L)} + \sum_{p=1}^{d_{L-1}} w_{p,1}^{(L)} c_p^{(L-1)}$, and $m_{p,q}^{(j)}, c_p^{(j)}$ are defined as in Proposition \ref{prop:localGLM} with $w_{k,p}^{(j)}$ replaced by $\gamma_{k,p}^{*(j)} w_{k,p}^{(j)}$. \newline{\textbf{Step} $\mathbf{II}$:} The LBBNN explanation uses $\boldsymbol{\beta}^*, \beta_0^*$ from $\E(\mathbf{W} \mid \mathbf{\Gamma}^*, D)$. \newline{\textbf{Step} $\mathbf{III}$:} Sampling $\mathbf{W} \sim \pi(\mathbf{W} \mid \mathbf{\Gamma}^*, D)$ generates a distribution over $\boldsymbol{\beta}, \beta_0$ via varying weights and possibly $\sigma_p^{(j)}$, with credible intervals from sample quantiles.
\end{proof}

\begin{proof}[Proof of Corollary \ref{coro:IGExplain}]
{\textbf{Step} $\mathbf{I}$:} For \(\mathbf{x}_i \in \mathbb{R}^v\), Proposition \ref{prop:localGLM} gives \(\zeta(\mathbf{x}) = \beta_0 + \sum_{j=1}^v \beta_j x_j\) in \(B_\varepsilon(\mathbf{x}_i) := \{ \mathbf{x} : \|\mathbf{x} - \mathbf{x}_i\|_2 < \varepsilon \}\), with \(\sigma_p^{(j)} = 1\) if \(z_p^{(j)} = w_{0,p}^{(j)} + \sum_{k=1}^{d_{j-1}} a_k^{(j-1)} w_{k,p}^{(j)} > 0\), else 0, fixed in \(j=1,\dots,L-1\). The gradient is \(\nabla_{\mathbf{x}} \zeta(\mathbf{x}) = (\beta_1, \dots, \beta_v) = \boldsymbol{\beta}\), \(\forall \mathbf{x} \in B_\varepsilon(\mathbf{x}_i)\). \newline{\textbf{Step} $\mathbf{II}$:} For a baseline \(\mathbf{x}' \in \mathbb{R}^v\), assuming the straight-line path \(\mathbf{x}' + \alpha (\mathbf{x}_i - \mathbf{x}') \in B_\varepsilon(\mathbf{x}_i)\) for \(\alpha \in [0,1]\), Integrated Gradients for feature \(j\) is:
\[
\text{IG}_j(\mathbf{x}_i) = (x_{i,j} - x'_j) \int_0^1 \frac{\partial \zeta}{\partial x_j} (\mathbf{x}' + \alpha (\mathbf{x}_i - \mathbf{x}')) \, d\alpha.
\]
\newline{\textbf{Step} $\mathbf{III}$:} Since \(\frac{\partial \zeta}{\partial x_j} = \beta_j\) is constant in \(B_\varepsilon(\mathbf{x}_i)\), the integral is:
\[
\int_0^1 \beta_j \, d\alpha = \beta_j.
\]
Thus, \(\text{IG}_j(\mathbf{x}_i) = (x_{i,j} - x'_j) \beta_j\), providing the local linear coefficients scaled by the input difference.
\end{proof}

\begin{proof}[Proof of Corollary \ref{coro:LBBNNIG}]
{\textbf{Step} $\mathbf{I}$:} For any \(\mathbf{W} \sim \pi(\mathbf{W} \mid \mathbf{\Gamma}^*, D)\), \(\exists \varepsilon > 0\) such that \(\forall \mathbf{x} \in B_\varepsilon(\mathbf{x}_i) := \{ \mathbf{x} : \|\mathbf{x} - \mathbf{x}_i\|_2 < \varepsilon \}\), Proposition \ref{prop:localGLM} gives \(\zeta(\mathbf{x}) = \beta_0 + \sum_{j=1}^v \beta_j x_j\), where \(\beta_j = \sum_{p=1}^{d_{L-1}} w_{p,1}^{(L)} m_{p,j}^{(L-1)}\), \(\beta_0 = w_{0,1}^{(L)} + \sum_{p=1}^{d_{L-1}} w_{p,1}^{(L)} c_p^{(L-1)}\), and \(m_{p,q}^{(j)}, c_p^{(j)}\) are defined as in Proposition \ref{prop:localGLM} with \(w_{k,p}^{(j)}\) replaced by \(\gamma_{k,p}^{*(j)} w_{k,p}^{(j)}\). The gradient is \(\nabla_{\mathbf{x}} \zeta(\mathbf{x}) = \boldsymbol{\beta}\), \(\forall \mathbf{x} \in B_\varepsilon(\mathbf{x}_i)\). \newline{\textbf{Step} $\mathbf{II}$:} For a baseline \(\mathbf{x}' \in \mathbb{R}^v\), assuming \(\mathbf{x}' + \alpha (\mathbf{x}_i - \mathbf{x}') \in B_\varepsilon(\mathbf{x}_i)\) for \(\alpha \in [0,1]\), Integrated Gradients is:
\[
\text{IG}_j(\mathbf{x}_i) = (x_{i,j} - x'_j) \int_0^1 \frac{\partial \zeta}{\partial x_j} (\mathbf{x}' + \alpha (\mathbf{x}_i - \mathbf{x}')) \, d\alpha.
\]
Since \(\frac{\partial \zeta}{\partial x_j} = \beta_j\) in \(B_\varepsilon(\mathbf{x}_i)\), we have \(\text{IG}_j(\mathbf{x}_i) = (x_{i,j} - x'_j) \beta_j\). \newline{\textbf{Step} $\mathbf{III}$:} The explanation coefficients \(\boldsymbol{\beta}^*, \beta_0^*\) from \(\E(\mathbf{W} \mid \mathbf{\Gamma}^*, D)\) give \(\text{IG}_j(\mathbf{x}_i) = (x_{i,j} - x'_j) \beta_j^*\). \newline{\textbf{Step} $\mathbf{IV}$:} Sampling \(\mathbf{W} \sim \pi(\mathbf{W} \mid \mathbf{\Gamma}^*, D)\) generates a distribution over \(\boldsymbol{\beta}\), yielding a distribution over \(\text{IG}_j(\mathbf{x}_i)\) with credible intervals from sample quantiles.
\end{proof}

\section{Hyperparameters and Initialization}\label{appendix:A}

Table \ref{tab:initInclusionProb} shows the initial inclusion probabilities, where $\alpha=1/(1+\exp{(-\lambda}))$, for all weights. Initial values are drawn randomly from uniform distributions, with $\lambda\sim \text{Unif}(\lambda_{min}, \lambda_{max})$. We draw $\lambda$ from the same probability distribution regardless of the position in the network. That is, the same distribution is used for connections in, e.g., the first hidden layer and the last hidden layer.

All hyper- and tuning parameters used in the experiments can be found in Table \ref{tab:paramMat}. 

\begin{table}[!htp]
    \caption{Initial inclusion probability, $\alpha=1/(1+\exp{(-\lambda}))$, for all weights. ``weights'' indicates initialization for connections going from hidden nodes, while ``covariate'' indicate connections coming from a covariate.}
    \small{
    \centering
    \begin{tabular}
    {lcccc}
    \hline
    Experiment & $\lambda_{min}$ weights & $\lambda_{max}$ weights & $\lambda_{min}$ covariates & $\lambda_{max}$ covariates \\
    \hline
        Sim. lin. - LRT        & -10 & -7 & 5  & 5 \\
        Sim. lin. - FLOW       & -10 & -7 & 5  & 5 \\
        Sim. lin. - BLR-LRT    & -   & -  & 0  & 1 \\
        Sim. lin. - BLR-FLOW   & -   & -  & 0  & 1 \\
        \hline
        Sim. nonlin. - LRT      & -5  & -4 & 5  & 5 \\
        Sim. nonlin. - FLOW     & 4   & 6  & 5  & 15 \\
        Sim. nonlin. - BLR-LRT  & -   & -  & 0  & 1 \\
        Sim. nonlin. - BLR-FLOW & -   & -  & 0  & 1 \\
        \hline
        WBC - LRT              & -9  & -5 & 5  & 5 \\
        WBC - FLOW             & -9  & -5 & 5  & 15 \\
        WBC - BLR-LRT          & -   & -  & 0  & 1 \\
        WBC - BLR-FLOW         & -   & -  & 0  & 1 \\
        \hline
        Abalone - LRT          & -9  & -4 & 5  & 5 \\
        Abalone - FLOW         & -9  & -4 & 5  & 5 \\
        Abalone - BLR-LRT      & -   & -  & 0  & 1 \\
        Abalone - BLR-FLOW     & -   & -  & 0  & 1 \\
        \hline
        MiM prot. - LRT        &  10 & 15 & 10  & 15 \\
        MiM prot. - FLOW       &  10 & 15 & 10  & 15 \\
        MiM prot. - BLR-LRT    & -   & -  & 10  & 15 \\
        MiM prot. - BLR-FLOW   & -   & -  & 10  & 15 \\
        \hline
        MNIST - LRT            & 5   & 15 & 5   & 15 \\
        MNIST - FLOW           & 10  & 15 & 10  & 15 \\
        FER2013 - BCNN-LRT     & 5   & 10 & -10 & -10 \\
        FER2013 - BCNN-FLOW    & 5   & 10 & -15 & -15 \\
        MNIST - BLR-LRT        & -   & -  & 0   & 1 \\
        MNIST - BLR-FLOW       & -   & -  & 0   & 1 \\
        \hline
        FER2013 - LRT          & -8  & -6 & -8  & -6 \\
        FER2013 - FLOW         & -8  & -6 & -8  & -6 \\
        FER2013 - BCNN-LRT     & 3   & 6  & -10 & -10 \\
        FER2013 - BCNN-FLOW    & 3   & 6  & -15 & -15 \\
        FER2013 - BLR-LRT      & -   & -  & 0   & 1 \\
        FER2013 - BLR-FLOW     & -   & -  & 0   & 1 \\
        \hline
        CIFAR10 - BCNN-LRT     & 10  & 15 & -10 & -10 \\
        CIFAR10 - BCNN-FLOW    & 10  & 15 & -15 & -15 \\
        CIFAR10 - ViT-LRT      & 5   & 8  & 5   & 8   \\
        CIFAR10 - ViT-FLOW     & 5   & 8  & 5   & 8   \\
        \hline
    \end{tabular}
    }
    \label{tab:initInclusionProb}
\end{table}

\begin{table}[!htp]
    \caption{Hyperparameters and tuning parameters used in the experiments. $p(w)$ denotes the prior weight distribution, $\psi$ represents the prior inclusion probability for all weights, ``lr.'' is the learning rate,``Epochs'' refers to the number of training epochs. ``Iter. epoch'' is the number of iterations per epoch, corresponding to the number of batches the training dataset is divided into, and ``Lik.'' is the likelihood function used for modeling the responses, with N for normal, B for Bernoulli, and C for Categorical. For the flow-based methods, we use two transformations.}
    \resizebox{\textwidth}{!}{
    \tiny
    \begin{tabular}{p{0.25\linewidth}|p{0.115\linewidth}p{0.05\linewidth}p{0.075\linewidth}p{0.11\linewidth}p{0.1\linewidth}p{0.05\linewidth}}
     \hline
     Experiment              & $p(w)$        & $\psi$ & lr.    & Epochs & Iter. epoch & Lik. \\
     \hline
     Sim. lin. - LRT         & $\N(0,2.5^2)$ & 0.001  & 0.1    & 200    & 50 & B \\
     Sim. lin. - FLOW        & $\N(0,2.5^2)$ & 0.001  & 0.01   & 200    & 10 & B \\
     Sim. lin. - BLR-LRT     & $\N(0,2.5^2)$ & 0.001  & 0.01   & 200    & 50 & B \\
     Sim. lin. - BLR-FLOW    & $\N(0,2.5^2)$ & 0.001  & 0.01   & 200    & 50 & B \\
     Sim. lin. - ANN-L1      & -             & -      & 0.01   & 750    & 50 & B\\
     \hline
     Sim. nonlin. - LRT       & $\N(0,30^2)$  & 0.01   & 0.01   & 750    & 50 & B \\
     Sim. nonlin. - FLOW      & $\N(0,30^2)$  & 0.05   & 0.01   & 750    & 10 & B \\
     Sim. nonlin. - BLR-LRT   & $\N(0,30^2)$  & 0.001  & 0.01   & 200    & 50 & B \\
     Sim. nonlin. - BLR-FLOW  & $\N(0,30^2)$  & 0.001  & 0.01   & 200    & 50 & B \\
     Sim. nonlin. - ANN-L1   & - & -                  & 0.01   & 750    & 50 & B\\
     \hline
     WBC - LRT               & $\N(0,1^2)$   & 0.01   & 0.1    & 200    & 8  & B \\
     WBC - FLOW              & $\N(0,1^2)$   & 0.01   & 0.05   & 200    & 8  & B\\
     WBC - BLR-LRT           & $\N(0,1^2)$   & 0.01   & 0.01   & 200    & 8  & B \\
     WBC - BLR-FLOW          & $\N(0,1^2)$   & 0.01   & 0.01   & 200    & 8  & B \\
     WBC - ANN-L1            & -             & -      & 0.01   & 200    & 8  & B \\
     WBC - BNN-HORSE            & -             & -      & 0.01   & 1000    & 8  & B \\
     WBC - BNN-CONCRETE            & -             & -      & 0.01   & 200    & 8  & B \\
     WBC - Laplace-SpaM           & -             & -      & 0.1   & 200    & 8  & B \\
     \hline
     Abalone - LRT           & $\N(0,25^2)$  & 0.25   & 0.01   & 5,000  & 5  & N \\
     Abalone - FLOW          & $\N(0,25^2)$  & 0.25   & 0.01   & 2,000  & 5  & N \\
     Abalone - BLR-LRT       & $\N(0,25^2)$  & 0.25   & 0.005  & 2,000  & 5  & N \\
     Abalone - BLR-FLOW      & $\N(0,15^2)$  & 0.25   & 0.01   & 2,000  & 5  & N \\
     Abalone - ANN-L1        & -             & -      & 0.01   & 2,000  & 5  & N \\
     Abalone - BNN-HORSE     & -             & -      & 0.01   & 2,000  & 5  & N \\
     Abalone - BNN-CONCRETE  & -             & -      & 0.005   & 2,000  & 5  & N \\
     Abalone - Laplace-SpaM  & -             & -      & 0.01  & 2,000  & 5  & N \\
     \hline
     MiM prot. - LRT         & $\N(0,25^2)$  & 0.05   & 0.01   & 2,000  & 27 & C\\
     MiM prot. - FLOW        & $\N(0,25^2)$  & 0.05   & 0.01   & 500    & 27 & C \\
     MiM prot. - BLR-LRT     & $\N(0,25^2)$  & 0.05   & 0.01   & 500    & 27 & C \\
     MiM prot. - BLR-FLOW    & $\N(0,25^2)$  & 0.05   & 0.01   & 500    & 27 & C \\
     MiM prot. - ANN-L1      & -             & -      & 0.001  & 20,000 & 1  & C \\
     MiM prot. - BNN-HORSE   & -             & -      & 0.001  & 20,000 & 1  & C \\
     MiM prot. - BNN-CONCRETE& -             & -      & 0.001  & 20,000 & 1  & C \\
     MiM prot. - Laplace-SpaM& -             & -      & 0.01  & 500 & 27  & C \\
     \hline
     MNIST - LRT             & $\N(0,15^2)$  & 0.01   & 0.01   & 1,000  & 50  & C \\
     MNIST - FLOW            & $\N(0,10^2)$  & 0.01   & 0.005  & 2,000  & 50  & C \\
     MNIST - BCNN-LRT        & $\N(0,1^2)$   & 0.1    & 0.005  & 250    & 600 & C \\
     MNIST - BCNN-FLOW       & $\N(0,1^2)$   & 0.1    & 0.01   & 100    & 600 & C \\
     MNIST - BLR-LRT         & $\N(0,15^2)$  & 0.1    & 0.01   & 250    & 50  & C \\
     MNIST - BLR-FLOW        & $\N(0,10^2)$  & 0.1    & 0.005  & 250    & 50  & C \\
     MNIST - ANN-L1          & -             & -      & 0.01   & 250    & 50  & C \\
     MNIST - BNN-HORSE       & -             & -      & 0.001  & 250    & 600 & C \\
     MNIST - BNN-CONCRETE    & -             & -      & 0.001  & 250    & 600 & C \\
     MNIST - Laplace-SpaM    & -             & -      & 0.001  & 250    & 50  & C \\
     \hline
     FER2013 - LRT           &  $\N(0,30^2)$ & 0.05   & 0.005  & 10,000 & 10 & C \\
     FER2013 - FLOW          &  $\N(0,30^2)$ & 0.1    & 0.0025 & 10,000 & 10 & C \\
     FER2013 - BCNN-LRT      &  $\N(0,30^2)$ & 0.05   & 0.001  & 10,000 & 10 & C \\
     FER2013 - BCNN-FLOW     &  $\N(0,30^2)$ & 0.3    & 0.001  & 10,000 & 10 & C \\
     FER2013 - BLR-LRT       &  $\N(0,30^2)$ & 0.05   & 0.005  & 1,000  & 10 & C \\
     FER2013 - BLR-FLOW      &  $\N(0,30^2)$ & 0.1    & 0.0025 & 1,000  & 10 & C \\
     FER2013 - ANN-L1        & -             & -      & 0.005  & 1,000  & 10 & C \\
     FER2013 - BNN-HORSE     & -             & -      & 0.005  & 1,000  & 10 & C \\
     FER2013 - BNN-CONCRETE  & -             & -      & 0.005  & 1,000  & 10 & C \\
     FER2013 - Laplace-SpaM  & -             & -      & 0.001  & 1,000  & 10 & C \\
     \hline
     CIFAR10 - BCNN-LRT      & $\N(0,1^2)$   & 0.1    & 0.001  & 250    & 50 & C \\
     CIFAR10 - BCNN-FLOW     & $\N(0,1^2)$   & 0.1    & 0.0001 & 100    & 50 & C \\
     CIFAR10 - ViT-LRT       & $\N(0,1^2)$   & 0.1    & 0.001  & 100    & 50 & C \\
     CIFAR10 - ViT-FLOW      & $\N(0,1^2)$   & 0.1    & 0.001  & 100    & 50 & C \\
     \hline
    \end{tabular}
}
\label{tab:paramMat}
\end{table}

\section{Further Detail for the Mice Data}

The classes, explanation, and content of each class in the mice protein dataset are provided in Table~\ref{tab:miceClassDesc}. Explanations are retrieved from \citet{mice_protein_expression_342}.

\begin{table}[htp]
    \caption{Class descriptions of the mice protein dataset.}
    \small{
    \centering
    \begin{tabular}{p{0.4\linewidth}p{0.6\linewidth}}
     \hline
     Class name (class nr.) & Description \\
     \hline
     c-CS-s (1) & control mice, stimulated to learn, injected with saline \\
     c-CS-m (2) & control mice, stimulated to learn, injected with memantine \\
     c-SC-s (3) & control mice, not stimulated to learn, injected with saline \\
     c-SC-m (4) & control mice, not stimulated to learn, injected with memantine \\
     t-CS-s (5) & trisomy mice, stimulated to learn, injected with saline \\
     t-CS-m (6) & trisomy mice, stimulated to learn, injected with memantine \\
     t-SC-s (7) & trisomy mice, not stimulated to learn, injected with saline \\
     t-SC-m (8) & trisomy mice, not stimulated to learn, injected with memantine \\
     \hline
    \end{tabular}
}
\label{tab:miceClassDesc}
\end{table}

\section{Different Initial $\alpha$-values}\label{appendix:abalon_diff_init}

\begin{figure}[htp]
\centering
\begin{subfigure}{.33\textwidth}
  \centering
  
  \includegraphics[width=.99\linewidth]{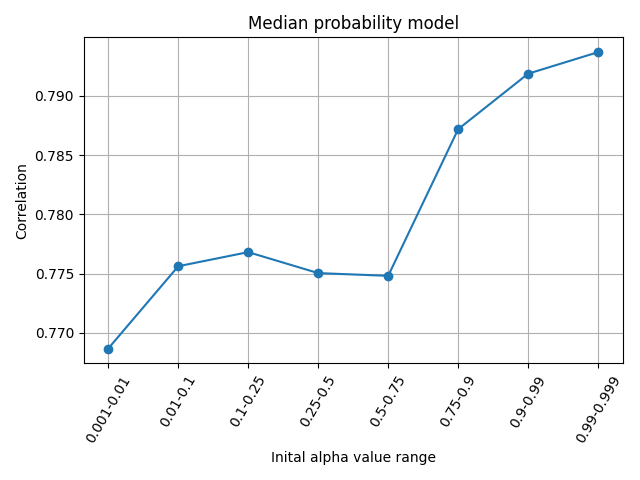}
  \caption{Correlation}
  \label{fig:correlation_mpm_lrt}
\end{subfigure}%
\begin{subfigure}{.33\textwidth}
  \centering
  \includegraphics[width=.99\linewidth]{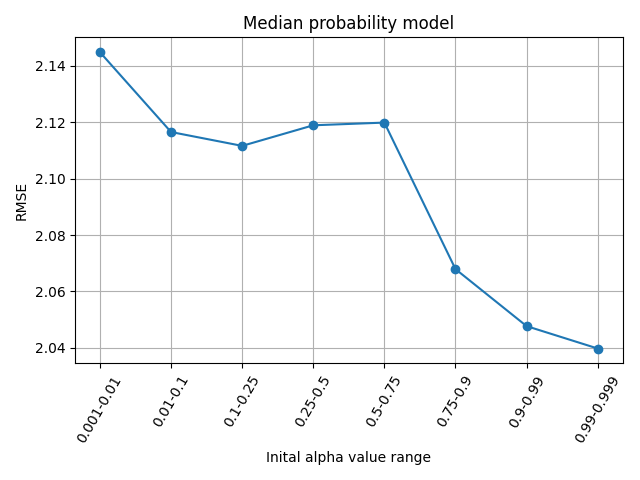}
  \caption{RMSE}
  \label{fig:rmse_mpm_lrt}
\end{subfigure}
\begin{subfigure}{.33\textwidth}
  \centering
  \includegraphics[width=.99\linewidth]{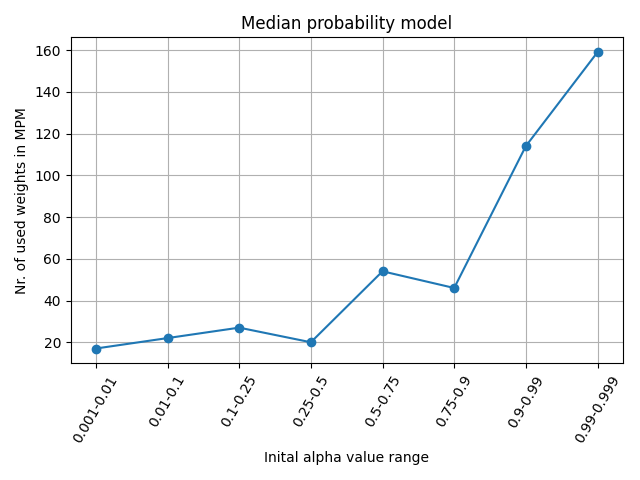}
  \caption{Nr. of used weights}
  \label{fig:nr_used_weights_mpm_lrt}
\end{subfigure}
\caption{Different initialization of $\alpha$ on the Abalone dataset using ISLaB-LRT with the MPM.}
\label{fig:abaloneDiffInitMPMLRT}
\end{figure}

\begin{figure}[htp]
\centering
\begin{subfigure}{.33\textwidth}
  \centering
  \includegraphics[width=.99\linewidth]{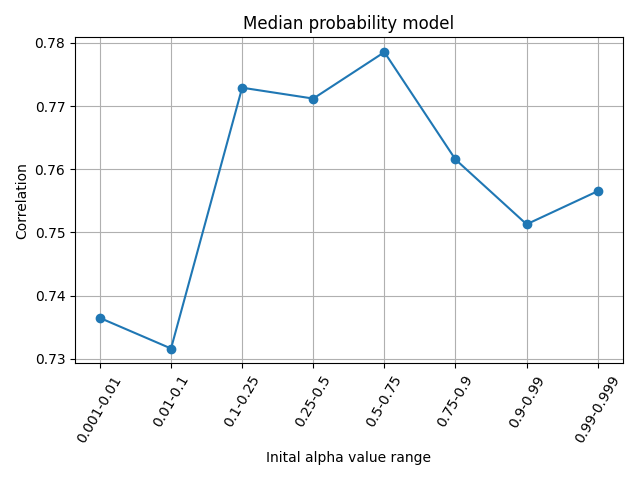}
  \caption{Correlation}
  \label{fig:correlation_mpm_flow}
\end{subfigure}%
\begin{subfigure}{.33\textwidth}
  \centering
  \includegraphics[width=.99\linewidth]{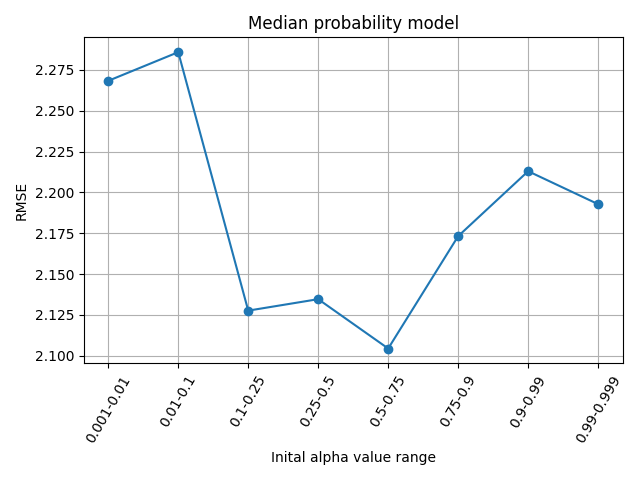}
  \caption{RMSE}
  \label{fig:rmse_mpm_flow}
\end{subfigure}
\begin{subfigure}{.33\textwidth}
  \centering
  \includegraphics[width=.99\linewidth]{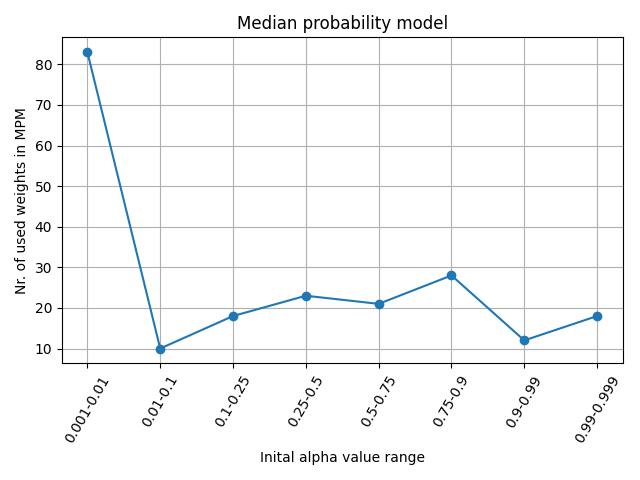}
  \caption{Nr. of used weights}
  \label{fig:nr_used_weights_mpm_flow}
\end{subfigure}
\captionsetup{justification=raggedright,singlelinecheck=false}
\caption{Different initialization of $\alpha$ on the Abalone dataset using ISLaB-FLOW with the MPM.}
\label{fig:abaloneDiffInitMPMFLOW}
\end{figure}

The initial inclusion probability affects the learning procedure, influencing whether the resulting structure is dense or sparse. This, in turn, impacts performance, as suboptimal structures may be chosen. Figures \ref{fig:abaloneDiffInitMPMLRT} and \ref{fig:abaloneDiffInitMPMFLOW} show experiments on the Abalone dataset \citep{nash1995abalone} using different initial inclusion probabilities, $\alpha$, for LRT and normalizing flow implementations. Initializations range from sparse (0.001-0.01) to dense (0.99-0.999), with the remaining hyperparameter as in Table \ref{tab:paramMat} and training procedure matching the Abalone experiments presented in Section \ref{sec:abaloneExp}. Only one model is trained per initialization range.

For LRT models (Figure \ref{fig:abaloneDiffInitMPMLRT}), higher initial density improves both performance and final model size, indicating that LRT based models benefits from more parameters for this specific problem. In contrast, normalizing flows models (Figure \ref{fig:abaloneDiffInitMPMFLOW}) shows no clear pattern; density drops sharply when $\alpha$ increases from (0.001-0.01) to (0.01-0.1), increases from (0.01-0.1) to (0.75-0.9) and drops down again at (0.9-0.99). Performance peaks when $\alpha$ is at (0.5-0.75), despite the densest model occurs at the lowest $\alpha$ values. This illustrates that more weight parameters do not guarantee better performance and that the choice of initialization can be non-trivial for obtaining optimal results.

\section{CIFAR-10 Experiments}\label{appendix:cifar10exp}

ISLaB models can be combined with modern architectures engineered for specific applications in hybrid models.  This is done in transfer learning settings, for instance by extending it with pre-trained vision transformer (ViT) models. Table \ref{tab:CIFAR10} reports results using ViT-B/16 from \citet{dosovitskiy2020image}, pre-trained on the ImageNet dataset, combined with ISLaB using both LRT and normalizing flows (denoted ViT-ISLaB-LRT and ViT-ISLaB-FLOW). The ISLaB part consists of a single hidden layer with 256 nodes, and is trained for 100 epochs. The ViT is kept constant. 

For comparison, we train BCNNs with ISLaB as the feed-forward component. The BCNN models employ a kernel size of 3, stride of 2, and three convolutional layers with 64, 64 and 128 kernels. The feed-forward ISLaB component use two hidden layers with 256 nodes each for LRT and one layer with 512 nodes for normalizing flows. All models were trained without extensive parameter tuning. Results are shown in Table \ref{tab:CIFAR10}.  

The ViT-ISLaB models substantially outperforms the BCNN models. This shows that using pre-trained networks as part of the ISLaB model can significantly improve predictive performance in applications where these pre-trained parts are known to excel. It can be noted that the ECE scores are generally higher for the ViT based models, indicating poorer calibration despite higher accuracy. 

\begin{table}[htp]
    \caption{Results from the CIFAR-10 dataset.}
    \resizebox{\textwidth}{!}{
    \centering
    \begin{tabular}{p{0.25\linewidth}p{0.22\linewidth}p{0.21\linewidth}p{0.25\linewidth}p{0.22\linewidth}}
     \hline
     \multicolumn{5}{c}{Accuracy and ECE score, CIFAR10 dataset} \\
     \hline
     Model & ACC full & ACC sparse &  ECE full &  ECE median \\
     \hline
     ViT-ISLaB-LRT   & 90.7 (90.2, 90.9)\% & 92.4 (92.2, 92.8)\% & 0.204 (0.200, 0.208) & 0.068 (0.064, 0.070) \\ 
     ViT-ISLaB-FLOW  & 95.1 (94.9, 95.3)\% & 75.4 (64.4, 81.8)\% & 0.046 (0.042, 0.048) & 0.051 (0.018, 0.106) \\ 
     BCNN-ISLaB-LRT  & 47.8 (47.5, 48.3)\% & 47.2 (46.6, 47.6)\% & 0.011 (0.007, 0.015) & 0.012 (0.009, 0.024) \\ 
     BCNN-ISLaB-FLOW & 47.8 (45.0, 48.6)\% & 44.9 (38.9, 47.5)\% & 0.025 (0.021, 0.032) & 0.022 (0.013, 0.078) \\ 
     \hline
    \end{tabular}
}
\label{tab:CIFAR10}
\end{table}

\section{Computational Cost and Memory Efficiency}\label{appendix:CompCost}

Table~\ref{tab:computational_cost} provides per iteration training and memory costs for different methods used in this paper.

\begin{table}[htp]
    \caption{Training cost (per iteration of first order optimizer) and Memory Cost (for saving the trained model's parameters) for different methods. Assume (only) for this table the following notation:
 \( n \) - number of data points in the minibatch;
\( p \) - number of trainable parameters in the architecture;
\( p_j \) - number of trainable parameters in layer $j$  of the architecture;
\( l_{j} \) - largest block sizes in the block-diagonal approximation of the Hessian used by Laplace-SpaM used in layer $j$;
$j$  of the architecture;
\( s \) - number of samples from parameters for gradient estimation;
$m$ - the amortization factor indicating how often computation of the marginal likelihood is done;
\( f \) - additional flow parameters;
\( J \) - number of layers;
\( k_j \) - number of neurons at layer $j$;
\( v \) - number of input variables;
\( q \) - additional input skip parameters, where $q = v(k_2,+\dots,+k_{J})$;
\( h \) - for horseshoe priors, we have a dropout parameter (with a mean and variance), but only for the input at each layer, so $h=2(k_1+\dots +k_{J-1})$;
\( w \) - speed-up factor due to sampling directly from neurons (number of weights per neuron);
\( \alpha \) - ratio of non-pruned weights. The training iteration cost reflects the computational complexity for each method. The memory cost includes considerations for additional parameters and storage efficiency, especially for pruned weights. A non-pruned weight is assumed to need a double-precision number to be stored (64 bits). In the pruned model, we further assume to need 1 bit per position of whether the weight there is pruned.}
    \resizebox{\textwidth}{!}{
    \centering
    \begin{tabular}{p{0.22\linewidth}p{0.40\linewidth}p{0.35\linewidth}p{0.38\linewidth}}
     \multicolumn{4}{c}{} \\
     \hline
     Method & Training Cost & Memory Cost(bits) &  Sparse Memory Cost(bits) \\
     \hline
     IS-ANN-L1  & $\mathcal{O}(n \cdot (p+q))$ &$64 \cdot (p+q)$  & $\alpha\cdot (p+q) +65\cdot (1-\alpha)\cdot (p + q)$\\ 
     ISLaB-LRT  &$\mathcal{O}(n \cdot 3(p+q)\cdot s/w)$ &$64 \cdot 3(p+q
     )$ & $\alpha\cdot 2(p+q) +65\cdot (1-\alpha)\cdot 2(p + q)$ \\ 
     ISLaB-FLOW & $\mathcal{O}(n \cdot (3(p+q)+f)\cdot s/w)$&$64 \cdot (3(p+q
     )+f)$&  $\alpha\cdot 2(p+q) +65\cdot ((1-\alpha)\cdot 2(p + q)+f)$ \\ 
     BLR-LRT & $\mathcal{O}(n\cdot 3v \cdot s/w)$& $64\cdot3v$ & $\alpha\cdot 2v+65\cdot (1-\alpha)\cdot 2v$\\
     BLR-FLOW & $\mathcal{O}(n\cdot (3v+f) \cdot s/w)$ &$64\cdot(3v+f)$ & $\alpha\cdot 2v+65\cdot (\ (1-\alpha)\cdot 2v + f)$\\
     BNN-HORSE & $\mathcal{O}(n\cdot (2p+h))$ &$64\cdot (2p+h)$& $\alpha\cdot 2p +65\cdot (1-\alpha)\cdot 2p  $\\
     BNN-CONCRETE & $\mathcal{O}(n\cdot (p + J))$ &$64\cdot (p + J)$& NA\\
     Laplace-SpaM &$\mathcal{O}(n \cdot p  + m^{-1}\cdot (k_1\cdot l_{1}^3+\dots+k_J\cdot l_{J}^3))$ & $64 \cdot (p + k_1\cdot l_{1}^2+\dots +k_J\cdot l_{J}^2)$&$\alpha\cdot p +64\cdot (1-\alpha)\cdot p   + k_1\cdot l_{1}^2+\dots +k_J\cdot l_{J}^2$\\

     \hline
    \end{tabular}
}
\label{tab:computational_cost}
\end{table}

\end{document}